\documentclass[nohyperref]{article}

\usepackage{microtype}
\usepackage{graphicx}
\usepackage{subfigure}
\usepackage{booktabs} 

\usepackage{hyperref}

\usepackage[accepted]{icml2022}

\usepackage{amsmath}
\usepackage{amssymb}
\usepackage{mathtools}
\usepackage{amsthm}

\usepackage[capitalize,noabbrev]{cleveref}

\theoremstyle{plain}

\theoremstyle{definition}

\theoremstyle{remark}

\usepackage[textsize=small]{todonotes}

\icmltitlerunning{Neural-Symbolic Models for Logical Queries on Knowledge Graphs}

\usepackage{tikz}
\usepackage{enumitem}
\usepackage{xspace}
\usepackage{xcolor}
\usepackage{wrapfig}
\usepackage{caption}
\usepackage{amssymb}
\usepackage{multirow}
\usepackage{booktabs}
\usepackage{adjustbox}
\usepackage{thmtools}
\usepackage{thm-restate}

\newcommand{\INDSTATE}[1][1]{\STATE\hspace{#1em}}

\newcommand{\method}{GNN-QE\xspace}

\newcommand{\stack}[1]{\begin{tabular}{@{}c@{}}#1\end{tabular}}

\usepackage{pifont}
\newcommand{\cmark}{\ding{51}}
\newcommand{\xmark}{\ding{55}}
\definecolor{pltred}{RGB}{214, 39, 40}
\definecolor{pltgreen}{RGB}{44, 160, 44}
\definecolor{pltblue}{RGB}{31, 119, 180}
\newcommand{\easy}[1]{#1}
\newcommand{\tp}[1]{#1 \cmark}
\newcommand{\fp}[1]{#1 \xmark}

\usepackage{amsmath,amsfonts,bm}

\def\eqref#1{equation~\ref{#1}}

\def\1{\bm{1}}

\def\vb{{\bm{b}}}

\def\vh{{\bm{h}}}

\def\vq{{\bm{q}}}

\def\vx{{\bm{x}}}
\def\vy{{\bm{y}}}

\def\mW{{\bm{W}}}

\DeclareMathAlphabet{\mathsfit}{\encodingdefault}{\sfdefault}{m}{sl}
\SetMathAlphabet{\mathsfit}{bold}{\encodingdefault}{\sfdefault}{bx}{n}

\def\gA{{\mathcal{A}}}

\def\gC{{\mathcal{C}}}
\def\gD{{\mathcal{D}}}
\def\gE{{\mathcal{E}}}

\def\gG{{\mathcal{G}}}

\def\gL{{\mathcal{L}}}

\def\gN{{\mathcal{N}}}

\def\gP{{\mathcal{P}}}

\def\gR{{\mathcal{R}}}

\def\gU{{\mathcal{U}}}
\def\gV{{\mathcal{V}}}

\begin{document}

\twocolumn[
\icmltitle{Neural-Symbolic Models for Logical Queries on Knowledge Graphs}



\icmlsetsymbol{equal}{*}

\begin{icmlauthorlist}
\icmlauthor{Zhaocheng Zhu}{mila,udem}
\icmlauthor{Mikhail Galkin}{mila,mcgill}
\icmlauthor{Zuobai Zhang}{mila,udem}
\icmlauthor{Jian Tang}{mila,hec,cifar}
\end{icmlauthorlist}

\icmlaffiliation{mila}{Mila - Qu\'ebec AI Institute}
\icmlaffiliation{udem}{Universit\'e de Montr\'eal}
\icmlaffiliation{mcgill}{McGill University}
\icmlaffiliation{hec}{HEC Montr\'eal}
\icmlaffiliation{cifar}{CIFAR AI Chair}

\icmlcorrespondingauthor{Zhaocheng Zhu}{zhaocheng.zhu@umontreal.ca}
\icmlcorrespondingauthor{Jian Tang}{jian.tang@hec.ca}

\icmlkeywords{ICML, Machine Learning, Knowledge Graphs, Graph Neural Networks, Fuzzy Sets}

\vskip 0.3in
]



\printAffiliationsAndNotice{}  

\begin{abstract}

Answering complex first-order logic (FOL) queries on knowledge graphs is a fundamental task for multi-hop reasoning. Traditional symbolic methods traverse a complete knowledge graph to extract the answers, which provides good interpretation for each step. Recent neural methods learn geometric embeddings for complex queries. These methods can generalize to incomplete knowledge graphs, but their reasoning process is hard to interpret. In this paper, we propose Graph Neural Network Query Executor (\method), a neural-symbolic model that enjoys the advantages of both worlds. \method decomposes a complex FOL query into relation projections and logical operations over fuzzy sets, which provides interpretability for intermediate variables. To reason about the missing links, \method adapts a graph neural network from knowledge graph completion to execute the relation projections, and models the logical operations with product fuzzy logic. Experiments on 3 datasets show that \method significantly improves over previous state-of-the-art models in answering FOL queries. Meanwhile, \method can predict the number of answers without explicit supervision, and provide visualizations for intermediate variables.\footnote{Code is available at \url{https://github.com/DeepGraphLearning/GNN-QE}}

\end{abstract}

\section{Introduction}

\begin{figure*}
    \centering
    \includegraphics[width=\textwidth]{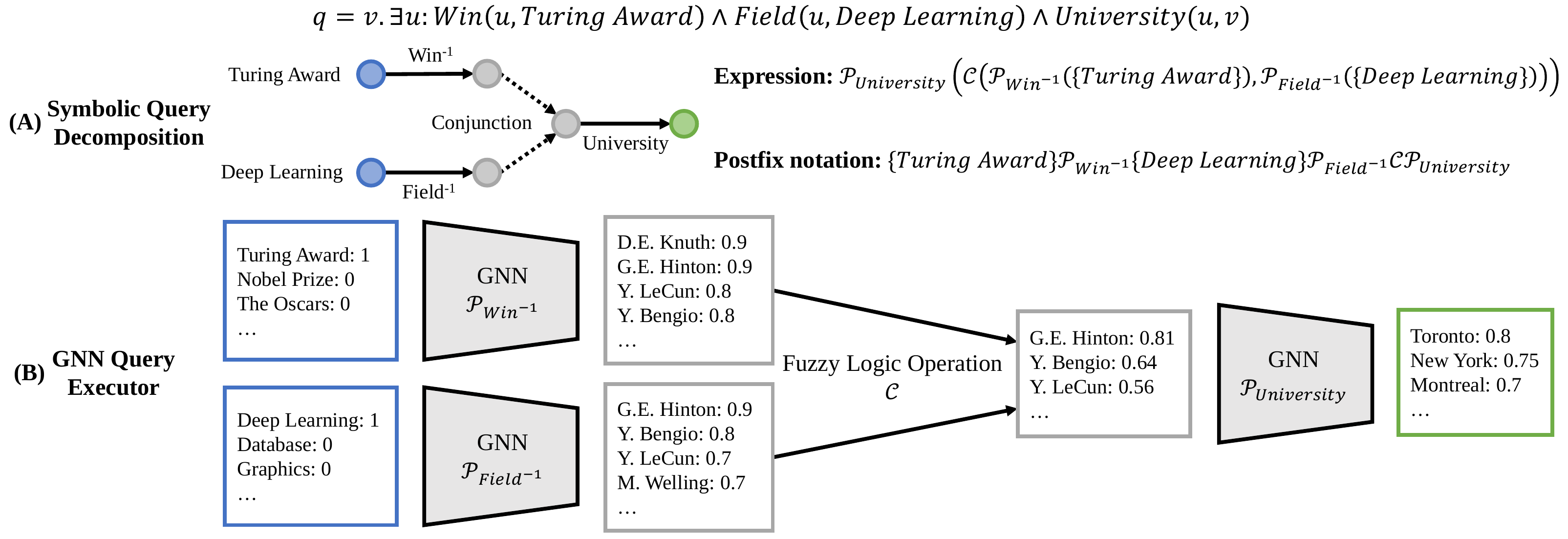}
    \caption{Overview of \method. \textbf{(A)} \method decomposes a FOL query into an expression of relation projections ($\gP$) and logic operations ($\gC, \gD, \gN$). We convert the query into an expression execution problem, where we use the postfix notation to efficiently batch multiple expressions. \textbf{(B)} The expression is executed with relation projection learned by GNNs and fuzzy logic operations. All the input, intermediate and output variables are fuzzy sets of entities. Best viewed in color.}
    \label{fig:method}
\end{figure*}

Knowledge graphs (KGs) encapsulate knowledge about the world in a collection of relational edges between entities, and are widely adopted by many domains~\cite{miller1998wordnet, vrandevcic2014wikidata, himmelstein2017systematic, szklarczyk2019string}. Reasoning on knowledge graphs has attracted much attention in artificial intelligence, since it can be used to infer new knowledge or answer queries based on existing knowledge. One particular reasoning task we are interested in is answering complex First-Order Logic (FOL) queries on knowledge graphs, which involves logic operations like existential quantifier ($\exists$), conjunction ($\land$), disjunction ($\lor$) and negation ($\neg$). For example, the question ``\emph{Which universities do the Turing Award winners of deep learning work in?}'' can be represented as a FOL query, as showed in Fig.~\ref{fig:method}.

Traditionally, the problem of reasoning is handled by symbolic approaches, such as logic programming~\cite{lloyd2012foundations}, fuzzy logic~\cite{klir1995fuzzy} or probabilistic reasoning~\cite{pearl2014probabilistic}. In the same vein, several algorithms~\cite{dalvi2007efficient, schmidt2010foundations, zou2011gstore} have been developed for searching the answers to complex queries on graph databases. These methods traverse a graph and extract all possible assignments for intermediate variables, which provides good interpretation for each step. Besides, symbolic methods are guaranteed to produce the correct answer if all facts are given~\cite{stuart2016artificial}. However, many real-world knowledge graphs are known to be incomplete~\cite{nickel2015review}, which limits the usage of symbolic methods on knowledge graphs.

Recently, neural methods, such as embedding methods~\cite{bordes2013translating, trouillon2016complex, sun2018rotate} and graph neural networks (GNNs)~\cite{schlichtkrull2018modeling, vashishth2019composition, teru2020inductive, zhu2021neural}, have achieved significant progress in knowledge graph completion. Based on the success of these neural methods, many works have been proposed to solve FOL queries on incomplete graphs by learning an embedding for each FOL query~\cite{hamilton2018embedding, ren2019query2box, ren2020beta, chen2021fuzzy, zhang2021cone}. Typically, these methods 
translate the logic operations into neural logic operators in the embedding space. Nevertheless, it is hard to interpret what set of entities an intermediate embedding encodes, leaving the reasoning process unknown to users. The only interpretable method is CQD-Beam~\cite{arakelyan2020complex}, which applies beam search to a pretrained embedding model in the entity space. However, the complexity of exhaustive search prevents CQD-Beam from being trained directly on complex queries.

In this paper, we marry the advantages from both neural and symbolic approaches, and propose Graph Neural Network Query Executor (\method), a neural-symbolic method for answering FOL queries on incomplete knowledge graphs. Following symbolic methods that output a set of assignments for each intermediate variable, we decompose a complex FOL query into an expression over fuzzy sets (i.e., a continuous relaxation of sets), which attains interpretability for intermediate variables. Each basic operation in the expression is either a relation projection or a logic operation (e.g., conjunction, disjunction and negation). We design the relation projection to be a GNN that predicts the fuzzy set of tail entities given a fuzzy set of head entities and a relation. The logic operations are transformed to the product fuzzy logic operations over fuzzy sets, which satisfy logic laws and enable differentiation of logic operations. We also propose traversal dropout to regularize the model, and batch expression execution to speed up training and inference.

We evaluate our method on 3 standard datasets for FOL queries. Experiments show that \method achieves new state-of-the-art performance on all datasets, with an average relative gain of 22.3\% on existential positive first-order (EPFO) queries and 95.1\% on negation queries (Sec.~\ref{sec:main_experiment}). By disentangling the contribution of knowledge graph completion and complex query framework, we find that \method achieves one of the best generalization performances from knowledge graph completion to EPFO queries among different methods. Additionally, the symbolic formulation of our method enables us to predict the number of answers without explicit supervision (Sec.~\ref{sec:cardinality}), and visualize intermediate variables (Sec.~\ref{sec:visualization} \& App.~\ref{app:visualization}). The visualization provided by \method may help us better understand the reasoning process taken by the model, leading to more interpretable multi-hop reasoning.
\section{Related Work}

\textbf{Knowledge Graph Completion}
Recent years have witnessed a significant progress in reasoning about missing links on a knowledge graph. Notably, embedding methods~\cite{bordes2013translating, yang2014embedding, trouillon2016complex, sun2018rotate, amin2020lowfer} learn a low-dimensional vector for each entity and relation, which preserves the structure of the knowledge graph. Reinforcement learning methods~\cite{xiong2017deeppath, das2017go, hildebrandt2020reasoning, zhang2021learning} train an agent to collect necessary paths for predicting the link between entities. Rule learning methods~\cite{yang2017differentiable, sadeghian2019drum, qu2020rnnlogic} first extract interpretable logic rules from the knowledge graph, and then use the rules to predict the links. Another stream of works adopts graph neural networks (GNNs) to learn the entity representations~\cite{schlichtkrull2018modeling, vashishth2019composition}, or the pairwise representations~\cite{teru2020inductive, zhu2021neural} for knowledge graph completion. Our method adapts a GNN from knowledge graph completion~\cite{zhu2021neural} to implement the relation projection on knowledge graphs. However, \method is designed to answer complex logical queries, a more challenging task than KG completion.

\textbf{Complex Logical Query}
Complex logical query extends knowledge graph completion to predict answer entities for queries with conjunction, disjunction or negation operators. \citet{guu2015traversing} proposes compositional training for embedding methods to predict answers for path queries. GQE~\cite{hamilton2018embedding} learns a geometric intersection operator to answer conjunctive queries ($\land$) in the embedding space, which is later extended by Query2Box~\cite{ren2019query2box} to EPFO queries ($\exists$, $\land$, $\lor$) and BetaE~\cite{ren2020beta} to FOL queries ($\exists$, $\land$, $\lor$, $\neg$). FuzzQE~\cite{chen2021fuzzy} improves embedding methods with t-norm fuzzy logic, which satisfies the axiomatic system of classical logic. Some recent works utilize advanced geometric embeddings to achieve desired properties for operators, e.g., hyperboloid embeddings in HypE~\cite{choudhary2021self} and cone embeddings in ConE~\cite{zhang2021cone}. Generally, all these methods compute an embedding for the query, and decode the answers with nearest neighbor search or dot product. However, the interpretability of embedding methods is usually compromised, i.e., there is no simple way to understand intermediate reasoning results. 

Some other works combine neural methods with symbolic algorithms to solve the complex query answering problem. EmQL~\cite{sun2020faithful} ensembles an embedding model and a count-min sketch, and is able to find logically entailed answers. CQD~\cite{arakelyan2020complex} extends a pretrained knowledge graph embedding model to infer answers for complex queries, with CQD-CO based on continuous optimization and CQD-Beam based on beam search. Our method shares a similar spirit with CQD-Beam in the sense that both models wrap a knowledge graph completion model with symbolic algorithms. However, CQD-Beam cannot be directly trained on complex query due to the complexity incurred by exhaustive search. By contrast, \method is trained directly on complex queries without pretrained embedding models.

\section{Preliminary}
In this section, we introduce the background knowledge of FOL queries on knowledge graphs and fuzzy sets.

\subsection{First-Order Logic Queries on Knowledge Graphs}
Given a set of entities $\gV$ and a set of relations $\gR$, a knowledge graph $\gG=(\gV, \gE, \gR)$ is a collection of triplets $\gE=\{(h_i, r_i, t_i)\}\subseteq \gV \times \gR \times \gV$, where each triplet is a fact from head entity $h_i$ to tail entity $t_i$ with the relation type $r_i$.

A FOL query on a knowledge graph is a formula composed of constants (denoted with English terms), variables (denoted with $a$, $b$, $c$), relation symbols (denoted with $R(a, b)$) and logic symbols ($\exists$, $\land$, $\lor$, $\neg$). In the context of knowledge graphs, each constant or variable is an entity in $\gV$. A variable is bounded if it is quantified in the expression, and free otherwise. Each relation symbol $R(a, b)$ is a binary function that indicates whether there is a relation $R$ between a pair of constants or variables. For logic symbols, we consider queries that contain conjunction ($\land$), disjunction ($\lor$), negation ($\neg$) and existential quantification ($\exists$)\footnote{Note universal quantification ($\forall$) is excluded, since none of the entities connects to all entities in a real-world knowledge graph.}. Fig.~\ref{fig:method} illustrates the FOL query for the natural language question ``\emph{Which universities do the Turing Award winners of deep learning work in?}''. Given a FOL query, the goal is to find answers to the free variables, such that the formula is true.

\subsection{Fuzzy Sets and Fuzzy Logic Operations}
Fuzzy sets~\cite{klir1995fuzzy} are a continuous relaxation of sets whose elements have degrees of membership. A fuzzy set $\gA=(\gU, x)$ contains a universal set $\gU$ and a membership function $x: \gU \to [0, 1]$. For each $u \in \gU$, the value of $x(u)$ defines the degree of membership (i.e., probability) for $u$ in $\gA$. Similar to Boolean logic, fuzzy logic defines three logic operations, \text{AND}, \text{OR} and \text{NOT}, over the real-valued degree of membership. There are several alternative definitions for these operations, such as product fuzzy logic, Gödel fuzzy logic and Łukasiewicz fuzzy logic.

In this paper, fuzzy sets are used to represent the assignments of variables in FOL queries, where the universe $\gU$ is always the set of entities $\gV$ in the knowledge graph. Since the universe is a finite set, we represent the membership function $x$ as a vector $\vx$. We use $x_u$ to denote the degree of membership for element $u$. For simplicity, we abbreviate a fuzzy set $\gA=(\gU, x)$ as $\vx$ throughout the paper.
\section{Proposed Method}

Here we present our model, Graph Neural Network Query Executor (\method). The high-level idea of \method is to first decompose a FOL query into an expression of 4 basic operations (relation projection, conjunction, disjunction and negation) over fuzzy sets, then parameterize the relation projection with a GNN adapted from KG completion, and instantiate the logic operations with product fuzzy logic operations. Besides, we introduce traversal dropout to prevent the GNN from converging to a trivial solution, and batched expression execution for speeding up training and inference.

\subsection{Symbolic Query Decomposition}

Given a FOL query, the first step is to convert it into an expression of basic operations, so that we can retrieve answers by executing the expression. Previous works define basic operations as either relation projections and logic operations over \emph{embeddings}~\cite{ren2019query2box, ren2020beta, chen2021fuzzy, zhang2021cone}, or a score function over triplets~\cite{arakelyan2020complex}. To achieve better interpretability for intermediate variables, we explicitly define 4 basic operations over \emph{fuzzy sets of entities} as follows

\begin{itemize}[leftmargin=*, nosep]
    \item \textbf{Relation Projection}: $\gP_q(\vx)$ computes the fuzzy set of \emph{tail} entities that are reachable by the input fuzzy set of \emph{head} entities through relation $q$. $\gP_{q^{-1}}(\vx)$ computes the fuzzy set of \emph{head} entities that can reach the input fuzzy set of \emph{tail} entities through relation $q$.
    \item \textbf{Conjunction}: $\gC(\vx, \vy)$ computes the logical conjunction for each element in $\vx$ and $\vy$.
    \item \textbf{Disjunction}: $\gD(\vx, \vy)$ computes the logical disjunction for each element in $\vx$ and $\vy$.
    \item \textbf{Negation}: $\gN(\vx)$ computes the logical negation for each element in $\vx$.
\end{itemize}

where $\vx, \vy \in [0,1]^\gV$ are two vector representations of fuzzy sets. We then decompose a FOL query into an expression of the above operations. For the example in Fig.~\ref{fig:method}, the corresponding expression is
\begin{align}
    \label{eqn:decomposition}
    \hspace{-0.5em}
    \begin{adjustbox}{max width=0.92\columnwidth}
        $\gP_\textit{University}\left(\gC\left(\gP_{\textit{Win}^{-1}}(\{\textit{Turing Award}\}), \gP_{\textit{Field}^{-1}}(\{\textit{Deep Learning}\})\right)\right)$
    \end{adjustbox}
\end{align}
where \{\textit{Turing Award}\} and \{\textit{Deep Learning}\} denote singleton sets of \emph{Turing Award} and \emph{Deep Learning}, respectively.

\subsection{Neural Relation Projection}

In order to solve complex queries on incomplete knowledge graphs, we learn a neural model to perform the relation projection $\vy = \gP_q(\vx)$. Specifically, the neural relation projection model should predict the fuzzy set of tail entities $\vy$ given the fuzzy set of head entities $\vx$ and a relation $q$ in the presence of missing links. This is in contrast to the common GNNs~\cite{schlichtkrull2018modeling, vashishth2019composition} and embedding methods~\cite{bordes2013translating, sun2018rotate} for knowledge graph completion, which operate on individual entities $x$ and $y$. While it is possible to apply such GNNs or embedding methods for relation projection, it takes at least $O(|\gV|^2d)$ time to compute them for every $x \in \vx$ and $y \in \vy$, which is not scalable.

Recently, \citet{zhu2021neural} introduced a new GNN framework for knowledge graph completion, which can predict the set of tail entities $\vy$ given an entity $x$ and a relation $q$ in $O(|\gV|d^2 + |\gE|d)$ time. Inspired by such a framework, we propose a scalable GNN solution for relation projection.

\textbf{Graph Neural Networks.}
Our goal is to design a GNN model that predicts a fuzzy set of tail entities given a fuzzy set of head entities and a relation. A special case of the input is a singleton set, where we need to model the probability $p_q(y|x)$ for every $y \in \vy$. Such a problem can be solved by GNNs in a single-source fashion~\cite{you2021identity, zhu2021neural}. For example, the recent work NBFNet~\cite{zhu2021neural} derives a GNN framework based on the generalized Bellman-Ford algorithm for single-source problems on graphs. Given a head entity $u$ and a projection relation $q$, we use the following iteration to compute a representation $\vh_v$ for each entity $v \in \gV$ w.r.t.\ the source entity $u$:
\begin{align}
    &\vh^{(0)}_v \leftarrow \textsc{Indicator}(u, v, q) \label{eqn:boundary} \\
    &\begin{aligned}
    \vh^{(t)}_v \leftarrow \textsc{Aggregate}(\{&\textsc{Message}(\vh^{(t-1)}_z, (z, r, v)) \\
    &| (z, r, v)\in\gE(v)\}) \label{eqn:bellman}
    \end{aligned}
\end{align}
where the \textsc{Indicator} function initializes a relation embedding $\vq$ on entity $v$ if $v$ equals to $u$ and a zero embedding otherwise, and $\gE(v)$ is the set of edges going into $v$. The \textsc{Message} and \textsc{Aggregate} functions can be instantiated with any neural function from popular GNNs. To apply the above framework to a fuzzy set $\vx$ of head entities, we propose to replace Eqn.~\ref{eqn:boundary} with the following initialization
\begin{align}
    &\vh^{(0)}_v \leftarrow x_v \vq
    \label{eqn:fuzzy_set}
\end{align}
where $x_v$ is the probability of entity $v$ in $\vx$. Intuitively, this GNN model initializes an embedding $\vq$ for the projection relation $q$ on all entities, where the scale of the initialization on an entity depends on its probability in the fuzzy set. The original \textsc{Indicator} function can be viewed as a special case of Eqn.~\ref{eqn:fuzzy_set}, with the fuzzy set being a singleton set.

For the \textsc{Aggregate} and the \textsc{Message} functions, we follow the design in NBFNet~\cite{zhu2021neural} and parameterize the \textsc{Message} function as
\begin{align}
    \hspace{-0.5em}
    \begin{adjustbox}{max width=0.92\columnwidth}
        $\textsc{Message}(\vh^{(t-1)}_z, (z, r, v)) = \vh^{(t-1)}_z \odot (\mW_r \vq + \vb_r)$
    \end{adjustbox}
    \label{eqn:message}
\end{align}
where $\mW^{(t)}_r$ and $\vb^{(t)}_r$ are the weight matrix and bias vector for relation $r$ in the $t$-th iteration respectively, and $\odot$ is the element-wise multiplication operator. The \textsc{Aggregate} function is parameterized as the principal neighborhood aggregation (PNA)~\cite{corso2020principal}. Our GNN has the same time complexity as NBFNet, and therefore takes $O(|\gV|d^2 + |\gE|d)$ time for each message passing iteration. Note it is possible to parameterize the framework with other GNN models, such as RGCN~\cite{schlichtkrull2018modeling} or CompGCN~\cite{vashishth2019composition}. See Sec.~\ref{sec:ablation} for experiments with different GNN models.

To apply the GNN framework for relation projection, we propagate the representations with Eqn.~\ref{eqn:bellman} for $T$ layers. Then we take the representations in the last layer, and pass them into a multi-layer perceptron (MLP) $f$ followed by a sigmoid function $\sigma$ to predict the fuzzy set of tail entities.
\begin{align}
    \gP_q(\vx) = \sigma(f(\vh^{(T)}))
    \label{eqn:predict}
\end{align}

\subsection{Fuzzy Logic Operations}

The logic operations (i.e., $\gC(\vx, \vy)$, $\gD(\vx, \vy)$, $\gN(\vx)$) glue multiple relation projection results and generate the input fuzzy set for the next relation projection. Ideally, they should satisfy certain logic laws, such as commutativity, associativity 
and non-contradiction. Most previous works~\cite{hamilton2018embedding, ren2019query2box, ren2020beta, zhang2021cone} propose dedicated geometric operations to learn these logic operations in the embedding space. Nevertheless, these neural operators are not guaranteed to satisfy most logic laws, which may introduce additional error when they are chained together. 

Here we model the conjunction, disjunction and negation with product fuzzy logic operations. Given two fuzzy sets $\vx, \vy \in [0,1]^\gV$, the operations are defined as follows
\begin{align}
    \gC(\vx, \vy) &= \vx \odot \vy \\
    \gD(\vx, \vy) &= \vx + \vy - \vx \odot \vy \\
    \gN(\vx) &= \bm{1} - \vx
\end{align}
where $\odot$ is the element-wise multiplication and $\bm{1}$ is a vector of all ones (i.e., the universe). Compared to geometric operations in previous works, such fuzzy logic operations satisfy many logic laws, e.g., De Morgan's laws $\gN(\gC(\vx, \vy)) = \gD(\gN(\vx), \gN(\vy))$, $\gN(\gD(\vx, \vy)) = \gC(\gN(\vx), \gN(\vy))$. Note FuzzQE~\cite{chen2021fuzzy} also adopts fuzzy logic operations and satisfies logic laws. However, FuzzQE applies fuzzy logic operations to \emph{embeddings}. By contrast, our \method applies fuzzy logic operations to \emph{fuzzy sets of entities}, which provides better interpretability (See Sec.~\ref{sec:visualization}).

\subsection{Learning}
Following previous works~\cite{ren2019query2box, ren2020beta, zhang2021cone}, we train our model to minimize the binary cross entropy loss.
\begin{align}
    \begin{aligned}
    \gL = &-\frac{1}{|\gA_Q|}\sum_{a \in \gA_Q}\log p(a|Q) \\
    &-\frac{1}{|\gV\backslash\gA_Q|}\sum_{a' \in \gV\setminus\gA_Q}\log (1 - p(a'|Q))
    \end{aligned}
    \label{eqn:loss}
\end{align}
where $\gA_Q$ is the set of answers to the complex query $Q$ and $p(a|Q)$ is the probability of entity $a$ in the final output fuzzy set. Since \method always outputs the probability for all entities (Eqn.~\ref{eqn:predict}), we do not perform negative sampling and compute the loss with all negative answers.

\textbf{Traversal Dropout.} One challenge in training \method is to let the model generalize to incomplete KGs at test time. This is because all the training queries are generated by assuming the training graph is complete~\cite{ren2020beta}. In other words, all the training queries can be perfectly solved by a simple relation traversal model on the training graph, without modeling any missing link. GNN models can easily discover this mode, which does not generalize to incomplete knowledge graphs at test time.

To solve this issue, we introduce traversal dropout to create an incomplete KG at training time. Specifically, we first run a relation traversal model to extract all the edges corresponding to the query. We then randomly mask out the traversed edges in each relation projection with probability $p$. Intuitively, the probability $p$ trades off between a simple relation traversal model and a full reasoning model. If $p$ is small, the GNN model may converge to a trivial relation traversal model, otherwise it is forced to encode non-trivial reasoning features. Since some of the edges in the test queries may be present in the KG, it is not always optimal to use a large $p$ to discourage a relation traversal model. In practice, we treat $p$ as a hyperparameter, and tune it based on the performance on the validation set. See Sec.~\ref{sec:ablation} for experiments with different values of $p$.

\textbf{Batched Expression Execution\footnote{Expression execution is formally known as expression evaluation in computer science. In this paper, we use the term ``expression execution'' to avoid ambiguity in machine learning contexts.}.}
Modern machine learning relies on batch processing on GPUs to accelerate the computation of neural (or even symbolic) models. However, it is challenging to batch the expressions of FOL queries, since different query structures require different recursive computation steps. Previous works~\cite{hamilton2018embedding, ren2019query2box, ren2020beta} divide a batch based on the query structure of each sample, and only batch the computation of samples that have the same structure. However, such an implementation needs to enumerate every query structure, and is not scalable when the vocabulary of query structures grows large.

To solve this issue, we need to find a way to execute the expressions without recursion. This can be achieved by converting the expressions into postfix notation. The postfix notation, a.k.a.\ reverse Polish notation~\cite{lukasiewicz1951aristotle}, writes operators \emph{after} their operands in an expression. For example, the postfix expression of Eqn.~\ref{eqn:decomposition} is
\begin{align}
    \hspace{-0.5em}
    \begin{adjustbox}{max width=0.9\columnwidth}
        $\{\textit{Turing Award}\}\gP_{\textit{Win}^{-1}}\{\textit{Deep Learning}\}\gP_{\textit{Field}^{-1}}\gC\gP_\textit{University}$
    \end{adjustbox}
\end{align}
The advantage of postfix expressions is that they are unambiguous without parentheses, and therefore can be executed easily without recursion. To execute a postfix expression, we allocate a stack and scan the expression from left to right. When we encounter an operand, we push it into the stack. When we encounter an operator, we pop the corresponding number of operands from the stack, apply the operation and push the result into the stack. Such an algorithm can be easily batched for the same operator even in samples of different query types. Examples and pseudo code for batched expression execution are provided in App.~\ref{app:code}.
\section{Experiments}

In this section, we evaluate \method by answering FOL queries on 3 standard datasets. Our experiments demonstrate that: (1) \method outperforms existing methods on both EPFO queries and queries with negation. (2) \method can predict the number of answers out-of-the-box without any explicit supervision. (3) We can visualize the intermediate variables of \method and interpret its reasoning process.

\subsection{Experiment Setup}

We evaluate our method on FB15k~\cite{bordes2013translating}, FB15k-237~\cite{toutanova2015observed} and NELL995~\cite{xiong2017deeppath} knowledge graphs. To make a fair comparison with baselines, we use the standard train, validation and test FOL queries generated by the BetaE paper~\cite{ren2020beta}, which consist of 9 EPFO query types and 5 query types with negation. We follow previous works~\cite{ren2020beta, chen2021fuzzy, zhang2021cone} and train our model with 10 query types (\emph{1p/2p/3p/2i/3i/2in/3in/inp/pni/pin}). The model is evaluated on 10 training query types, plus 4 query types (\emph{ip/pi/2u/up}) that have never been seen during training. A full list of query types and their statistics is provided in App.~\ref{app:datasets}.

\textbf{Evaluation Protocol.}
Following the evaluation protocol in \cite{ren2019query2box}, we separate the answers to each query into two sets: easy answers and hard answers. For test (validation) queries, easy answers are the entities that can be reached on the validation (train) graph via a symbolic relation traverse model. Hard answers are those that can only be reached with predicted links. In other words, the model must perform reasoning to get the hard answers. We compute the ranking of each hard answer against all non-answer entities. The performance is measured by mean reciprocal rank (MRR) and HITS at K (H@K) metrics.

\textbf{Implementation Details.}
Our work is implemented based on the open-source codebase of GNNs for KG completion\footnote{\url{https://github.com/DeepGraphLearning/NBFNet}}. Following \cite{zhu2021neural}, we augment each triplet with a flipped one of its inverse relation, so that the GNN can propagate information in both directions. The neural relation projection model is set to a 4-layer GNN model. We train the model with the self-adversarial negative sampling~\cite{sun2018rotate}. Note we only instantiate 1 GNN model and share it across all neural relation projections in the query. For query types that contain multiple relation projections in a chain (\emph{2p/3p/inp/pni/pin}), we observe very noisy gradients for the relation projections early in the chain. Therefore, we zero out the gradients of those relation projections, and only update the GNN with gradients from the last relation projections close to the loss. Our model is trained with Adam optimizer~\cite{kingma2014adam} on 4 Tesla V100 GPUs. Hyperparameters of \method are given in App.~\ref{app:hyperparameter}.

\textbf{Baselines.} We compare \method against both embedding methods and neural-symbolic methods. The embedding methods include GQE~\cite{hamilton2018embedding}, Q2B~\cite{ren2019query2box}, BetaE~\cite{ren2020beta}, FuzzQE~\cite{chen2021fuzzy} and ConE~\cite{zhang2021cone}. The neural-symbolic methods include CQD-CO~\cite{arakelyan2020complex} and CQD-Beam~\cite{arakelyan2020complex}. For CQD-CO and CQD-Beam, we obtain their performance using the codebase\footnote{\url{https://github.com/pminervini/KGReasoning}} provided by the original authors.

\begin{table*}[t]
    \centering
    \caption{Test MRR results (\%) on answering FOL queries. avg$_p$ is the average MRR on EPFO queries ($\land$, $\lor$). avg$_n$ is the average MRR on queries with negation. Results of GQE and Q2B are taken from \cite{ren2020beta}. Results of BetaE, FuzzQE and ConE are taken from their original papers. Results of other metrics can be found in App.~\ref{app:result}.}
    \begin{adjustbox}{width=\textwidth}
    \begin{tabular}{lcccccccccccccccc}
        \toprule
        \bf{Model} & \bf{avg$_p$} & \bf{avg$_n$} & \bf{1p} & \bf{2p} & \bf{3p} & \bf{2i} & \bf{3i} & \bf{pi} & \bf{ip} & \bf{2u} & \bf{up} & \bf{2in} & \bf{3in} & \bf{inp} & \bf{pin} & \bf{pni} \\
        \midrule
        \multicolumn{17}{c}{FB15k} \\
        \midrule
        GQE & 28.0 & - & 54.6 & 15.3 & 10.8 & 39.7 & 51.4 & 27.6 & 19.1 & 22.1 & 11.6 & - & - & - & - & - \\
        Q2B & 38.0 & - & 68.0 & 21.0 & 14.2 & 55.1 & 66.5 & 39.4 & 26.1 & 35.1 & 16.7 & - & - & - & - & - \\
        BetaE & 41.6 & 11.8 & 65.1 & 25.7 & 24.7 & 55.8 & 66.5 & 43.9 & 28.1 & 40.1 & 25.2 & 14.3 & 14.7 & 11.5 & 6.5 & 12.4 \\
        CQD-CO & 46.9 & - & \bf{89.2} & 25.3 & 13.4 & 74.4 & 78.3 & 44.1 & 33.2 & 41.8 & 21.9 & - & - & - & - & - \\
        CQD-Beam & 58.2 & - & \bf{89.2} & 54.3 & 28.6 & 74.4 & 78.3 & 58.2 & 67.7 & 42.4 & 30.9 & - & - & - & - & - \\
        ConE & 49.8 & 14.8 & 73.3 & 33.8 & 29.2 & 64.4 & 73.7 & 50.9 & 35.7 & 55.7 & 31.4 & 17.9 & 18.7 & 12.5 & 9.8 & 15.1 \\
        \midrule
        \method & \bf{72.8} & \bf{38.6} & 88.5 & \bf{69.3} & \bf{58.7} & \bf{79.7} & \bf{83.5} & \bf{69.9} & \bf{70.4} & \bf{74.1} & \bf{61.0} & \bf{44.7} & \bf{41.7} & \bf{42.0} & \bf{30.1} & \bf{34.3} \\
        \midrule[0.08em]
        \multicolumn{17}{c}{FB15k-237} \\
        \midrule
        GQE & 16.3 & - & 35.0 & 7.2 & 5.3 & 23.3 & 34.6 & 16.5 & 10.7 & 8.2 & 5.7 & - & - & - & - & - \\
        Q2B & 20.1 & - & 40.6 & 9.4 & 6.8 & 29.5 & 42.3 & 21.2 & 12.6 & 11.3 & 7.6 & - & - & - & - & - \\
        BetaE & 20.9 & 5.5 & 39.0 & 10.9 & 10.0 & 28.8 & 42.5 & 22.4 & 12.6 & 12.4 & 9.7 & 5.1 & 7.9 & 7.4 & 3.5 & 3.4 \\
        CQD-CO & 21.8 & - & \bf{46.7} & 9.5 & 6.3 & 31.2 & 40.6 & 23.6 & 16.0 & 14.5 & 8.2 & - & - & - & - & - \\
        CQD-Beam & 22.3 & - & \bf{46.7} & 11.6 & 8.0 & 31.2 & 40.6 & 21.2 & 18.7 & 14.6 & 8.4 & - & - & - & - & - \\
        FuzzQE & 24.0 & 7.8 & 42.8 & 12.9 & 10.3 & 33.3 & 46.9 & 26.9 & 17.8 & 14.6 & 10.3 & 8.5 & 11.6 & 7.8 & 5.2 & 5.8 \\
        ConE & 23.4 & 5.9 & 41.8 & 12.8 & 11.0 & 32.6 & 47.3 & 25.5 & 14.0 & 14.5 & 10.8 & 5.4 & 8.6 & 7.8 & 4.0 & 3.6 \\
        \midrule
        \method & \bf{26.8} & \bf{10.2} & 42.8 & \bf{14.7} & \bf{11.8} & \bf{38.3} & \bf{54.1} & \bf{31.1} & \bf{18.9} & \bf{16.2} & \bf{13.4} & \bf{10.0} & \bf{16.8} & \bf{9.3} & \bf{7.2} & \bf{7.8} \\
        \midrule[0.08em]
        \multicolumn{17}{c}{NELL995} \\
        \midrule
        GQE & 18.6 & - & 32.8 & 11.9 & 9.6 & 27.5 & 35.2 & 18.4 & 14.4 & 8.5 & 8.8 & - & - & - & - & - \\
        Q2B & 22.9 & - & 42.2 & 14.0 & 11.2 & 33.3 & 44.5 & 22.4 & 16.8 & 11.3 & 10.3 & - & - & - & - & - \\
        BetaE & 24.6 & 5.9 & 53.0 & 13.0 & 11.4 & 37.6 & 47.5 & 24.1 & 14.3 & 12.2 & 8.5 & 5.1 & 7.8 & 10.0 & 3.1 & 3.5 \\
        CQD-CO & \bf{28.8} & - & \bf{60.4} & 17.8 & 12.7 & 39.3 & 46.6 & 30.1 & 22.0 & 17.3 & \bf{13.2} & - & - & - & - & - \\
        CQD-Beam & 28.6 & - & \bf{60.4} & \bf{20.6} & 11.6 & 39.3 & 46.6 & 25.4 & \bf{23.9} & \bf{17.5} & 12.2 & - & - & - & - & - \\
        FuzzQE & 27.0 & 7.8 & 47.4 & 17.2 & 14.6 & 39.5 & 49.2 & 26.2 & 20.6 & 15.3 & 12.6 & 7.8 & 9.8 & 11.1 & 4.9 & 5.5 \\
        ConE & 27.2 & 6.4 & 53.1 & 16.1 & 13.9 & 40.0 & 50.8 & 26.3 & 17.5 & 15.3 & 11.3 & 5.7 & 8.1 & 10.8 & 3.5 & 3.9 \\
        \midrule
        \method & \bf{28.9} & \bf{9.7} & 53.3 & 18.9 & \bf{14.9} & \bf{42.4} & \bf{52.5} & \bf{30.8} & 18.9 & 15.9 & 12.6 & \bf{9.9} & \bf{14.6} & \bf{11.4} & \bf{6.3} & \bf{6.3} \\
        \bottomrule
    \end{tabular}
    \end{adjustbox}
    \label{tab:main}
\end{table*}

\begin{figure*}[!h]
    \centering
    \includegraphics[width=0.74\textwidth]{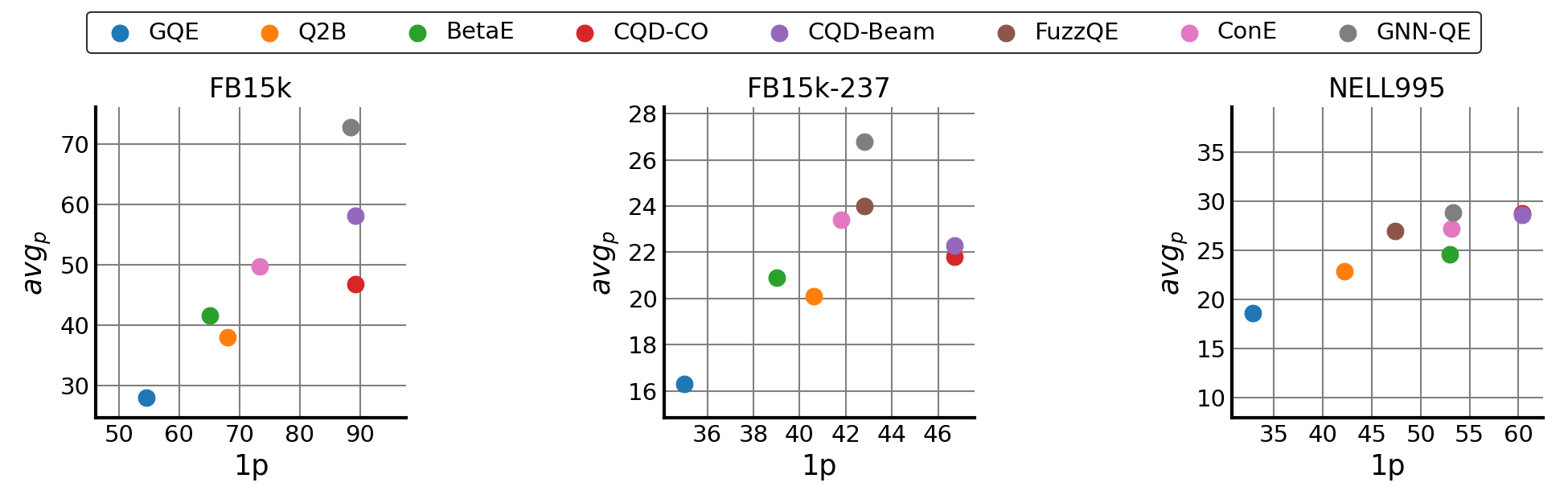}
    \caption{MRR results on EPFO queries w.r.t.\ MRR results on knowledge graph completion (1p queries). Methods on the top left boundary of each plot generalize better from knowledge graph completion to EPFO queries. Best viewed in color.}
    \label{fig:generalization}
\end{figure*}

\subsection{Complex Query Answering}
\label{sec:main_experiment}

Tab.~\ref{tab:main} shows the MRR results of different models for answering FOL queries. GQE, Q2B, CQD-CO and CQD-Beam do not support queries with negation, so the corresponding entries are empty. We observe that \method achieves the best result for both EPFO queries and queries with negation on all 3 datasets. Notably, \method achieves an average relative gain of 22.3\% in avg$_p$ and 95.1\% in avg$_n$ compared to previous best model ConE. We attribute this gain to the advantage of fuzzy sets over geometric embeddings. Fuzzy sets can easily model intermediate variables with many possible assignments, while it is hard to embed a large number of entities in a low-dimensional vector. Such an advantage is especially useful for negation operations, since the output of a negation operation usually contains nearly $|\gV|$ entities.

Intuitively, the performance of complex query models should benefit from better KG completion performance, i.e., \emph{1p} queries. Here we disentangle the contribution of KG completion and complex query framework in answering EPFO queries. Fig.~\ref{fig:generalization} plots the performance of EPFO queries w.r.t.\ the performance of KG completion on all datasets. Methods on the top-left corner of each plot show a better generalization from KG completion to EPFO queries, which implies their complex query frameworks are better. These include GQE, BetaE, FuzzQE, ConE and \method. By contrast, CQD-CO and CQD-Beam generalize worse than other methods, because they rely on a pretrained embedding model and cannot be trained for complex queries.

\subsection{Answer Set Cardinality Prediction}
\label{sec:cardinality}

One advantage of \method is that it can predict the cardinality of the answer set (i.e., the number of answers) without explicit supervision. Specifically, the cardinality of a fuzzy set is computed as the sum of entity probabilities exceeding a certain threshold. We use 0.5 for the threshold as it is a natural choice for our binary classification loss (Eqn.~\ref{eqn:loss}). Tab.~\ref{tab:error} shows the mean absolute percentage error (MAPE) between our model prediction and the ground truth. Note none of existing methods can predict the number of answers without explicit supervision. \citet{ren2020beta} and \citet{zhang2021cone} observe that the uncertainty of Q2B, BetaE and ConE are positively correlated with the number of answers. We follow their setting and report the Spearman's rank correlation between our model prediction and the ground truth. As showed in Tab.~\ref{tab:correlation}, \method outperforms existing methods by a large margin on all query types.

\subsection{Intermediate Variables Visualization}
\label{sec:visualization}

\begin{table*}[!h]
    \centering
    \caption{MAPE (\%) of the number of answers predicted by \method. $avg$ is the average on all query types.}
    \label{tab:error}
    \footnotesize
    \begin{tabular}{lccccccccccccccc}
        \toprule
        \bf{Dataset} & \bf{avg} & \bf{1p} & \bf{2p} & \bf{3p} & \bf{2i} & \bf{3i} & \bf{pi} & \bf{ip} & \bf{2u} & \bf{up} & \bf{2in} & \bf{3in} & \bf{inp} & \bf{pin} & \bf{pni} \\
        \midrule
        FB15k & 37.1 & 34.4 & 29.7 & 34.7 & 39.1 & 57.3 & 47.8 & 34.6 & 13.5 & 26.5 & 31.4 & 50.3 & 50.3 & 39.4 & 29.8 \\
        FB15k-237 & 38.9 & 40.9 & 23.6 & 27.4 & 34.8 & 53.4 & 39.9 & 60.0 & 27.8 & 20.3 & 40.3 & 52.6 & 49.6 & 44.8 & 29.0 \\
        NELL995 & 44.0 & 61.9 & 38.2 & 47.1 & 56.6 & 72.3 & 49.5 & 45.8 & 19.9& 36.2& 30.0 & 47.0 & 42.3 & 39.8 & 29.4 \\
        \bottomrule
    \end{tabular}
\end{table*}

\begin{table*}[!h]
    \centering
    \caption{Spearman's rank correlation between the model prediction and the number of ground truth answers on FB15k-237. $avg$ is the average correlation on all 12 query types in the table. Results of baselines are taken from \cite{zhang2021cone}. Results on FB15k and NELL can be found in Tab.~\ref{tab:correlation_app} in Appendix.}
    \begin{adjustbox}{max width=\textwidth}
        \footnotesize
        \begin{tabular}{lccccccccccccc}
            \toprule
            \bf{Model} & \bf{avg} & \bf{1p} & \bf{2p} & \bf{3p} & \bf{2i} & \bf{3i} & \bf{pi} & \bf{ip} & \bf{2in} & \bf{3in} & \bf{inp} & \bf{pin} & \bf{pni} \\
            \midrule
            Q2B & - & 0.184 & 0.226 & 0.269 & 0.347 & 0.436 & 0.361 & 0.199 & - & - & - & - & - \\
            BetaE & 0.540 & 0.396 & 0.503 & 0.569 & 0.598 & 0.516 & 0.540 & 0.439 & 0.685 & 0.579 & 0.511 & 0.468 & 0.671 \\
            ConE & 0.738 & 0.70 & 0.71 & 0.74 & 0.82 & 0.72 & 0.70 & 0.62 & 0.90 & 0.83 & 0.66 & 0.57 & 0.88 \\
            \midrule
            \method & \bf{0.940} & \bf{0.948} & \bf{0.951} & \bf{0.895} & \bf{0.992} & \bf{0.970} & \bf{0.911} & \bf{0.937} & \bf{0.981} & \bf{0.968} & \bf{0.864} & \bf{0.880} & \bf{0.987} \\
            \bottomrule
        \end{tabular}
    \end{adjustbox}
    \label{tab:correlation}
\end{table*}

\begin{table*}[!h]
    \centering
    \caption{Visualization of a \emph{3p} query from FB15k-237 test set. More visualizations can be found in App.~\ref{app:visualization}.}
    \begin{minipage}{0.17\textwidth}
        \centering
        \includegraphics[width=\textwidth]{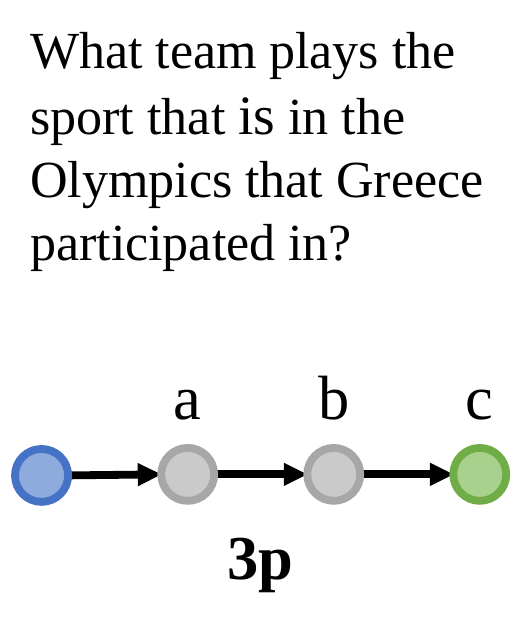}
    \end{minipage}
    \hspace{0.02\textwidth}
    \begin{minipage}{0.8\textwidth}
        \begin{adjustbox}{max width=\textwidth}
            \scriptsize
            \begin{tabular}{lccccc}
                \toprule
                \bf{Query} & \multicolumn{5}{l}{$q = ?c : \exists a, b: \text{ParticipateCountry}(a, \text{Greece}) \land \text{OlympicSports}(a, b) \land \text{TeamSports}(c, b)$} \\
                \midrule
                \multirow{2}{*}{\bf{Variable}} & \multicolumn{3}{c}{\bf{Top Predictions ($\geq 0.1$)}} & \bf{Random} & \bf{Filtered} \\
                & \bf{Easy} & \multicolumn{2}{c}{\bf{Hard}} & \bf{Ground Truth} & \bf{Ranking} \\
                \midrule
                a & \stack{\easy{1936 Summer Olympics} \\ \easy{1980 Winter Olympics} \\ \easy{2002 Winter Olympics}} & \stack{\tp{2010 Winter Olympics} \\ \fp{2012 Summer Olympics} \\ \fp{1920 Summer Olympics}} & \stack{\fp{1988 Summer Olympics} \\ \fp{1928 Summer Olympics} \\ \fp{1992 Summer Olympics}} & \stack{2010 Winter Olympics \\ (hard)} & 1 \\
                \midrule
                b & \stack{\easy{soccer} \\ \easy{track and field} \\ \easy{water polo}} & \stack{\tp{luge} \\ \tp{ice hockey} \\ \tp{short track speed skating}} & \stack{\fp{tennis} \\ - \\ - } & \stack{ice hockey \\ (hard)} & 1 \\
                \midrule
                c & \stack{\easy{Sacramento Kings} \\ \easy{Utah Jazz} \\ \easy{Seattle SuperSonics}} & \stack{\tp{Algeria soccer team} \\ \tp{Cincinnati Reds} \\ \tp{Washington Nationals}} & \stack{\tp{Chile soccer team} \\ \tp{Cardiff City} \\ \tp{Blackburn Rovers}} & \stack{Florida Panthers \\ (hard)} & 433 \\
                \bottomrule
            \end{tabular}
        \end{adjustbox}
    \end{minipage}
    \label{tab:visualization}
\end{table*}

\begin{figure*}[!h]
    \centering
    \begin{minipage}{0.48\textwidth}
        \centering
        \includegraphics[width=0.52\textwidth]{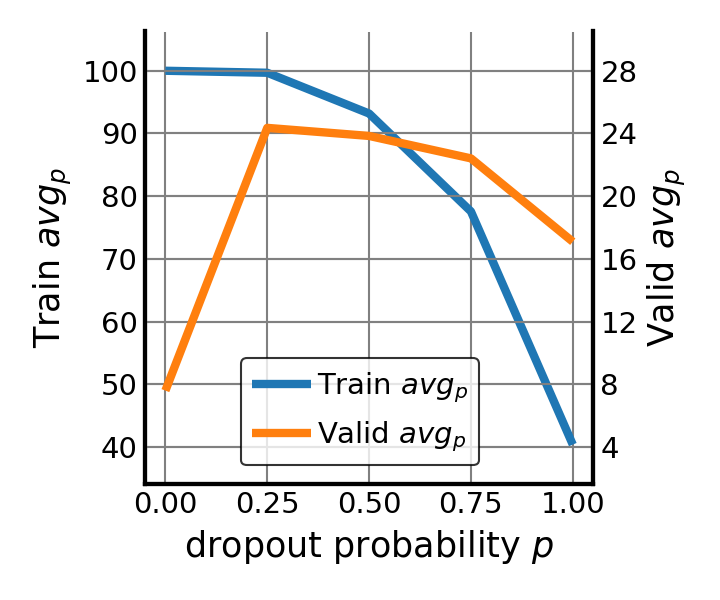}
        \caption{Average MRR on EPFO queries (\%) of train / validation sets w.r.t. traversal dropout probability $p$. The best validation performance is achieved with $p = 0.25$.}
        \label{fig:dropout}
    \end{minipage}
    \hfill
    \begin{minipage}{0.48\textwidth}
        \centering
        \includegraphics[width=0.65\textwidth]{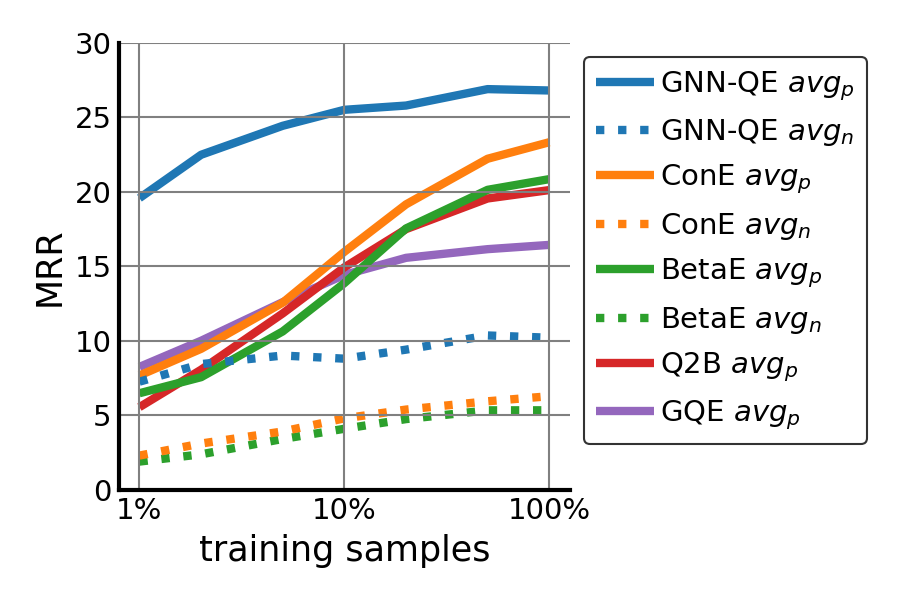}
        \caption{Test MRR results w.r.t. number of training samples. \method is not only better than embeddings, but also less sensitive to the number of training samples.}
        \label{fig:few_sample}
    \end{minipage}
\end{figure*}

Another advantage of \method is that we can interpret its reasoning process by investigating the intermediate variables. As the intermediate fuzzy sets may contain hundreds of entities, we consider two kinds of visualization to qualitatively analyze the precision and the recall of our model. The first one examines the entities with the top probabilities in each fuzzy set, and checks if they are an easy entity (i.e., those can be traversed on the training graph), a hard entity (i.e., those require reasoning) or a false positive one. For each fuzzy set, we visualize the top-3 easy entities and top-6 hard entities that have a minimum probability of 0.1. The second one draws a random ground truth assignment for each variable, such that the assignments form a valid grounding of the query and lead to a hard answer. We report the filtered ranking for each entity in the grounding.

Tab.~\ref{tab:visualization} shows the visualization of \method on a \emph{3p} query from FB15k-237 test set. Among the top hard entities, \method correctly predicts most of the intermediate entities, which indicates our method has a good precision for this sample. For the random ground truth assignments, \method recalls the first two hops (2010 Winter Olympics \& ice hockey) perfectly, but fails for the last hop. Such analysis would be beneficial to identify the steps where error occurs. 

\subsection{Ablation Study}
\label{sec:ablation}

To provide a more comprehensive understanding of \method, we conduct three ablation studies on FB15k-237.

\textbf{Traversal Dropout Probability $p$.} Fig.~\ref{fig:dropout} shows the average MRR on EPFO queries of train and validation sets w.r.t.\ different probability $p$. The model can achieve a perfect training MRR of 1 when $p = 0$, which suggests that the model is able to learn the behavior of a relation traversal model. However, a relation traversal model cannot solve queries on incomplete graphs, which is revealed by its low performance on the validation set. With a non-zero probability $p$, traversal dropout makes the training problem more difficult, and enforces the model to learn a reasoning model that predicts the dropped link from its surrounding graph structure. However, it is not optimal to learn a fully reasoning model with $p = 1$, since it cannot perform relation traversal and some links in the validation queries can be perfectly solved by a relation traversal model.

\textbf{Performance w.r.t.\ Number of Training Samples.}
Fig.~\ref{fig:few_sample} plots the MRR curves of different query types in \method and BetaE under different number of training samples. It is observed that the performance of \method is not only better than BetaE, but also less sensitive to the number of training samples. Even with 1\% training samples (i.e., only 8,233 training queries for FB15k-237), \method achieves a comparative $avg_p$ and better $avg_n$ compared with BetaE trained with the full dataset. We conjecture the reason is that BetaE needs to learn a separate embedding for each entity, while our neural-symbolic method only learns relation embeddings (Eqn.~\ref{eqn:message}) for relation projection, which requires less samples to converge.

\textbf{GNN Parameterization.}
Tab.~\ref{tab:gnn} shows the MRR results of \method w.r.t.\ different GNN parameterizations. We consider three parameterizations for the \textsc{Message} and \textsc{Aggregate} functions in Eqn.~\ref{eqn:bellman}, namely RGCN~\cite{schlichtkrull2018modeling}, CompGCN~\cite{vashishth2019composition} and NBFNet~\cite{zhu2021neural}. It is observed that all three parameterizations outperform BetaE with significant improvement on $avg_n$, which suggests the advantages of fuzzy sets in modeling negation queries. Besides, \method benefits from stronger GNN models (NBFNet $>$ CompGCN $>$ RGCN). The performance of \method might be further improved with better GNN models.
\begin{table}[!h]
    \centering
    \caption{Test MRR results (\%) w.r.t.\ GNN models. \method benefits from better GNN models.}
    \footnotesize
    \begin{tabular}{lcc}
        \toprule
        \bf{Model} & \bf{avg$_p$} & \bf{avg$_n$} \\
        \midrule
        BetaE   & 20.9 & 5.5 \\
        \midrule
        \method (RGCN)    & 20.9 & 7.3 \\
        \method (CompGCN) & 22.5 & 7.3 \\
        \method (NBFNet)  & \bf{26.8} & \bf{10.2} \\
        \bottomrule
    \end{tabular}
    \label{tab:gnn}
\end{table}
\section{Conclusion}

In this paper, we present a novel neural-symbolic model, namely Graph Neural Network Query Executor (\method), for answering complex FOL queries on incomplete knowledge graphs. Our method decomposes complex queries into an expression of basic operations over fuzzy sets, and executes the expression with a learned GNN relation projection model and fuzzy logic operations. \method not only significantly outperforms previous state-of-the-art models on 3 datasets, but also provides interpretability for intermediate variables. Besides, \method can predict the number of answers without explicit supervision. Future works include combining \method with a parser to answer logical queries in the natural language form, and scaling up \method to large-scale knowledge graphs with millions of entities.
\section*{Acknowledgements}

This project is supported by the Natural Sciences and Engineering Research Council (NSERC) Discovery Grant, the Canada CIFAR AI Chair Program, collaboration grants between Microsoft Research and Mila, Samsung Electronics Co., Ltd., Amazon Faculty Research Award, Tencent AI Lab Rhino-Bird Gift Fund and a NRC Collaborative R\&D Project (AI4D-CORE-06). This project was also partially funded by IVADO Fundamental Research Project grant PRF-2019-3583139727. The computation resource of this project is supported by Calcul Qu\'ebec\footnote{\url{https://www.calculquebec.ca/}} and Compute Canada\footnote{\url{https://www.computecanada.ca/}}.

We would like to thank Meng Qu for discussion on research ideas, and Pasquale Minervini for discussion on the CQD model. We also appreciate all anonymous reviewers for their constructive suggestions on this paper.

\bibliography{reference}
\bibliographystyle{icml2022}

\newpage
\appendix
\onecolumn

\section{Dataset Statistics}
\label{app:datasets}

\begin{figure*}[!h]
    \centering
    \includegraphics[width=0.98\textwidth]{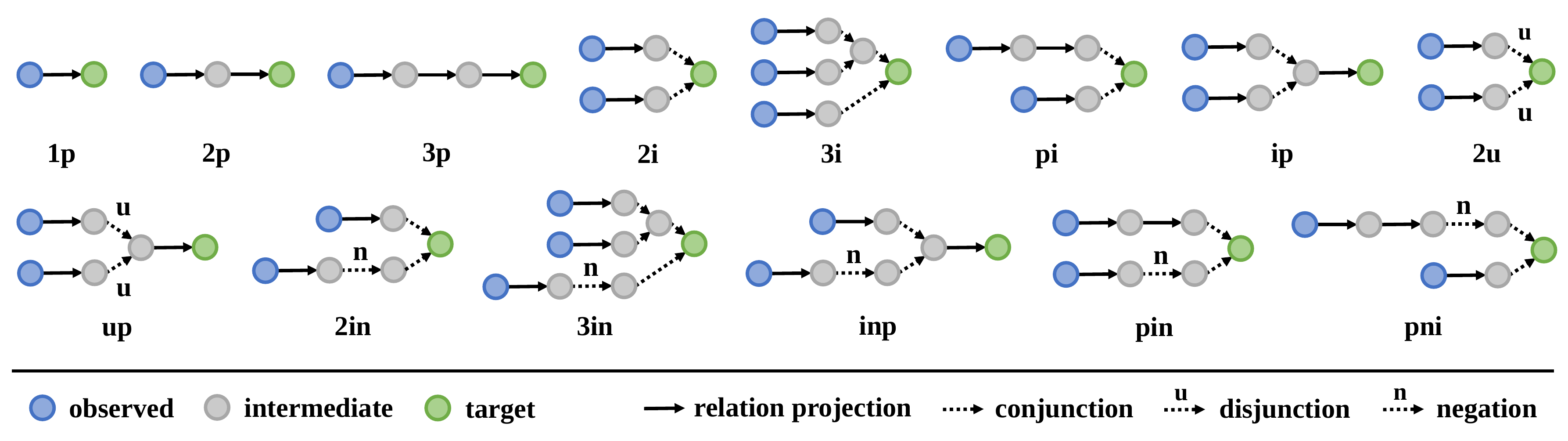}
    \caption{Types of complex FOL queries used in training and inference.}
    \label{fig:query_type}
\end{figure*}

We use the complex query datasets generated by \cite{ren2020beta}. There is a total number of 14 query types, as showed in Fig.~\ref{fig:query_type}. Statistics of all query types is summarized in Tab.~\ref{tab:statistics}.

\begin{table}[!h]
    \centering
    \caption{Statistics of different query types used in the benchmark datasets.}
    \begin{adjustbox}{max width=0.48\textwidth}
        \begin{tabular}{llcccc}
            \toprule
            \bf{Split} & \bf{Query Type} & \bf{FB15k} & \bf{FB15k-237} & \bf{NELL995} \\
            \midrule
            \multirow{2}{*}{Train}
            & 1p/2p/3p/2i/3i & 273,710 & 149,689 & 107,982 \\
            & 2in/3in/inp/pin/pni & 27,371 & 14,968 & 10,798 \\
            \midrule
            \multirow{2}{*}{Valid}
            & 1p & 59,078 & 20,094 & 16,910 \\
            & Others & 8,000 & 5,000 & 4,000 \\
            \midrule
            \multirow{2}{*}{Test}
            & 1p & 66,990 & 22,804 & 17,021 \\
            & Others & 8,000 & 5,000 & 4,000 \\
            \bottomrule
        \end{tabular}
    \end{adjustbox}
    \label{tab:statistics}
\end{table}

\section{Hyperparameters}
\label{app:hyperparameter}

Tab.~\ref{tab:hyperparameter} lists the hyperparameter configurations of \method on different datasets.

\begin{table*}[!h]
    \centering
    \caption{Hyperparameters of \method on different datasets. All the hyperparameters are selected by the performance on the validation set.}
    \footnotesize
    \begin{tabular}{llccc}
        \toprule
        \multicolumn{2}{l}{\bf{Hyperparameter}} & \bf{FB15k} & \bf{FB15k-237} & \bf{NELL-995} \\
        \midrule
        \multirow{2}{*}{\bf{GNN}}
        & \#layer & 4 & 4 & 4 \\
        & hidden dim. & 32 & 32 & 32 \\
        \midrule
        \multirow{2}{*}{\bf{MLP}}
        & \#layer & 2 & 2 & 2 \\
        & hidden dim. & 64 & 64 & 64 \\
        \midrule
        \bf{Traversal Dropout} & probability & 0.25 & 0.25 & 0.25 \\
        \midrule
        \multirow{6}{*}{\bf{Learning}}
        & batch size & 192 & 192 & 32 \\
        & sample weight & uniform across queries & uniform across queries & uniform across answers \\
        & optimizer        & Adam & Adam   & Adam \\
        & learning rate    & 5e-3 & 5e-3   & 5e-3 \\
        & iterations (\#batch) & 10,000 & 10,000 & 30,000 \\
        & adv. temperature & 0.2 & 0.2 & 0.2 \\
        \bottomrule
    \end{tabular}
    \label{tab:hyperparameter}
\end{table*}

\section{Batched Expression Execution}
\label{app:code}

Alg.~\ref{alg:postfix} shows the pseudo code for converting expression to postfix notation. The idea is to recursively parse the expression from outside to inside, and construct the postfix notation from inside to outside. We preprocess all query samples in training and evaluation with Alg.~\ref{alg:postfix}.

\begin{figure*}[!h]
    \centering
    \includegraphics[width=0.9\textwidth]{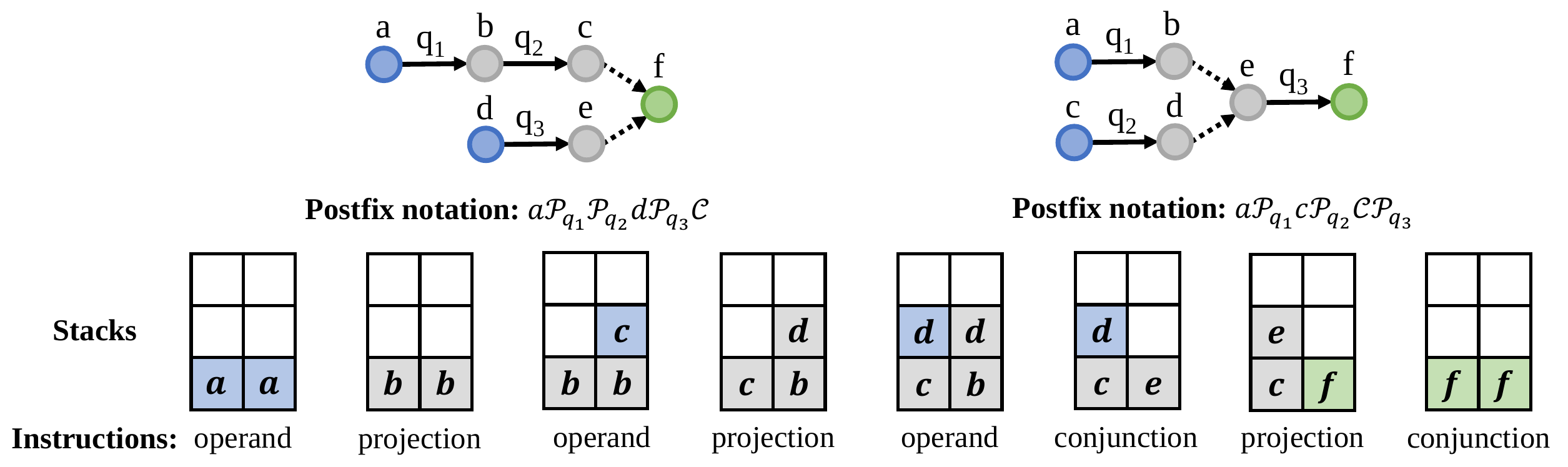}
    \caption{Illustration of batched expression execution (Alg.~\ref{alg:batch_execution}) over a batch of two queries.}
    \label{fig:batch_execution}
\end{figure*}

Alg.~\ref{alg:batch_execution} illustrates the steps of batch execution over postfix expressions. For clarity, we describe the algorithm as one for loop over samples in the pseudo code, while samples that fall into the same case (Line 8, 10, 13, 16 \& 19) are executed in parallel. Since the GNN in relation projection takes ($O(|\gV|d^2 + |\gE|d)$ time (see App. C of \cite{zhu2021neural} for proofs), i.e., much more time than fuzzy logic operations ($O(|\gV|)$ time), we synchronize different samples before neural relation projection (Line 20) to maximize the utilization of GPU. Fig.~\ref{fig:batch_execution} shows the procedure of Alg.~\ref{alg:batch_execution} over a batch of two queries.

The overall time complexity of our batched execution is $O(t(|\gV|d^2 + |\gE|d))$, where $t$ is the maximal number of projections in a single query in the batch. Compared to existing implementation~\cite{hamilton2018embedding, ren2019query2box, ren2020beta} that scales \emph{linearly} w.r.t.\ the number of query types, batched expression execution scales \emph{independently} w.r.t.\ the number of query types, and can be applied to arbitrary large number of query types without scalability issues.

\begin{minipage}[!h]{\textwidth}
\begin{minipage}[t]{0.49\textwidth}
    \begin{algorithm}[H]
        \footnotesize
        \captionsetup{font=footnotesize}\caption{Convert expression to postfix notation}
        \begin{algorithmic}[1]
            \STATE {\bfseries Input:} a query expression
            \STATE {\bfseries Output:} postfix notation of the query expression
            \FUNCTION{GetPostfix($exp$)}
                \STATE $\textit{postfix} \gets []$
                \STATE $\textit{op}, \textit{exps\_or\_vars} \gets \text{GetOutmostOperation}(\textit{exp})$
                \FOR{$\textit{exp\_or\_var}$ in $\textit{exps\_or\_vars}$}
                    \IF{$\textit{exp\_or\_var}$ is an expression}
                        \STATE $\textit{postfix} \gets \textit{postfix} + \text{GetPostfix}(\textit{exp\_or\_var})$
                    \ELSE
                        \STATE $\textit{postfix} \gets \textit{postfix} + \textit{exp\_or\_var}$
                    \ENDIF
                \ENDFOR
                \STATE {\bfseries return} $\textit{postfix} + \textit{op}$
            \ENDFUNCTION
        \end{algorithmic}
        \label{alg:postfix}
    \end{algorithm}
    \vfill
\end{minipage}
\hfill
\begin{minipage}[t]{0.49\textwidth}
    \begin{algorithm}[H]
        \footnotesize
        \captionsetup{font=footnotesize}\caption{Batched expression execution}
        \begin{algorithmic}[1]
            \STATE {\bfseries Input:} a batch of expressions in postfix notation
            \STATE {\bfseries Output:} a batch of fuzzy sets for answers
            \STATE $\textit{stacks} \gets$ allocate $\textit{batch\_size}$ stacks for fuzzy sets
            \FOR {$i \gets 0$ to $\textit{batch\_size} - 1$}
                \STATE // parallelized loop
                \FOR {$\textit{instruction}$ in $\textit{queries[i]}$}    
                    \STATE {\bfseries switch} $\textit{instruction}$ {\bfseries do}
                        \INDSTATE {\bfseries case} operand
                            \INDSTATE[2] \textit{stacks}[i].push($\textit{instruction}$)
                        \INDSTATE {\bfseries case} conjunction
                            \INDSTATE[2] $\bm{x}, \bm{y} \gets$ \textit{stacks}[i].pop(), \textit{stacks}[i].pop()
                            \INDSTATE[2] \textit{stacks}[i].push($\gC(\bm{x}, \bm{y})$)
                        \INDSTATE {\bfseries case} disjunction
                            \INDSTATE[2] $\bm{x}, \bm{y} \gets$ \textit{stacks}[i].pop(), \textit{stacks}[i].pop()
                            \INDSTATE[2] \textit{stacks}[i].push($\gD(\bm{x}, \bm{y})$)
                        \INDSTATE {\bfseries case} negation
                            \INDSTATE[2] $\bm{x} \gets$ \textit{stacks}[i].pop()
                            \INDSTATE[2] \textit{stacks}[i].push($\gN(\bm{x})$)
                        \INDSTATE {\bfseries case} projection
                            \INDSTATE[2] wait until all samples are in this case
                            \INDSTATE[2] $\bm{x} \gets$ \textit{stacks}[i].pop()
                            \INDSTATE[2] \textit{relation} $\gets$ instruction.relation
                            \INDSTATE[2] \textit{stacks}[i].push($\gP_\textit{relation}(\bm{x})$)
                    \STATE {\bfseries end switch}
                \ENDFOR
            \ENDFOR
            \STATE {\bfseries return} \textit{stacks}.pop()
        \end{algorithmic}
        \label{alg:batch_execution}
    \end{algorithm}
\end{minipage}
\end{minipage}

\section{More Experiment Results}
\label{app:result}

Here we provide additional experiment results.

Tab.~\ref{tab:main_app} shows the H@1 results of different models for answering FOL queries. \method significantly outperforms existing methods in both EPFO queries and negation queries on all datasets.

Tab.~\ref{tab:correlation_app} compares the Spearman's rank correlation for answer set cardinality prediction on FB15k and NELL-995. \method achieves the best rank correlation on all query types.

\begin{table*}[!h]
    \centering
    \caption{Test H@1 results (\%) on answering FOL queries. avg$_p$ is the average H@1 on EPFO queries ($\land$, $\lor$). avg$_n$ is the average H@1 on queries with negation. Results of GQE and Q2B are taken from \cite{ren2020beta}.}
    \begin{adjustbox}{width=\textwidth}
    \begin{tabular}{lcccccccccccccccc}
        \toprule
        \bf{Model} & \bf{avg$_p$} & \bf{avg$_n$} & \bf{1p} & \bf{2p} & \bf{3p} & \bf{2i} & \bf{3i} & \bf{pi} & \bf{ip} & \bf{2u} & \bf{up} & \bf{2in} & \bf{3in} & \bf{inp} & \bf{pin} & \bf{pni} \\
        \midrule
        \multicolumn{17}{c}{FB15k} \\
        \midrule
        GQE & 16.6 & - & 34.2 & 8.3 & 5.0 & 23.8 & 34.9 & 15.5 & 11.2 & 11.5 & 5.6 & - & - & - & - & - \\
        Q2B & 26.8 & - & 52.0 & 12.7 & 7.8 & 40.5 & 53.4 & 26.7 & 16.7 & 22.0 & 9.4 & - & - & - & - & - \\
        BetaE & 31.3 & 5.2 & 52.0 & 17.0 & 16.9 & 43.5 & 55.3 & 32.3 & 19.3 & 28.1 & 16.9 & 6.4 & 6.7 & 5.5 & 2.0 & 5.3 \\
        CQD-CO & 39.7 & - & 85.8 & 17.8 & 9.0 & 67.6 & 71.7 & 34.5 & 24.5 & 30.9 & 15.5 & - & - & - & - & - \\
        CQD-Beam & 51.9 & - & 85.8 & 48.6 & 22.5 & 67.6 & 71.7 & 51.7 & 62.3 & 31.7 & 25.0 & - & - & - & - & - \\
        ConE & 39.6 & 7.3 & 62.4 & 23.8 & 20.4 & 53.6 & 64.1 & 39.6 & 25.6 & 44.9 & 21.7 & 9.4 & 9.1 & 6.0 & 4.3 & 7.5 \\
        \midrule
        \method & \bf{67.3} & \bf{28.6} & \bf{86.1} & \bf{63.5} & \bf{52.5} & \bf{74.8} & \bf{80.1} & \bf{63.6} & \bf{65.1} & \bf{67.1} & \bf{53.0} & \bf{35.4} & \bf{33.1} & \bf{33.8} & \bf{18.6} & \bf{21.8} \\
        \midrule[0.08em]
        \multicolumn{17}{c}{FB15k-237} \\
        \midrule
        GQE & 8.8 & - & 22.4 & 2.8 & 2.1 & 11.7 & 20.9 & 8.4 & 5.7 & 3.3 & 2.1 & - & - & - & - & - \\
        Q2B & 12.3 & - & 28.3 & 4.1 & 3.0 & 17.5 & 29.5 & 12.3 & 7.1 & 5.2 & 3.3 & - & - & - & - & - \\
        BetaE & 13.4 & 2.8 & 28.9 & 5.5 & 4.9 & 18.3 & 31.7 & 14.0 & 6.7 & 6.3 & 4.6 & 1.5 & 7.7 & 3.0 & 0.9 & 0.9 \\
        CQD-CO & 14.7 & - & \bf{36.6} & 4.7 & 3.0 & 20.7 & 29.6 & 15.5 & 9.9 & 8.6 & 4.0 & - & - & - & - & - \\
        CQD-Beam & 15.1 & - & \bf{36.6} & 6.3 & 4.3 & 20.7 & 29.6 & 13.5 & 12.1 & 8.7 & 4.3 & - & - & - & - & - \\
        ConE & 15.6 & 2.2 & 31.9 & 6.9 & 5.3 & 21.9 & 36.6 & 17.0 & 7.8 & 8.0 & 5.3 & 1.8 & 3.7 & 3.4 & 1.3 & 1.0 \\
        \midrule
        \method & \bf{19.1} & \bf{4.3} & 32.8 & \bf{8.2} & \bf{6.5} & \bf{27.7} & \bf{44.6} & \bf{22.4} & \bf{12.3} & \bf{9.8} & \bf{7.6} & \bf{4.1} & \bf{8.1} & \bf{4.1} & \bf{2.5} & \bf{2.7} \\
        \midrule[0.08em]
        \multicolumn{17}{c}{NELL-995} \\
        \midrule
        GQE & 9.9 & - & 15.4 & 6.7 & 5.0 & 14.3 & 20.4 & 10.6 & 9.0 & 2.9 & 5.0 & - & - & - & - & - \\
        Q2B & 14.1 & - & 23.8 & 8.7 & 6.9 & 20.3 & 31.5 & 14.3 & 10.7 & 5.0 & 6.0 & - & - & - & - & - \\
        BetaE & 17.8 & 2.1 & 43.5 & 8.1 & 7.0 & 27.2 & 36.5 & 17.4 & 9.3 & 6.9 & 4.7 & 1.6 & 2.2 & 4.8 & 0.7 & 1.2 \\
        CQD-CO & 21.3 & - & \bf{51.2} & 11.8 & 9.0 & 28.4 & 36.3 & 22.4 & 15.5 & 9.9 & \bf{7.6} & - & - & - & - & - \\
        CQD-Beam & 21.0 & - & \bf{51.2} & \bf{14.3} & 6.3 & 28.4 & 36.3 & 18.1 & \bf{17.4} & \bf{10.2} & 7.2 & - & - & - & - & - \\
        ConE & 19.8 & 2.2 & 43.6 & 10.7 & 9.0 & 28.6 & 39.8 & 19.2 & 11.4 & 9.0 & 6.6 & 1.4 & 2.6 & 5.2 & 0.8 & 1.2 \\
        \midrule
        \method & \bf{21.5} & \bf{3.6} & 43.5 & 12.9 & \bf{9.9} & \bf{32.5} & \bf{42.4} & \bf{23.5} & 12.9 & 8.8 & 7.4 & \bf{3.2} & \bf{5.9} & \bf{5.4} & \bf{1.6} & \bf{2.0} \\
        \bottomrule
    \end{tabular}
    \end{adjustbox}
    \label{tab:main_app}
\end{table*}

\begin{table*}[!h]
    \centering
    \caption{Spearman's rank correlation between the model prediction and the number of ground truth answers. $avg$ is the average correlation on all 12 query types in the table. Results of baseline methods are taken from \cite{zhang2021cone}.}
    \begin{adjustbox}{max width=\textwidth}
        \footnotesize
        \begin{tabular}{lccccccccccccc}
            \toprule
            \bf{Model} & \bf{avg} & \bf{1p} & \bf{2p} & \bf{3p} & \bf{2i} & \bf{3i} & \bf{pi} & \bf{ip} & \bf{2in} & \bf{3in} & \bf{inp} & \bf{pin} & \bf{pni} \\
            \midrule
            \multicolumn{13}{c}{FB15k} \\
            \midrule
            Q2B & - & 0.301 & 0.219 & 0.262 & 0.331 & 0.270 & 0.297 & 0.139 & - & - & - & - & - \\
            BetaE & 0.494 & 0.373 & 0.478 & 0.472 & 0.572 & 0.397 & 0.519 & 0.421 & 0.622 & 0.548 & 0.459 & 0.465 & 0.608 \\
            ConE & 0.659 & 0.60 & 0.68 & 0.70 & 0.68 & 0.52 & 0.59 & 0.56 & 0.84 & 0.75 & 0.61 & 0.58 & 0.80 \\
            \midrule
            \method & \bf{0.945} & \bf{0.958} & \bf{0.970} & \bf{0.940} & \bf{0.984} & \bf{0.927} & \bf{0.936} & \bf{0.916} & \bf{0.980} & \bf{0.907} & \bf{0.905} & \bf{0.944} & \bf{0.978} \\
            \midrule[0.08em]
            \multicolumn{13}{c}{NELL995} \\
            \midrule
            Q2B & - & 0.154 & 0.288 & 0.305 & 0.380 & 0.410 & 0.361 & 0.345 & - & - & - & - & - \\
            BetaE & 0.552 & 0.423 & 0.552 & 0.564 & 0.594 & 0.610 & 0.598 & 0.535 & 0.711 & 0.595 & 0.354 & 0.447 & 0.639 \\
            ConE & 0.688 & 0.56 & 0.61 & 0.60 & 0.79 & 0.79 & 0.74 & 0.58 & 0.90 & 0.79 & 0.56 & 0.48 & 0.85 \\
            \midrule
            \method & \bf{0.891} & \bf{0.913} & \bf{0.851} & \bf{0.780} & \bf{0.974} & \bf{0.935} & \bf{0.825} & \bf{0.737} & \bf{0.994} & \bf{0.980} & \bf{0.882} & \bf{0.848} & \bf{0.976} \\
            \bottomrule
        \end{tabular}
    \end{adjustbox}
    \label{tab:correlation_app}
\end{table*}

\section{More Visualization Results}
\label{app:visualization}

We provide more visualization for intermediate variables in Tab.~\ref{tab:visualization_app}. For each of the 14 query types, we \emph{randomly} draw 3 query samples from the test set of FB15k-237. Therefore, we can observe both successful and failure cases of our method. For all expressions in the visualization, the operations follow the priority $\pmb{\neg} > \pmb{\land} > \pmb{\lor}$.

Note the correctness of some predictions are contradictory to common sense. This is not a failure of our visualization, but is a result of the incomplete knowledge graph. For these contradictory tables, we add a footnote below to illustrate this problem.

\begin{table*}[!h]
    \centering
    \caption{Visualization of random samples from FB15k-237 test set.}
    \begin{minipage}{0.17\textwidth}
        \includegraphics[width=\textwidth]{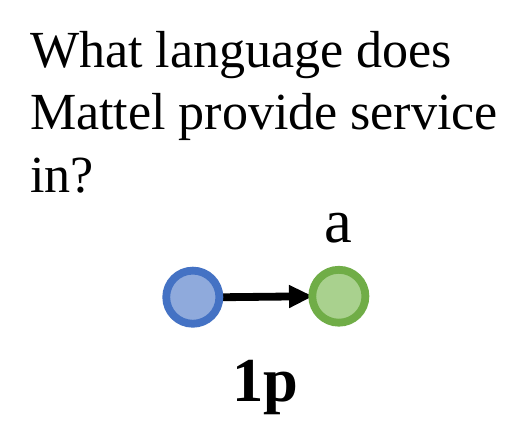}
    \end{minipage}
    \hspace{0.02\textwidth}
    \begin{minipage}{0.8\textwidth}
        \begin{adjustbox}{max width=\textwidth}
            \scriptsize
\begin{tabular}{lccccc}
                \toprule
                \bf{Query} & \multicolumn{5}{l}{$q = ?a: \text{ServiceLanguage}(\text{Mattel}, a)$} \\
                \midrule
                \multirow{2}{*}{\bf{Variable}} & \multicolumn{3}{c}{\bf{Top Predictions ($\geq 0.1$)}} & \bf{Random} & \bf{Filtered} \\
                & \bf{Easy} & \multicolumn{2}{c}{\bf{Hard}} & \bf{Ground Truth} & \bf{Ranking} \\
                \midrule
                a & \stack{\easy{English} \\ \easy{Spanish} \\ -} & \stack{\fp{Chinese} \\ \fp{Russia} \\ \tp{French}} & \stack{\fp{Arabic} \\ - \\ -} & \stack{French \\ (hard)} & 3 \\
                \bottomrule
            \end{tabular}
        \end{adjustbox}
    \end{minipage}
    \vspace{-0.5em}
    \label{tab:visualization_app}
\end{table*}
\begin{table*}[!h]
    \begin{minipage}{0.17\textwidth}
        \includegraphics[width=\textwidth]{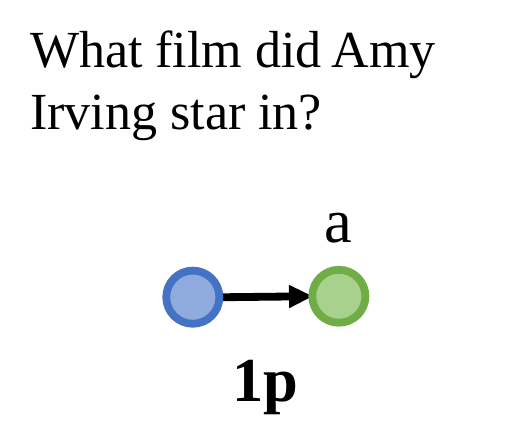}
    \end{minipage}
    \hspace{0.02\textwidth}
    \begin{minipage}{0.8\textwidth}
        \begin{adjustbox}{max width=\textwidth}
            \scriptsize
            \begin{tabular}{lccccc}
                \toprule
                \bf{Query} & \multicolumn{5}{l}{$q = ?a: \text{Star}(\text{Amy Irving}, a)$} \\
                \midrule
                \multirow{2}{*}{\bf{Variable}} & \multicolumn{3}{c}{\bf{Top Predictions ($\geq 0.1$)}} & \bf{Random} & \bf{Filtered} \\
                & \bf{Easy} & \multicolumn{2}{c}{\bf{Hard}} & \bf{Ground Truth} & \bf{Ranking} \\
                \midrule
                a & \stack{\easy{Never Say Never Again} \\ \easy{Carrie} \\ \easy{Who Framed Roger Rabbit}} & \stack{\fp{She's Having a Baby} \\ \fp{Mickey's Magical Christmas} \\ \tp{Terror in the Aisles}} & \stack{\fp{Think like a Man} \\ \fp{Evolution} \\ \fp{I Love You, Man}} & \stack{Terror in the Aisles \\ (hard)} & 3 \\
                \bottomrule
            \end{tabular}
        \end{adjustbox}
    \end{minipage}
    \vspace{-0.5em}
\end{table*}
\begin{table*}[!h]
    \begin{minipage}{0.17\textwidth}
        \includegraphics[width=\textwidth]{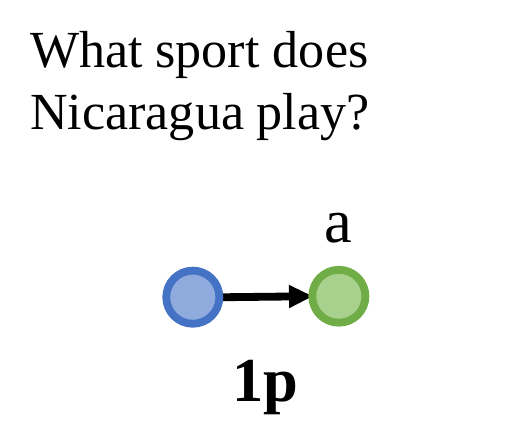}
    \end{minipage}
    \hspace{0.02\textwidth}
    \begin{minipage}{0.8\textwidth}
        \begin{adjustbox}{max width=\textwidth}
            \scriptsize
            \begin{tabular}{lccccc}
                \toprule
                \bf{Query} & \multicolumn{5}{l}{$q = ?a: \text{SportCountry}(a, Nicaragua)$} \\
                \midrule
                \multirow{2}{*}{\bf{Variable}} & \multicolumn{3}{c}{\bf{Top Predictions ($\geq 0.1$)}} & \bf{Random} & \bf{Filtered} \\
                & \bf{Easy} & \multicolumn{2}{c}{\bf{Hard}} & \bf{Ground Truth} & \bf{Ranking} \\
                \midrule
                a & \stack{\easy{beach volleyball} \\ \easy{shooting sport} \\ \easy{rowing}} & \stack{\fp{road running} \\ \tp{track and field} \\ \fp{judo}} & \stack{\fp{taekwondo} \\ \tp{boxing} \\ \fp{road cycling}} & \stack{boxing \\ (hard)} & 4 \\
                \bottomrule
            \end{tabular}
        \end{adjustbox}
    \end{minipage}
    \vspace{-0.5em}
\end{table*}
\begin{table*}[!h]
    \centering
    \begin{minipage}{0.17\textwidth}
        \includegraphics[width=\textwidth]{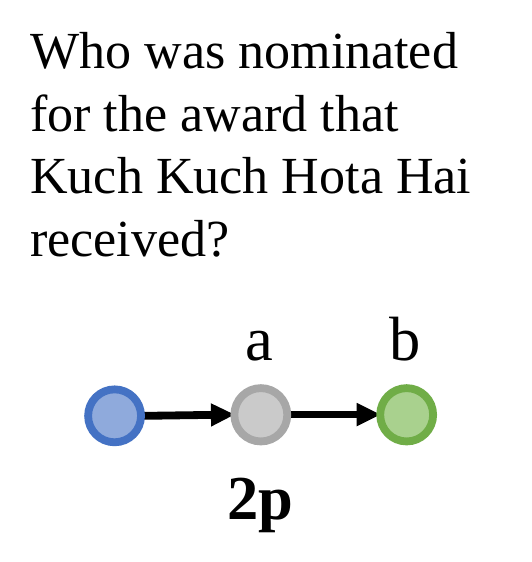}
    \end{minipage}
    \hspace{0.02\textwidth}
    \begin{minipage}{0.8\textwidth}
        \begin{adjustbox}{max width=\textwidth}
            \footnotesize
            \begin{tabular}{lccccc}
                \toprule
                \bf{Query} & \multicolumn{5}{l}{$q = ?b: \text{Award}(\text{Kuch Kuch Hota Hai}, a) \land \text{Nominated}(b, a)$} \\
                \midrule
                \multirow{2}{*}{\bf{Variable}} & \multicolumn{3}{c}{\bf{Top Predictions ($\geq 0.1$)}} & \bf{Random} & \bf{Filtered} \\
                & \bf{Easy} & \multicolumn{2}{c}{\bf{Hard}} & \bf{Ground Truth} & \bf{Ranking} \\
                \midrule
                a & \stack{\easy{Filmfare Award for Best Actor} \\ \easy{Filmfare Award for Best Film} \\ \easy{Filmfare Award for Best Actress}} & \stack{\fp{Filmfare Award for Best Director} \\ - \\ -} & \stack{- \\ - \\ -} & \stack{Filmfare Award for\\ Best Supporting Actor \\ (easy)} & 1 \\
                \midrule
                b & \stack{\easy{Girish Karnad} \\ \easy{Kamal Haasan} \\ \easy{Dilip Kumar}} & \stack{\fp{Vijay Anand} \\ \tp{Aamir Khan} \\ \fp{K. Balachander}} & \stack{\fp{Asit Sen} \\ \fp{Sridevi} \\ \fp{Aishwarya Rai}} & \stack{Arjun Rampal \\ (hard)} & 25 \\
                \bottomrule
            \end{tabular}
        \end{adjustbox}
    \end{minipage}
    \vspace{-0.5em}
\end{table*}
\begin{table*}[!h]
    \begin{minipage}{0.17\textwidth}
        \includegraphics[width=\textwidth]{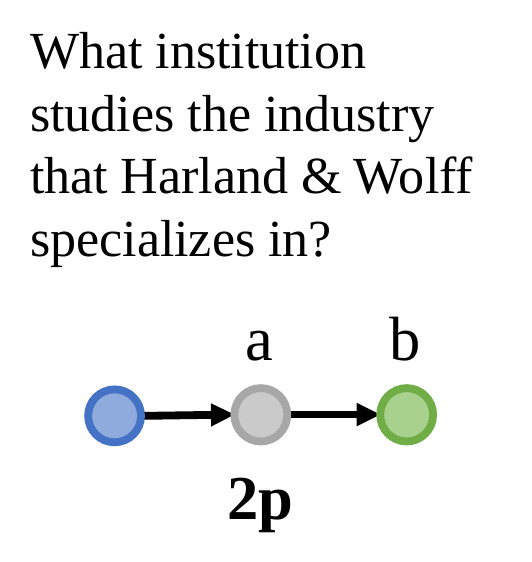}
    \end{minipage}
    \hspace{0.02\textwidth}
    \begin{minipage}{0.8\textwidth}
        \begin{adjustbox}{max width=\textwidth}
            \footnotesize
            \begin{tabular}{lccccc}
                \toprule
                \bf{Query} & \multicolumn{5}{l}{$q = ?b: \text{Industry}(\text{Harland \& Wolff}, a) \land \text{FieldOfStudy}(b, a)$} \\
                \midrule
                \multirow{2}{*}{\bf{Variable}} & \multicolumn{3}{c}{\bf{Top Predictions ($\geq 0.1$)}} & \bf{Random} & \bf{Filtered} \\
                & \bf{Easy} & \multicolumn{2}{c}{\bf{Hard}} & \bf{Ground Truth} & \bf{Ranking} \\
                \midrule
                a & \stack{\easy{civil engineering} \\ \easy{shipbuilding} \\ - } & \stack{\fp{video game} \\ \fp{media} \\ \fp{airline}} & \stack{\fp{publishing} \\ \fp{broadcasting} \\ \fp{entertainment}} & \stack{civil engineering \\ (easy)} & 1 \\
                \midrule
                b & \stack{\easy{West Point} \\ \easy{Yale University} \\ \easy{Newcastle University}} & \stack{\fp{University of Minnesota} \\ \fp{Princeton University} \\ \fp{Brown University}} & \stack{\fp{University of Edinburgh} \\ \fp{George Washington University} \\ \fp{Rutgers University}} & \stack{University of Utah \\ (hard)} & 26 \\
                \bottomrule
            \end{tabular}
        \end{adjustbox}
    \end{minipage}
    \vspace{-0.5em}
\end{table*}
\begin{table*}[!h]
    \begin{minipage}{0.17\textwidth}
        \includegraphics[width=\textwidth]{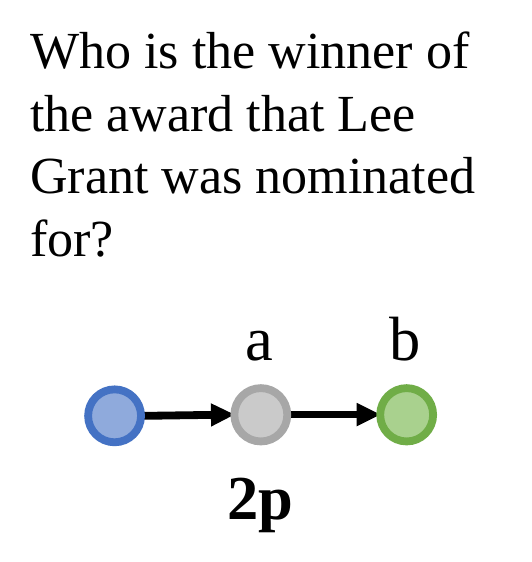}
    \end{minipage}
    \hspace{0.02\textwidth}
    \begin{minipage}{0.8\textwidth}
        \begin{adjustbox}{max width=\textwidth}
            \footnotesize
            \begin{tabular}{lccccc}
                \toprule
                \bf{Query} & \multicolumn{5}{l}{$q = ?b: \text{Nominated}(\text{Lee Grant}, a) \land \text{Winner}(a, b)$} \\
                \midrule
                \multirow{2}{*}{\bf{Variable}} & \multicolumn{3}{c}{\bf{Top Predictions ($\geq 0.1$)}} & \bf{Random} & \bf{Filtered} \\
                & \bf{Easy} & \multicolumn{2}{c}{\bf{Hard}} & \bf{Ground Truth} & \bf{Ranking} \\
                \midrule
                a & \stack{Emmy Award for \\ \easy{Best Supporting Actress (Drama)} \\ Emmy Award for\\ \easy{Best Lead Actress (Comedy)} \\ Academy Award for \\ \easy{Best Supporting Actress}} & \stack{Emmy Award for\\ \fp{Best Lead Actress (Movie)} \\ Cannes Award for\\ \fp{Best Actress} \\ Tony Award for\\ \fp{Best Actress (Play)}} & \stack{Tony Award for\\ \fp{Best Featured Actress (Play)} \\ SAG Award for\\ \fp{Best Female Actor (Movie)} \\ Emmy Award for\\ \fp{Best Guest Actress (Drama)}} & \stack{Academy Award for\\ Best Supporting Actress\\ (easy)} & 1 \\
                \midrule
                b & \stack{\easy{Cate Blanchett} \\ \easy{Anna Paquin} \\ \easy{Juliette Binoche}} & \stack{\fp{Laura Linney} \\ \fp{Jessica Lange} \\ \fp{Sally Field}} & \stack{\fp{Helen Mirren} \\ \fp{Jane Fonda} \\ \fp{Meryl Streep}} & \stack{Angelina Jolie \\ (hard)} & 32 \\
                \bottomrule
            \end{tabular}
        \end{adjustbox}
    \end{minipage}
    \vspace{-0.5em}
\end{table*}
\begin{table*}[!h]
    \centering
    \begin{minipage}{0.17\textwidth}
        \centering
        \includegraphics[width=\textwidth]{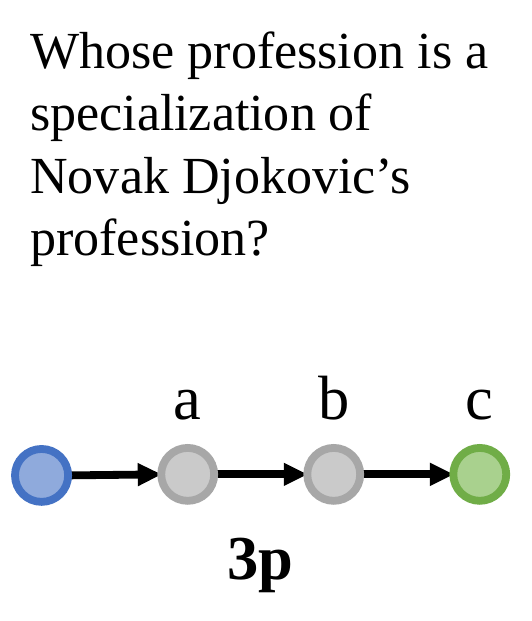}
    \end{minipage}
    \hspace{0.02\textwidth}
    \begin{minipage}{0.8\textwidth}
        \begin{adjustbox}{max width=\textwidth}
            \scriptsize
            \begin{tabular}{lccccc}
                \toprule
                \bf{Query} & \multicolumn{5}{l}{$q = ?c : \exists a, b: \text{Profession}(\text{Novak Djokovic}, a) \land \text{SpecializationOf}(b, a) \land \text{Profession}(c, b)$} \\
                \midrule
                \multirow{2}{*}{\bf{Variable}} & \multicolumn{3}{c}{\bf{Top Predictions ($\geq 0.1$)}} & \bf{Random} & \bf{Filtered} \\
                & \bf{Easy} & \multicolumn{2}{c}{\bf{Hard}} & \bf{Ground Truth} & \bf{Ranking} \\
                \midrule
                a & \stack{\easy{athlete} \\ - \\ -} & \stack{\fp{actor} \\ \fp{screenwriter} \\ \fp{voice actor}} & \stack{\fp{film director} \\ \fp{politician} \\ \fp{soccer player}} & \stack{athlete \\ (easy)} & 1 \\
                \midrule
                b & \stack{\easy{soccer player} \\ \easy{attacker (soccer)} \\ \easy{strongman}} & \stack{- \\ - \\ -} & \stack{- \\ - \\ -} & \stack{soccer player \\ (easy)} & 1 \\
                \midrule
                c & \stack{\easy{Jermaine Easter} \\ \easy{Jonathan Walters} \\ \easy{Taiwo Atieno}} & \stack{\tp{Delroy Facey}  \\ \tp{Kevin Betsy} \\ \fp{Carl Cort}} & \stack{\fp{Jung Sung-ryong} \\ \fp{Leon Cort} \\ \tp{Hameur Bouazza}} & \stack{Roy Carroll \\ (hard)} & 4 \\
                \bottomrule
            \end{tabular}
        \end{adjustbox}
    \end{minipage}
    \vspace{-0.5em}
\end{table*}
\begin{table*}[!h]
    \begin{minipage}{0.17\textwidth}
        \centering
        \includegraphics[width=\textwidth]{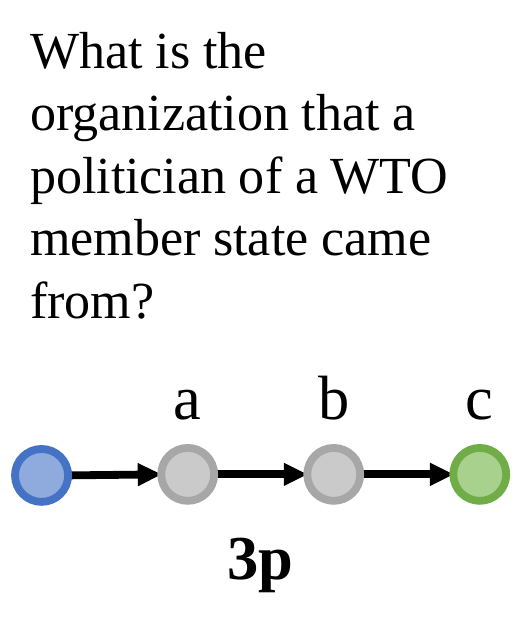}
    \end{minipage}
    \hspace{0.02\textwidth}
    \begin{minipage}{0.8\textwidth}
        \begin{adjustbox}{max width=\textwidth}
            \scriptsize
            \begin{tabular}{lccccc}
                \toprule
                \bf{Query} & \multicolumn{5}{l}{$q = ?c : \exists a, b: \text{MemberStates}(\text{WTO}, a) \land \text{JurisdicationOfOffice}(b, a) \land \text{Organization}(b, c)$} \\
                \midrule
                \multirow{2}{*}{\bf{Variable}} & \multicolumn{3}{c}{\bf{Top Predictions ($\geq 0.1$)}} & \bf{Random} & \bf{Filtered} \\
                & \bf{Easy} & \multicolumn{2}{c}{\bf{Hard}} & \bf{Ground Truth} & \bf{Ranking} \\
                \midrule
                a & \stack{\easy{Philippines} \\ \easy{Hong Kong} \\ \easy{Fiji}} & \stack{\fp{Yugoslavia} \\ \tp{Guyana} \\ \tp{Luxembourg}} & \stack{\tp{Greece} \\ \tp{Ghana} \\ \fp{Tajikistan}} & \stack{Fiji \\ (easy)} & 1 \\
                \midrule
                b & \stack{\easy{Robert F. Kennedy} \\ \easy{Thomas Jefferson} \\ \easy{John F. Kennedy}} & \stack{\tp{Mao Zedong} \\ \fp{Saddam Hussein} \\ \fp{Yasser Arafat}} & \stack{\fp{Chiang Kai-shek} \\ \fp{Georgy Zhukov} \\ \fp{Donald M. Payne}} & \stack{president \\ (easy)} & 1 \\
                \midrule
                c & \stack{\easy{University of Manitoba} \\ \easy{Kansas State University} \\ \easy{Columbia University}} & \stack{\fp{West Point} \\ \fp{U.S. Naval Academy} \\ \fp{Spyglass Media Group}} & \stack{\fp{Yale Law School} \\ \fp{Harvard Law School} \\ \fp{London School of Economics}} & \stack{Manhattan School of Music \\ (hard)} & 121 \\
                \bottomrule
            \end{tabular}
        \end{adjustbox}
    \end{minipage}
    \vspace{-0.5em}
\end{table*}
\begin{table*}[!h]
    \begin{minipage}{0.17\textwidth}
        \centering
        \includegraphics[width=\textwidth]{figures/visualization/3p_greece.pdf}
    \end{minipage}
    \hspace{0.02\textwidth}
    \begin{minipage}{0.8\textwidth}
        \begin{adjustbox}{max width=\textwidth}
            \scriptsize
            \begin{tabular}{lccccc}
                \toprule
                \bf{Query} & \multicolumn{5}{l}{$q = ?c : \exists a, b: \text{ParticipateCountry}(a, \text{Greece}) \land \text{OlympicSports}(a, b) \land \text{TeamSports}(c, b)$} \\
                \midrule
                \multirow{2}{*}{\bf{Variable}} & \multicolumn{3}{c}{\bf{Top Predictions ($\geq 0.1$)}} & \bf{Random} & \bf{Filtered} \\
                & \bf{Easy} & \multicolumn{2}{c}{\bf{Hard}} & \bf{Ground Truth} & \bf{Ranking} \\
                \midrule
                a & \stack{\easy{1936 Summer Olympics} \\ \easy{1980 Winter Olympics} \\ \easy{2002 Winter Olympics}} & \stack{\tp{2010 Winter Olympics} \\ \fp{2012 Summer Olympics} \\ \fp{1920 Summer Olympics}} & \stack{\fp{1988 Summer Olympics} \\ \fp{1928 Summer Olympics} \\ \fp{1992 Summer Olympics}} & \stack{2010 Winter Olympics \\ (hard)} & 1 \\
                \midrule
                b & \stack{\easy{soccer} \\ \easy{track and field} \\ \easy{water polo}} & \stack{\tp{luge} \\ \tp{ice hockey} \\ \tp{short track speed skating}} & \stack{\fp{tennis} \\ - \\ - } & \stack{ice hockey \\ (hard)} & 1 \\
                \midrule
                c & \stack{\easy{Sacramento Kings} \\ \easy{Utah Jazz} \\ \easy{Seattle SuperSonics}} & \stack{\tp{Algeria soccer team} \\ \tp{Cincinnati Reds} \\ \tp{Washington Nationals}} & \stack{\tp{Chile soccer team} \\ \tp{Cardiff City} \\ \tp{Blackburn Rovers}} & \stack{Florida Panthers \\ (hard)} & 433 \\
                \bottomrule
            \end{tabular}
        \end{adjustbox}
    \end{minipage}
    \vspace{-0.5em}
\end{table*}
\begin{table*}[!h]
    \centering
    \begin{minipage}{0.17\textwidth}
        \centering
        \includegraphics[width=\textwidth]{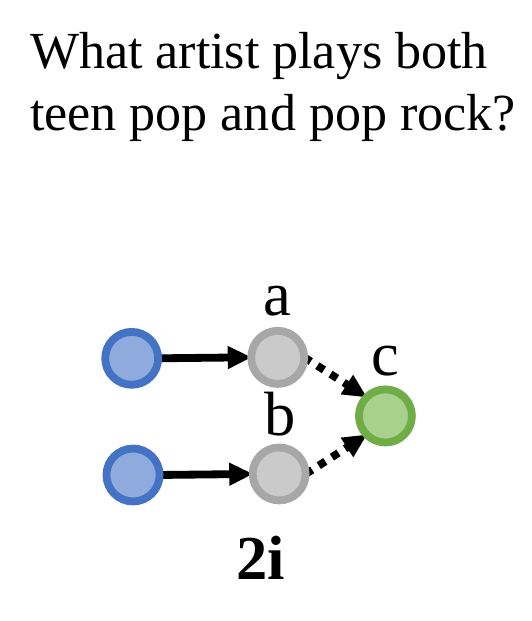}
    \end{minipage}
    \hspace{0.02\textwidth}
    \begin{minipage}{0.8\textwidth}
        \begin{adjustbox}{max width=\textwidth}
            \scriptsize
            \begin{tabular}{lccccc}
                \toprule
                \bf{Query} & \multicolumn{5}{l}{$q = ?c : \text{Artist}(\text{teen pop}, c) \land \text{Artist}(\text{pop rock}, c)$} \\
                \midrule
                \multirow{2}{*}{\bf{Variable}} & \multicolumn{3}{c}{\bf{Top Predictions ($\geq 0.1$)}} & \bf{Random} & \bf{Filtered} \\
                & \bf{Easy} & \multicolumn{2}{c}{\bf{Hard}} & \bf{Ground Truth} & \bf{Ranking} \\
                \midrule
                a & \stack{\easy{Jonas Brothers} \\ \easy{Ashley Tisdale} \\ \easy{Emma Roberts}} & \stack{\tp{Vanessa Hudgens} \\ \tp{Jesse McCartney} \\ \fp{Emily Osment}} & \stack{\fp{SM Town} \\ \fp{La Toya Jackson} \\ \tp{Justin Bieber}} & \stack{Kevin Jonas \\ (easy)} & 1 \\
                \midrule
                b & \stack{\easy{Blondie} \\ \easy{Rupert Holmes} \\ \easy{Bon Jovi}} & \stack{\tp{The Monkees} \\ \fp{Ringo Starr \& His All-Starr Band} \\ \tp{The Who}} & \stack{\fp{Queen} \\ \fp{Thin Lizzy} \\ \fp{Band Aid}} & \stack{Kevin Jonas \\ (hard)} & 336 \\
                \midrule
                c & \stack{\easy{Ashley Tisdale} \\ \easy{Emma Roberts} \\ \easy{Morning Musume}} & \stack{\tp{Jonas Brothers} \\ \fp{NSYNC} \\ \tp{Joe Jonas}} & \stack{\fp{Britney Spears} \\ \tp{Kevin Jonas} \\ \fp{Dr. Luke}} & \stack{Kevin Jonas \\ (hard)} & 4 \\
                \bottomrule
            \end{tabular}
        \end{adjustbox}
    \end{minipage}
    \vspace{-0.5em}
\end{table*}
\begin{table*}[!h]
    \begin{minipage}{0.17\textwidth}
        \centering
        \includegraphics[width=\textwidth]{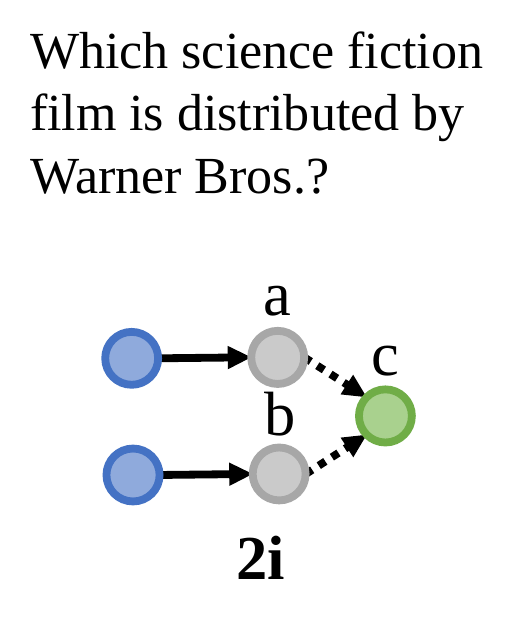}
    \end{minipage}
    \hspace{0.02\textwidth}
    \begin{minipage}{0.8\textwidth}
        \begin{adjustbox}{max width=\textwidth}
            \scriptsize
            \begin{tabular}{lccccc}
                \toprule
                \bf{Query} & \multicolumn{5}{l}{$q = ?c : \text{Genre}(c, \text{Science Fiction}) \land \text{DistributeFilm}(\text{Warner Bros.}, c)$} \\
                \midrule
                \multirow{2}{*}{\bf{Variable}} & \multicolumn{3}{c}{\bf{Top Predictions ($\geq 0.1$)}} & \bf{Random} & \bf{Filtered} \\
                & \bf{Easy} & \multicolumn{2}{c}{\bf{Hard}} & \bf{Ground Truth} & \bf{Ranking} \\
                \midrule
                a & \stack{\easy{Men in Black 3} \\ \easy{The Fifth Element} \\ \easy{Frankenweenie}} & \stack{\tp{Star Trek 10} \\ \tp{Star Trek 8} \\ \tp{Superman 4}} & \stack{\tp{Spiderman 3} \\ \fp{It} \\ \fp{Bleach Movie 3}} & \stack{Priest \\ (hard)} & 82 \\
                \midrule
                b & \stack{The Legend of Tarzan, \\ \easy{Lord of the Apes} \\ \easy{Never Say Never Again} \\ \easy{Sayonara}} & \stack{\fp{The Hangover 2} \\ \fp{Scooby-Doo} \\ Glee: The 3D \\ \fp{Concert Movie}} & \stack{\fp{Hamlet} \\ \fp{The Animatrix} \\ Superman/Batman: \\ \fp{Public Enemies}} & \stack{Priest \\ (easy)} & 1 \\
                \midrule
                c & \stack{\easy{A Clockwork Orange} \\ \easy{Terminator Salvation} \\ \easy{Superman: Man Of steel}} & \stack{\fp{The Animatrix} \\ \tp{Superman 4} \\ Superman/Batman: \\ \fp{Public Enemies}} & \stack{\fp{The Matrix 3} \\ \fp{Batman: Gotham Knight} \\ \tp{The Matrix 2}} & \stack{Priest \\ (hard)} & 28 \\
                \bottomrule
            \end{tabular}
        \end{adjustbox}
    \end{minipage}
    \vspace{-0.5em}
\end{table*}
\begin{table*}[!h]
    \begin{minipage}{0.17\textwidth}
        \centering
        \includegraphics[width=\textwidth]{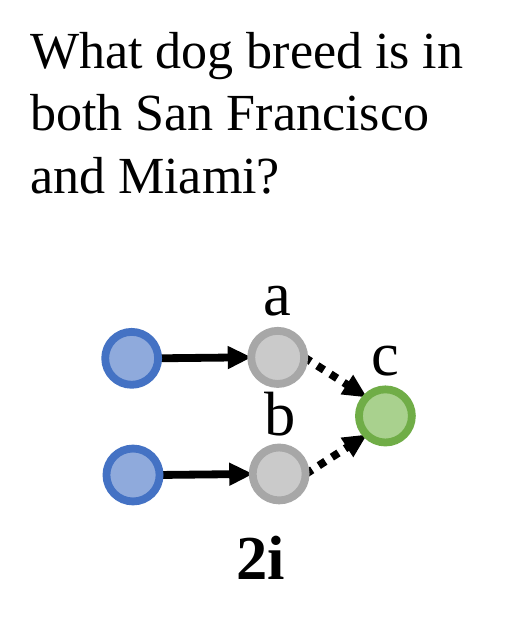}
    \end{minipage}
    \hspace{0.02\textwidth}
    \begin{minipage}{0.8\textwidth}
        \begin{adjustbox}{max width=\textwidth}
            \scriptsize
            \begin{tabular}{lccccc}
                \toprule
                \bf{Query} & \multicolumn{5}{l}{$q = ?c : \text{DogBreed}(\text{San Francisco}, c) \land \text{DogBreed}(\text{Miami}, c)$} \\
                \midrule
                \multirow{2}{*}{\bf{Variable}} & \multicolumn{3}{c}{\bf{Top Predictions ($\geq 0.1$)}} & \bf{Random} & \bf{Filtered} \\
                & \bf{Easy} & \multicolumn{2}{c}{\bf{Hard}} & \bf{Ground Truth} & \bf{Ranking} \\
                \midrule
                a & \stack{\easy{German Shepherd dog} \\ \easy{labrador retriever} \\ \easy{Bulldog}} & \stack{\tp{Yorkshire Terrier} \\ - \\ - } & \stack{- \\ - \\ -} & \stack{Yorkshire Terrier \\ (hard)} & 1 \\
                \midrule
                b & \stack{\easy{labrador retriever} \\ \easy{Bulldog} \\ \easy{Yorkshire Terrier}} & \stack{\tp{German Shepherd dog} \\ - \\ -} & \stack{- \\ - \\ -} & \stack{Yorkshire Terrier \\ (easy)} & 1 \\
                \midrule
                c & \stack{\easy{labrador retriever} \\ \easy{Bulldog} \\ \easy{Golden Retriever}} & \stack{\tp{German Shepherd dog} \\ \tp{Yorkshire Terrier} \\ - } & \stack{- \\ - \\ -} & \stack{Yorkshire Terrier \\ (hard)} & 1 \\
                \bottomrule
            \end{tabular}
        \end{adjustbox}
    \end{minipage}
    \vspace{-0.5em}
\end{table*}
\begin{table*}[!h]
    \centering
    \begin{minipage}{0.17\textwidth}
        \centering
        \includegraphics[width=\textwidth]{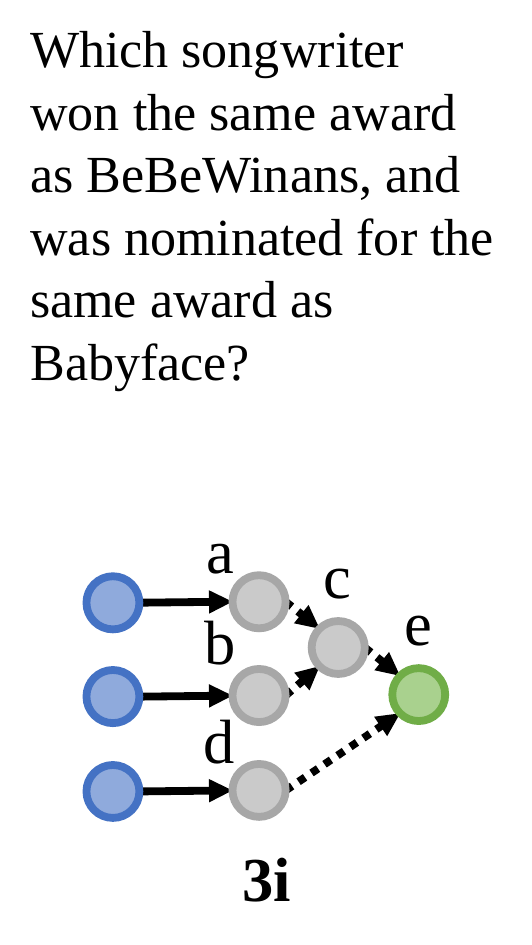}
    \end{minipage}
    \hspace{0.02\textwidth}
    \begin{minipage}{0.8\textwidth}
        \begin{adjustbox}{max width=\textwidth}
            \scriptsize
            \begin{tabular}{lccccc}
                \toprule
                \bf{Query} & \multicolumn{5}{l}{$q = ?e : \text{Profession}(e, \text{songwriter}) \land \text{WinnerOfSameAward}(e, \text{BeBe Winans}) \land \text{NominatedForSameAward}(\text{Babyface}, e)$} \\
                \midrule
                \multirow{2}{*}{\bf{Variable}} & \multicolumn{3}{c}{\bf{Top Predictions ($\geq 0.1$)}} & \bf{Random} & \bf{Filtered} \\
                & \bf{Easy} & \multicolumn{2}{c}{\bf{Hard}} & \bf{Ground Truth} & \bf{Ranking} \\
                \midrule
                a & \stack{\easy{Phil Vischer} \\ \easy{Dr. Seuss} \\ \easy{Michael Nesmith}} & \stack{\fp{Walter Scharf} \\ \tp{Paul Francis Webster} \\ \fp{DMX}} & \stack{\fp{Al Jourgensen} \\ \fp{Trey Anastasio} \\ \fp{George Duning}} & \stack{L.A. Reid \\ (easy)} & 1 \\
                \midrule
                b & \stack{\easy{Whitney Houston} \\ - \\ -} & \stack{\fp{CeCe Winans} \\ \fp{Babyface} \\ \fp{David Foster}} & \stack{- \\ - \\ -} & \stack{L.A. Reid \\ (hard)} & 6 \\
                \midrule
                c & \stack{\easy{Whitney Houston} \\ - \\ -} & \stack{\fp{Babyface} \\ \fp{David Foster} \\ \fp{CeCe Winans}} & \stack{- \\ - \\ -} & \stack{L.A. Reid \\ (hard)} & 6 \\
                \midrule
                d & \stack{\easy{Stephen Schwartz} \\ \easy{Eric Clapton} \\ \easy{T-Pain}} & \stack{\fp{Barry White} \\ \fp{David Banner} \\ \fp{Stevie Wonder}} & \stack{\fp{Static Major} \\ \fp{Whitney Houston} \\ \fp{BeBe Winans}} & \stack{L.A. Reid \\ (easy)} & 1 \\
                \midrule
                e & \stack{- \\ - \\ -} & \stack{\fp{Whitney Houston} \\ \fp{David Foster} \\ -} & \stack{- \\ - \\ -} & \stack{L.A. Reid \\ (hard)} & 3 \\
                \bottomrule
            \end{tabular}
        \end{adjustbox}
    \end{minipage}
    \vspace{-0.5em}
\end{table*}
\begin{table*}[!h]
    \begin{minipage}{0.17\textwidth}
        \centering
        \includegraphics[width=\textwidth]{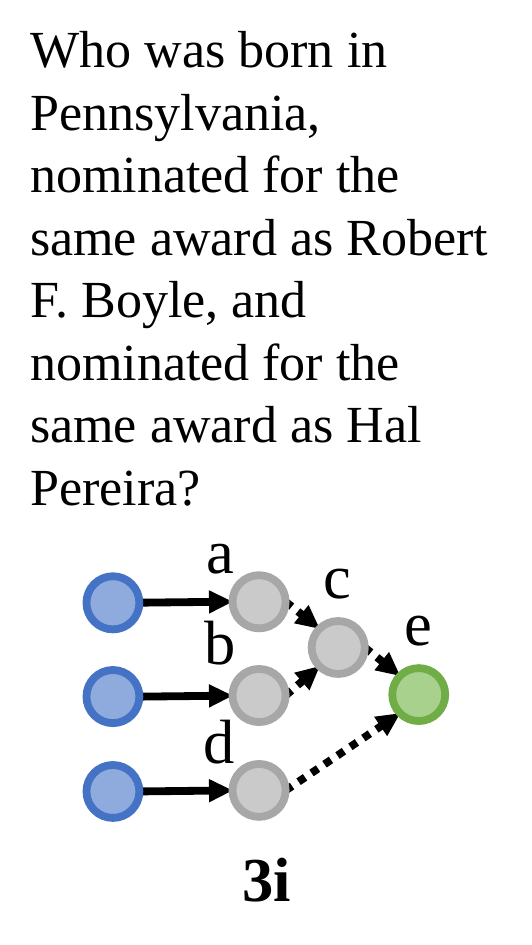}
    \end{minipage}
    \hspace{0.02\textwidth}
    \begin{minipage}{0.8\textwidth}
        \begin{adjustbox}{max width=\textwidth}
            \scriptsize
            \begin{tabular}{lccccc}
                \toprule
                \bf{Query} & \multicolumn{5}{l}{$q = ?e : \text{PlaceOfBirth}(e, \text{Pennsylvania}) \land \text{NominatedForSameAward}(e, \text{Robert F. Boyle}) \land \text{NominatedForSameAward}(\text{Hal Pereira}, e)$} \\
                \midrule
                \multirow{2}{*}{\bf{Variable}} & \multicolumn{3}{c}{\bf{Top Predictions ($\geq 0.1$)}} & \bf{Random} & \bf{Filtered} \\
                & \bf{Easy} & \multicolumn{2}{c}{\bf{Hard}} & \bf{Ground Truth} & \bf{Ranking} \\
                \midrule
                a & \stack{- \\ - \\ -} & \stack{\fp{Bam Mangera} \\ \fp{Jim Thorpe} \\ \fp{Matthew Fox}} & \stack{\fp{Dean Koontz} \\ \fp{Clark Gable} \\ \fp{Poul Anderson}} & \stack{Frank R. McKelvy \\ (hard)} & 770 \\
                \midrule
                b & \stack{\easy{Edward G. Boyle} \\ \easy{Frank R. McKelvy} \\ \easy{William A. Horning}} & \stack{\fp{Peter Lamont} \\ \fp{Ken Adam} \\ \tp{Henry Grace}} & \stack{\fp{Cedric Gibbons} \\ \fp{Edwin B. Willis} \\ \fp{Edward Carfagno}} & \stack{Frank R. McKelvy \\ (easy)} & 1 \\
                \midrule
                c & \stack{- \\ - \\ -} & \stack{\tp{Frank R. McKelvy} \\ \fp{Edward G. Boyle} \\ -} & \stack{- \\ - \\ -} & \stack{Frank R. McKelvy \\ (hard)} & 1 \\
                \midrule
                d & \stack{\easy{Frank R. McKelvy} \\ \easy{Ray Moyer} \\ \easy{Henry Bumstead}} & \stack{\fp{Joseph Kish} \\ \fp{Hans Dreier} \\ \fp{Richard Day}} & \stack{- \\ - \\ -} & \stack{Frank R. McKelvy \\ (easy)} & 1 \\
                \midrule
                e & \stack{- \\ - \\ -} & \stack{\tp{Frank R. McKelvy} \\ - \\ -} & \stack{- \\ - \\ -} & \stack{Frank R. McKelvy \\ (hard)} & 1 \\
                \bottomrule
            \end{tabular}
        \end{adjustbox}
    \end{minipage}
    \vspace{-0.5em}
\end{table*}
\begin{table*}[!h]
    \begin{minipage}{0.17\textwidth}
        \centering
        \includegraphics[width=\textwidth]{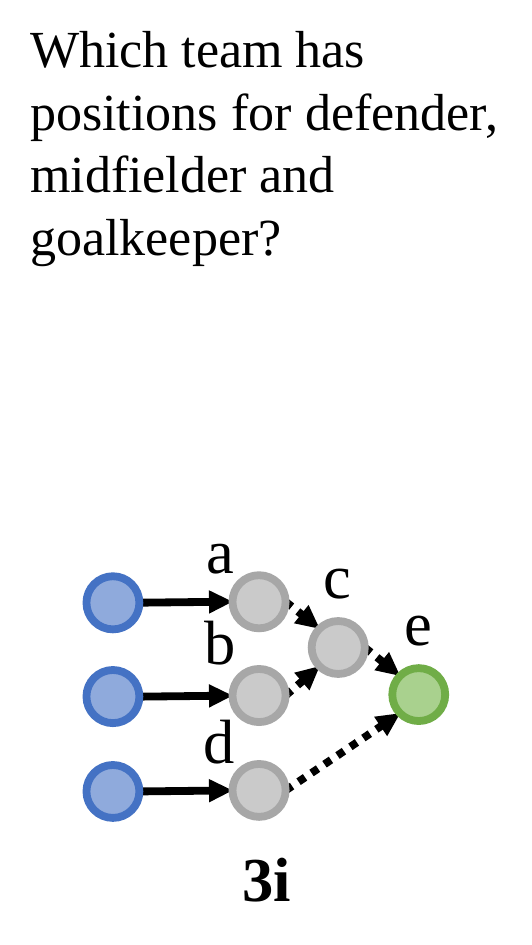}
    \end{minipage}
    \hspace{0.02\textwidth}
    \begin{minipage}{0.8\textwidth}
        \begin{adjustbox}{max width=\textwidth}
            \scriptsize
            \begin{tabular}{lccccc}
                \toprule
                \bf{Query} & \multicolumn{5}{l}{$q = ?e : \text{Position}(e, \text{defender (soccer)}) \land \text{Position}(e, \text{midfielder (soccer)}) \land \text{Position}(e, \text{goalkeeper (soccer)}$} \\
                \midrule
                \multirow{2}{*}{\bf{Variable}} & \multicolumn{3}{c}{\bf{Top Predictions ($\geq 0.1$)}} & \bf{Random} & \bf{Filtered} \\
                & \bf{Easy} & \multicolumn{2}{c}{\bf{Hard}} & \bf{Ground Truth} & \bf{Ranking} \\
                \midrule
                a & \stack{\easy{Hannover 96} \\ \easy{RB Leipzig} \\ \easy{Bayer 04 Leverkusen}} & \stack{\fp{Belarus soccer team} \\ \fp{U.S. Lecce} \\ \fp{A.C. Cesena}} & \stack{\fp{FC Torpedo Moscow} \\ \fp{Austria soccer team} \\ \fp{1. FC Union Berlin}} & \stack{Zamalek SC \\ (hard)} & 149 \\
                \midrule
                b & \stack{\easy{Hannover 96} \\ \easy{RB Leipzig} \\ \easy{FC Bayern Munich}} & \stack{\fp{Belarus soccer team} \\ \fp{Kazakhstan soccer team} \\ \fp{Croatia soccer team}} & \stack{\fp{Chile soccer team} \\ \fp{FC Torpedo Moscow} \\ \fp{FC Energie Cottbus}} & \stack{Zamalek SC \\ (easy)} & 1 \\
                \midrule
                c & \stack{\easy{Hannover 96} \\ \easy{RB Leipzig} \\ \easy{Bayer 04 Leverkusen}} & \stack{\fp{Bulgaria soccer team} \\ \fp{Belarus soccer team} \\ \fp{Kazakhstan soccer team}} & \stack{\fp{Croatia soccer team} \\ \fp{Chile soccer team} \\ \fp{U.S. Lecce}} & \stack{Zamalek SC \\ (hard)} & 218 \\
                \midrule
                d & \stack{\easy{RB Leipzig} \\ \easy{Hannover 96} \\ \easy{FC Volga Nizhny Novgorod}} & \stack{\fp{D.C. United} \\ \fp{F.C. Lorient} \\ \fp{Coritiba F.C.}} & \stack{\fp{South Africa soccer team} \\ \fp{Croatia soccer team} \\ \fp{Suwon Samsung Bluewings FC}} & \stack{Zamalek SC \\ (easy)} & 1 \\
\bottomrule
            \end{tabular}
        \end{adjustbox}
        \footnotetext{In reality, every soccer team has positions for defender, midfielder and goalkeeper. The predictions are considered wrong because the corresponding facts are missing in FB15k-237.}
    \end{minipage}
    \vspace{-0.5em}
\end{table*}
\begin{table*}[!h]
    \centering
    \begin{minipage}{0.17\textwidth}
        \centering
        \includegraphics[width=\textwidth]{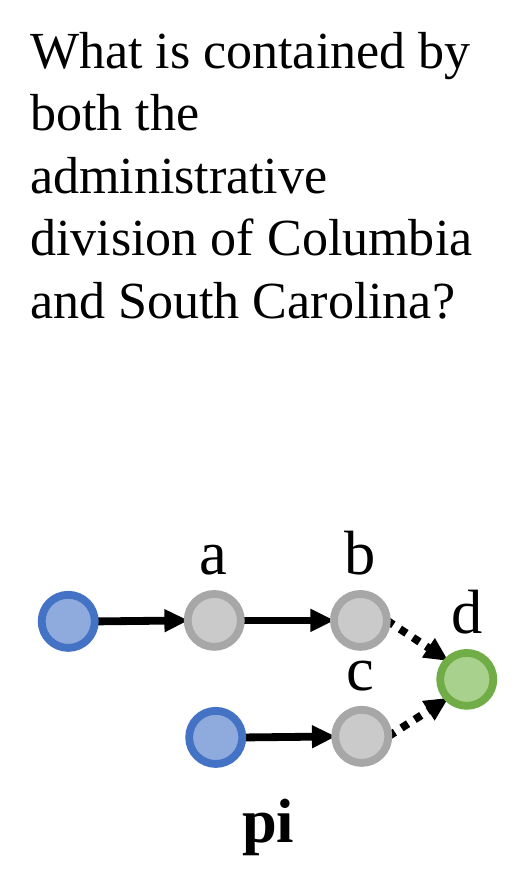}
    \end{minipage}
    \hspace{0.02\textwidth}
    \begin{minipage}{0.8\textwidth}
        \begin{adjustbox}{max width=\textwidth}
            \scriptsize
            \begin{tabular}{lccccc}
                \toprule
                \bf{Query} & \multicolumn{5}{l}{$q = ?d : \exists a: \text{AdministrativeDivision}(\text{Columbia}, a) \land \text{Contain}(a, d) \land \text{Contain}(\text{South Carolina}, d)$} \\
                \midrule
                \multirow{2}{*}{\bf{Variable}} & \multicolumn{3}{c}{\bf{Top Predictions ($\geq 0.1$)}} & \bf{Random} & \bf{Filtered} \\
                & \bf{Easy} & \multicolumn{2}{c}{\bf{Hard}} & \bf{Ground Truth} & \bf{Ranking} \\
                \midrule
                a & \stack{\easy{South Carolina} \\ - \\ -} & \stack{\fp{Lexington County} \\ - \\ -} & \stack{- \\ - \\ -} & \stack{South Carolina \\ (easy)} & 1 \\
                \midrule
                b & \stack{\easy{University of South Carolina} \\ \easy{Clemson University} \\ \easy{Furman University}} & \stack{\fp{Florida Keys} \\ \fp{Marquette County} \\ \fp{Bethesda}} & \stack{\fp{Passaic} \\ \fp{Fairbanks} \\ \fp{Johnstown}} & \stack{Greenville County \\ (hard)} & 34 \\
                \midrule
                c & \stack{\easy{Clemson University} \\ \easy{Furman University} \\ \easy{University of South Carolina}} & \stack{\fp{Marquette County} \\ \fp{Binghamton} \\ \fp{Passaic}} & \stack{\fp{Green Bay} \\ \tp{Spartanburg County} \\ \fp{Rochester}} & \stack{Greenville County \\ (hard)} & 34 \\
                \midrule
                d & \stack{\easy{Clemson University} \\ \easy{University of South Carolina} \\ \easy{Furman University}} & \stack{\fp{Marquette County} \\ \fp{Florida Keys} \\ -} & \stack{- \\ - \\ -} & \stack{Greenville County \\ (hard)} & 33 \\
                \bottomrule
            \end{tabular}
        \end{adjustbox}
        \footnotetext{Since the \emph{administrative division} of \emph{Columbia} is \emph{South Carolina}, this \emph{pi} query degenerates to a \emph{2p} query.}
    \end{minipage}
    \vspace{-0.5em}
\end{table*}
\begin{table*}
    \begin{minipage}{0.17\textwidth}
        \centering
        \includegraphics[width=\textwidth]{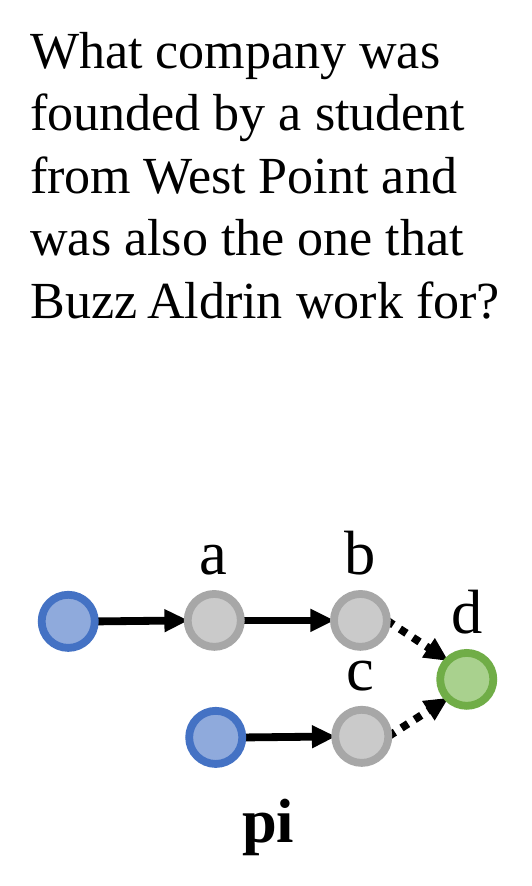}
    \end{minipage}
    \hspace{0.02\textwidth}
    \begin{minipage}{0.8\textwidth}
        \begin{adjustbox}{max width=\textwidth}
            \scriptsize
            \begin{tabular}{lccccc}
                \toprule
                \bf{Query} & \multicolumn{5}{l}{$q = ?d : \exists a: \text{Student}(\text{West Point}, a) \land \text{Found}(a, d) \land \text{Company}(\text{Buzz Aldrin}, d)$} \\
                \midrule
                \multirow{2}{*}{\bf{Variable}} & \multicolumn{3}{c}{\bf{Top Predictions ($\geq 0.1$)}} & \bf{Random} & \bf{Filtered} \\
                & \bf{Easy} & \multicolumn{2}{c}{\bf{Hard}} & \bf{Ground Truth} & \bf{Ranking} \\
                \midrule
                a & \stack{\easy{Timothy Leary} \\ \easy{David Petraeus} \\ \easy{Edgar Allan Poe}} & \stack{\fp{Grover Cleveland} \\ \fp{Theodore Roosevelt} \\ \fp{James Monroe}} & \stack{\fp{DeWitt Clinton} \\ \fp{Chris Elliott} \\ \fp{Gary Dourdan}} & \stack{Dwight D. Eisenhower \\ (easy)} & 1 \\
                \midrule
                b & \stack{\easy{CIA} \\ \easy{NASA} \\ -} & \stack{\fp{US Department of Defense} \\ \fp{University of California, Berkeley} \\ \fp{Columbia University}} & \stack{US Department of Housing \\ \fp{and Urban Development} \\ US Department of \\ \fp{the Air Force} \\ \fp{United Nations}} & \stack{NASA \\ (easy)} & 1 \\
                \midrule
                c & \stack{US Department of \\ \easy{the Air Force} \\ - \\ -} & \stack{\fp{ESPN } \\ \fp{Columbia University} \\ \fp{Microsoft}} & \stack{\fp{Havard University} \\ \fp{Oracle} \\ \fp{Dartmouth College}} & \stack{NASA \\ (hard)} & 934 \\
                \midrule
                d & \stack{- \\ - \\ -} & \stack{US Department of \\ \fp{the Air Force} \\ \tp{NASA} \\ -} & \stack{- \\ - \\ -} & \stack{NASA \\ (hard)} & 2 \\
                \bottomrule
            \end{tabular}
        \end{adjustbox}
    \end{minipage}
    \vspace{-0.5em}
\end{table*}
\begin{table*}
    \begin{minipage}{0.17\textwidth}
        \centering
        \includegraphics[width=\textwidth]{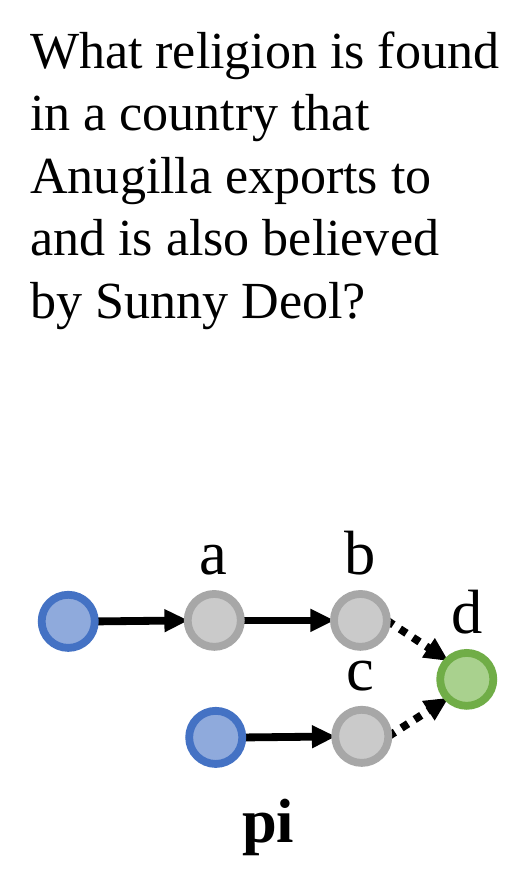}
    \end{minipage}
    \hspace{0.02\textwidth}
    \begin{minipage}{0.8\textwidth}
        \begin{adjustbox}{max width=\textwidth}
            \scriptsize
            \begin{tabular}{lccccc}
                \toprule
                \bf{Query} & \multicolumn{5}{l}{$q = ?d : \exists a: \text{ExportTo}(\text{Anguilla}, a) \land \text{Religion}(a, d) \land \text{Religion}(\text{Sunny Deol}, d)$} \\
                \midrule
                \multirow{2}{*}{\bf{Variable}} & \multicolumn{3}{c}{\bf{Top Predictions ($\geq 0.1$)}} & \bf{Random} & \bf{Filtered} \\
                & \bf{Easy} & \multicolumn{2}{c}{\bf{Hard}} & \bf{Ground Truth} & \bf{Ranking} \\
                \midrule
                a & \stack{\easy{United States of America} \\ \easy{United Kingdom} \\ -} & \stack{- \\ - \\ -} & \stack{- \\ - \\ -} & \stack{United Kingdom \\ (easy)} & 1 \\
                \midrule
                b & \stack{\easy{atheism} \\ \easy{Hinduism} \\ \easy{Judaism}} & \stack{\tp{Christianity} \\ \fp{Protestantism} \\ \fp{agnosticism}} & \stack{\fp{Lutheranism} \\ \tp{Methodism} \\ \fp{Anglicanism}} & \stack{Sikhism \\ (hard)} & 6 \\
                \midrule
                c & \stack{\easy{Hinduism} \\ \easy{Sikhism} \\ -} & \stack{\fp{Hindu} \\ \fp{Methodism} \\ \fp{Baptists}} & \stack{\fp{Presbyterianism} \\ \fp{Pentecostalism} \\ \fp{Nondenominational Christianity}} & \stack{Sikhism \\ (easy)} & 1 \\
                \midrule
                d & \stack{\easy{Hinduism} \\ - \\ -} & \stack{\fp{Islam} \\ \tp{Sikhism} \\ -} & \stack{- \\ - \\ -} & \stack{Sikhism \\ (hard)} & 2 \\
                \bottomrule
            \end{tabular}
        \end{adjustbox}
    \end{minipage}
    \vspace{-0.5em}
\end{table*}
\begin{table*}[!h]
    \centering
    \begin{minipage}{0.17\textwidth}
        \centering
        \includegraphics[width=\textwidth]{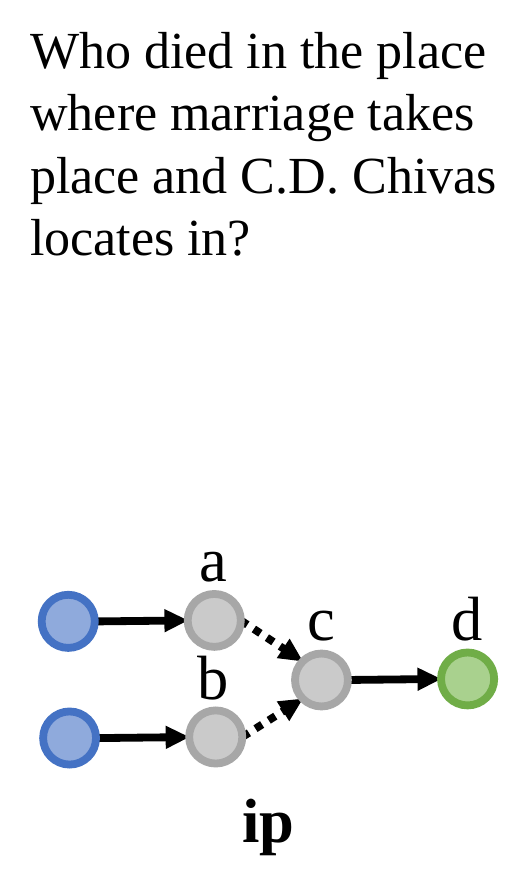}
    \end{minipage}
    \hspace{0.02\textwidth}
    \begin{minipage}{0.8\textwidth}
        \begin{adjustbox}{max width=\textwidth}
            \scriptsize
            \begin{tabular}{lccccc}
                \toprule
                \bf{Query} & \multicolumn{5}{l}{$q = ?d : \exists c: \text{LocationOfCeremony}(\text{marraige}, c) \land \text{LocalTeam}(c, \text{C.D. Chivas}) \land \text{PlaceOfDeath}(d, c)$} \\
                \midrule
                \multirow{2}{*}{\bf{Variable}} & \multicolumn{3}{c}{\bf{Top Predictions ($\geq 0.1$)}} & \bf{Random} & \bf{Filtered} \\
                & \bf{Easy} & \multicolumn{2}{c}{\bf{Hard}} & \bf{Ground Truth} & \bf{Ranking} \\
                \midrule
                a & \stack{\easy{Tijuana} \\ \easy{Jerusalem} \\ \easy{Puerto Rico}} & \stack{\fp{Tehran} \\ \tp{Genoa} \\ \fp{Tunis}} & \stack{\fp{Monterrey} \\ \fp{Green Bay} \\ \fp{Binghamton}} & \stack{Los Angeles \\ (easy)} & 1 \\
                \midrule
                b & \stack{\easy{Los Angeles} \\ - \\ -} & \stack{\fp{Museo del Prado} \\ \fp{Lund} \\ \fp{Kharkiv}} & \stack{\fp{Rawalpindi} \\ \fp{Innsbruck} \\ \fp{Katowice}} & \stack{Los Angeles \\ (easy)} & 1 \\
                \midrule
                c & \stack{\easy{Los Angeles} \\ - \\ -} & \stack{\fp{Museo del Prado} \\ \fp{Guatemala City} \\ \fp{Seville}} & \stack{\fp{Bilbao} \\ \fp{Rabat} \\ \fp{Tunis}} & \stack{Los Angeles \\ (easy)} & 1 \\
                \midrule
                d & \stack{\easy{Ralph Burns} \\ \easy{Robert F. Boyle} \\ \easy{Boris Leven}} & \stack{\fp{William Travilla} \\ \fp{Fred MacMurray} \\ \fp{Chuck Jones}} & \stack{\fp{Jerry Wald} \\ \fp{Louis Armstrong} \\ \fp{Gregory Peck}} & \stack{Ida Lupino \\ (hard)} & 15 \\
                \bottomrule
            \end{tabular}
        \end{adjustbox}
        \footnotetext{In reality, marriage can take place in any city. The predictions are considered wrong because the corresponding facts are missing in FB15k-237.}
    \end{minipage}
    \vspace{-0.5em}
\end{table*}
\begin{table*}[!h]
    \centering
    \begin{minipage}{0.17\textwidth}
        \centering
        \includegraphics[width=\textwidth]{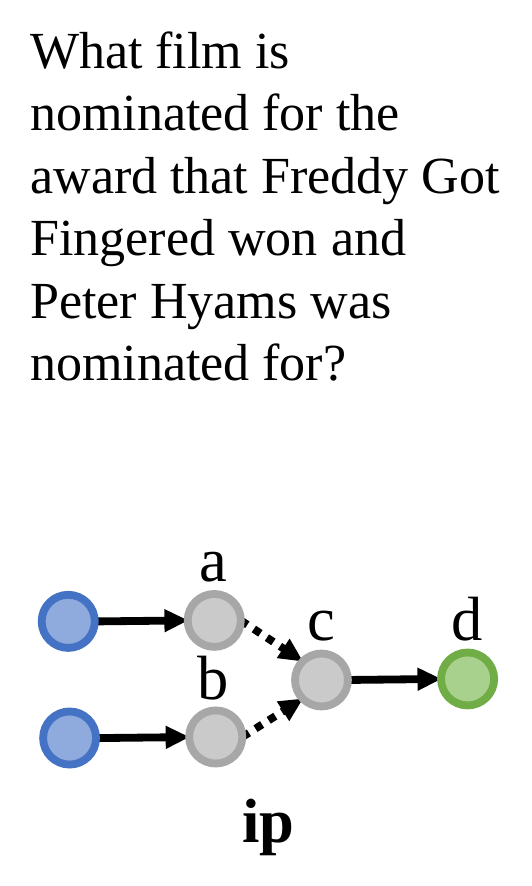}
    \end{minipage}
    \hspace{0.02\textwidth}
    \begin{minipage}{0.8\textwidth}
        \begin{adjustbox}{max width=\textwidth}
            \scriptsize
            \begin{tabular}{lccccc}
                \toprule
                \bf{Query} & \multicolumn{5}{l}{$q = ?d : \exists c: \text{Award}(\text{Freddy Got Fingered}, c) \land \text{Nominated}(\text{Peter Hyams}, c) \land \text{NominatedFor}(d, c)$} \\
                \midrule
                \multirow{2}{*}{\bf{Variable}} & \multicolumn{3}{c}{\bf{Top Predictions ($\geq 0.1$)}} & \bf{Random} & \bf{Filtered} \\
                & \bf{Easy} & \multicolumn{2}{c}{\bf{Hard}} & \bf{Ground Truth} & \bf{Ranking} \\
                \midrule
                a & \stack{Golden Raspberry Award for \\ \easy{Worst Screen Couple} \\ Golden Raspberry Award for \\ \easy{Worst Picture} \\ Golden Raspberry Award for \\ \easy{Worst Actor}} & \stack{Golden Raspberry Award for \\ \fp{Worst Remake} \\ Golden Raspberry Award for \\ \fp{Worst Supporting Actor} \\ Golden Raspberry Award for \\ \fp{Worst Supporting Actress}} & \stack{Golden Raspberry Award for \\ \fp{Worst Actress} \\ - \\ -} & \stack{Golden Raspberry Award for \\ Worst Director \\ (easy)} & 1 \\
                \midrule
                b & \stack{Golden Raspberry Award for \\ \easy{Worst Director} \\ - \\ -} & \stack{Golden Raspberry Award for \\ \fp{Worst Supporting Actor} \\ Academy Award for \\ \fp{Best Animated Short Film} \\ Academy Award for \\ \fp{Best Live Action Short Film}} & \stack{Academy Award for \\ \fp{Best Story} \\ Academy Award for \\ \fp{Best Documentary Feature} \\ Satellite Award for \\ \fp{Best Adapted Screenplay}} & \stack{Golden Raspberry Award for \\ Worst Director \\ (easy)} & 1 \\
                \midrule
                c & \stack{Golden Raspberry Award for \\ \easy{Worst Director} \\ - \\ -} & \stack{Golden Raspberry Award for \\ \fp{Worst Screenplay} \\ Golden Raspberry Award for \\ \fp{Worst Actor} \\ Golden Raspberry Award for \\ \fp{Worst Picture}} & \stack{Golden Raspberry Award for \\ \fp{Worst Screen Couple} \\ - \\ -} & \stack{Golden Raspberry Award for \\ Worst Director \\ (easy)} & 1 \\
\bottomrule
            \end{tabular}
        \end{adjustbox}
    \end{minipage}
    \vspace{-0.5em}
\end{table*}
\begin{table*}[!h]
    \centering
    \begin{minipage}{0.17\textwidth}
        \centering
        \includegraphics[width=\textwidth]{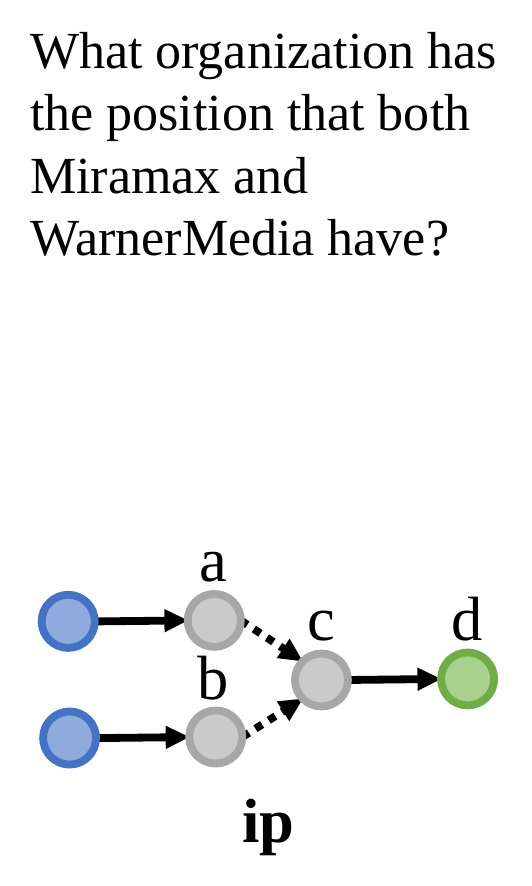}
    \end{minipage}
    \hspace{0.02\textwidth}
    \begin{minipage}{0.8\textwidth}
        \begin{adjustbox}{max width=\textwidth}
            \scriptsize
            \begin{tabular}{lccccc}
                \toprule
                \bf{Query} & \multicolumn{5}{l}{$q = ?d : \exists c: \text{Organization}(c, \text{Miramax}) \land \text{Company}(c, \text{WarnerMedia}) \land \text{Organization}(c, d)$} \\
                \midrule
                \multirow{2}{*}{\bf{Variable}} & \multicolumn{3}{c}{\bf{Top Predictions ($\geq 0.1$)}} & \bf{Random} & \bf{Filtered} \\
                & \bf{Easy} & \multicolumn{2}{c}{\bf{Hard}} & \bf{Ground Truth} & \bf{Ranking} \\
                \midrule
                a & \stack{\easy{chief executive officer} \\ - \\ -} & \stack{- \\ - \\ -} & \stack{- \\ - \\ -} & \stack{chief executive officer \\ (easy)} & 1 \\
                \midrule
                b & \stack{\easy{president} \\ \easy{chief operating officer} \\ \easy{chief executive officer}} & \stack{\fp{chief technology officer} \\ \fp{treasurer} \\ \fp{chief marketing officer}} & \stack{\fp{general manager} \\ \fp{chief administrative officer} \\ \fp{vice president}} & \stack{chief executive officer \\ (easy)} & 1 \\
                \midrule
                c & \stack{\easy{chief executive officer} \\ - \\ -} & \stack{- \\ - \\ -} & \stack{- \\ - \\ -} & \stack{chief executive officer \\ (easy)} & 1 \\
                \midrule
                d & \stack{\easy{Miramax} \\ \easy{New Line Cinema} \\ \easy{Lionsgate}} & \stack{\fp{KSA Network} \\ \fp{DC Comics} \\ \fp{ESPN}} & \stack{\fp{LucasArts} \\ \fp{GMA Network} \\ \fp{Spyglass Media Group}} & \stack{TV5 \\ (hard)} & 425 \\
                \bottomrule
            \end{tabular}
        \end{adjustbox}
        \footnotetext{In reality, most big companies have positions for CEO, CTO, etc. The predictions are considered wrong because the corresponding facts are missing in FB15k-237.}
    \end{minipage}
    \vspace{-0.5em}
\end{table*}
\begin{table*}[!h]
    \centering
    \begin{minipage}{0.17\textwidth}
        \centering
        \includegraphics[width=\textwidth]{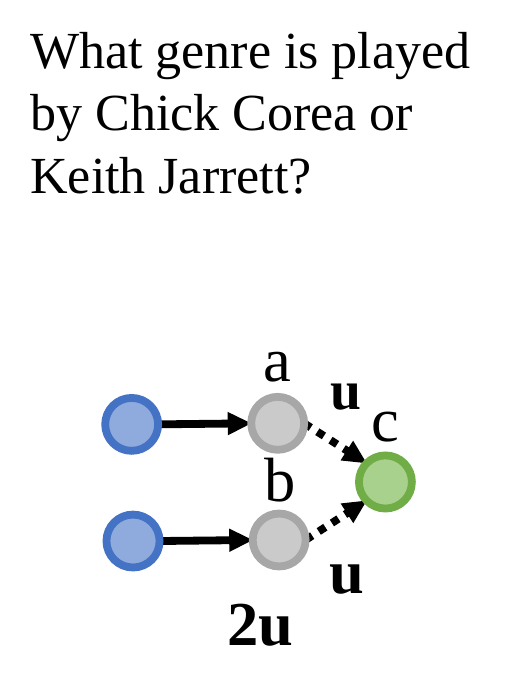}
    \end{minipage}
    \hspace{0.02\textwidth}
    \begin{minipage}{0.8\textwidth}
        \begin{adjustbox}{max width=\textwidth}
            \scriptsize
            \begin{tabular}{lccccc}
                \toprule
                \bf{Query} & \multicolumn{5}{l}{$q = ?c : \text{Artist}(c, \text{Chick Corea}) \lor \text{Artist}(c, \text{Keith Jarrett})$} \\
                \midrule
                \multirow{2}{*}{\bf{Variable}} & \multicolumn{3}{c}{\bf{Top Predictions ($\geq 0.1$)}} & \bf{Random} & \bf{Filtered} \\
                & \bf{Easy} & \multicolumn{2}{c}{\bf{Hard}} & \bf{Ground Truth} & \bf{Ranking} \\
                \midrule
                a & \stack{\easy{smooth jazz} \\ \easy{jazz-funk} \\ \easy{jazz fusion}} & \stack{\fp{free jazz} \\ \fp{hard bop} \\ \tp{bebop}} & \stack{\fp{funk} \\ \fp{sophisti-pop} \\ \fp{big band}} & \stack{bebop \\ (hard)} & 3 \\
                \midrule
                b & \stack{\easy{free jazz} \\ \easy{jazz fusion} \\ \easy{post-bop}} & \stack{\tp{avant-garde jazz} \\ \fp{hard bop} \\ \fp{bebop}} & \stack{\fp{Baroque music} \\ \tp{smooth jazz} \\ \fp{progressive rock}} & \stack{jazz \\ (easy)} & 1 \\
                \midrule
                c & \stack{\easy{post-bop} \\ \easy{classical music} \\ \easy{jazz}} & \stack{\fp{hard bop} \\ \tp{bebop} \\ \fp{rock music}} & \stack{\fp{Baroque music} \\ \fp{cool jazz} \\ \fp{funk}} & \stack{bebop \\ (hard)} & 2 \\
                \bottomrule
            \end{tabular}
        \end{adjustbox}
    \end{minipage}
    \vspace{-0.5em}
\end{table*}
\begin{table*}[!h]
    \centering
    \begin{minipage}{0.17\textwidth}
        \centering
        \includegraphics[width=\textwidth]{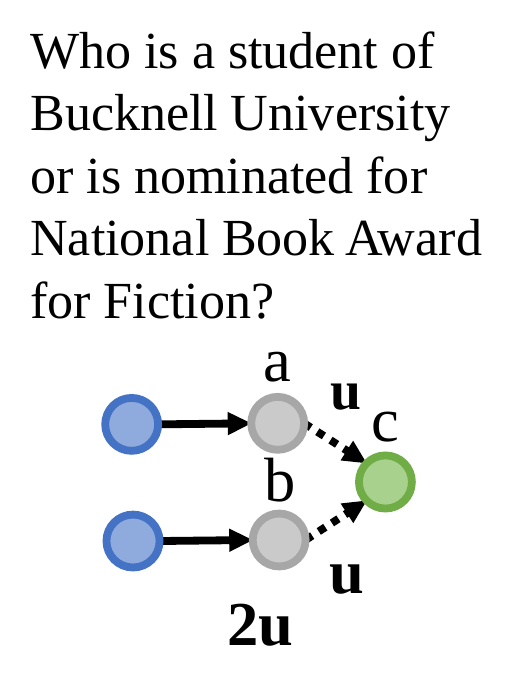}
    \end{minipage}
    \hspace{0.02\textwidth}
    \begin{minipage}{0.8\textwidth}
        \begin{adjustbox}{max width=\textwidth}
            \scriptsize
            \begin{tabular}{lccccc}
                \toprule
                \bf{Query} & \multicolumn{5}{l}{$q = ?c : \text{Student}(\text{Bucknell University}, c) \lor \text{Nominated}(c, \text{National Book Award for Fiction})$} \\
                \midrule
                \multirow{2}{*}{\bf{Variable}} & \multicolumn{3}{c}{\bf{Top Predictions ($\geq 0.1$)}} & \bf{Random} & \bf{Filtered} \\
                & \bf{Easy} & \multicolumn{2}{c}{\bf{Hard}} & \bf{Ground Truth} & \bf{Ranking} \\
                \midrule
                a & \stack{\easy{Philip Roth} \\ - \\ -} & \stack{\fp{Kate Flannery} \\ \fp{Hogan Sheffer} \\ \fp{James Buchanan}} & \stack{\fp{Pandro S. Berman} \\ \fp{Gary Dourdan} \\ \fp{Martie Maguire}} & \stack{Philip Roth \\ (easy)} & 1 \\
                \midrule
                b & \stack{\easy{James Baldwin} \\ \easy{J. D. Salinger} \\ \easy{Vladimir Nabokov}} & \stack{\fp{Joyce Carol Oates} \\ \fp{Michael Chabon} \\ \fp{James A. Michener}} & \stack{\fp{Larry McMurtry} \\ \fp{James Joyce} \\ \fp{Fyodor Dostoyevsky}} & \stack{Ursula K. Le Guin \\ (hard)} & 24 \\
                \midrule
                c & \stack{\easy{Philip Roth} \\ \easy{James Baldwin} \\ \easy{J. D. Salinger}} & \stack{\fp{Joyce Carol Oates} \\ \fp{James A. Michener} \\ \fp{Michael Chabon}} & \stack{\fp{Andy Warhol} \\ \fp{Lloyd Alexander} \\ \fp{Noam Chomsky}} & \stack{Ursula K. Le Guin \\ (hard)} & 40 \\
                \bottomrule
            \end{tabular}
        \end{adjustbox}
    \end{minipage}
    \vspace{-0.5em}
\end{table*}
\begin{table*}[!h]
    \centering
    \begin{minipage}{0.17\textwidth}
        \centering
        \includegraphics[width=\textwidth]{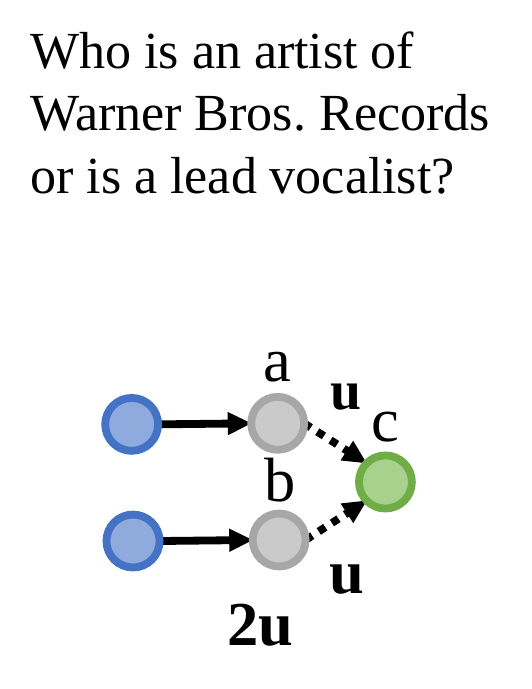}
    \end{minipage}
    \hspace{0.02\textwidth}
    \begin{minipage}{0.8\textwidth}
        \begin{adjustbox}{max width=\textwidth}
            \scriptsize
            \begin{tabular}{lccccc}
                \toprule
                \bf{Query} & \multicolumn{5}{l}{$q = ?c : \text{Artist}(\text{Warner Bros. Records}, c) \lor \text{Role}(c, \text{lead vocalist})$} \\
                \midrule
                \multirow{2}{*}{\bf{Variable}} & \multicolumn{3}{c}{\bf{Top Predictions ($\geq 0.1$)}} & \bf{Random} & \bf{Filtered} \\
                & \bf{Easy} & \multicolumn{2}{c}{\bf{Hard}} & \bf{Ground Truth} & \bf{Ranking} \\
                \midrule
                a & \stack{\easy{Sheila E.} \\ \easy{Linkin Park} \\ \easy{Metallica}} & \stack{\fp{Los Lobos} \\ \fp{Gong} \\ \fp{Dir En Grey}} & \stack{\fp{Gwar} \\ \fp{The Smashing Pumpkins} \\ \fp{Tony Levin}} & \stack{Randy Newman \\ (hard)} & 366 \\
                \midrule
                b & \stack{\easy{Bernadette Peters} \\ \easy{John Gielgud} \\ \easy{Andy Dick}} & \stack{\fp{Rick Springfield} \\ \tp{Steve Winwood} \\ \fp{Lou Reed}} & \stack{\fp{Chris Seefried} \\ \fp{David Grisman} \\ \fp{Arjen Anthony Lucassen}} & \stack{Eddie Vedder \\ (easy)} & 1 \\
                \midrule
                c & \stack{\easy{Sheila E.} \\ \easy{Chaka Khan} \\ \easy{Metallica}} & \stack{\fp{Los Lobos} \\ \fp{Glen Ballard} \\ \fp{Brad Paisley}} & \stack{\fp{Tony Levin} \\ \fp{Rick Springfield} \\ \fp{Meat Loaf}} & \stack{Randy Newman \\ (hard)} & 269 \\
                \bottomrule
            \end{tabular}
        \end{adjustbox}
    \end{minipage}
    \vspace{-0.5em}
\end{table*}
\begin{table*}[!h]
    \centering
    \begin{minipage}{0.17\textwidth}
        \centering
        \includegraphics[width=\textwidth]{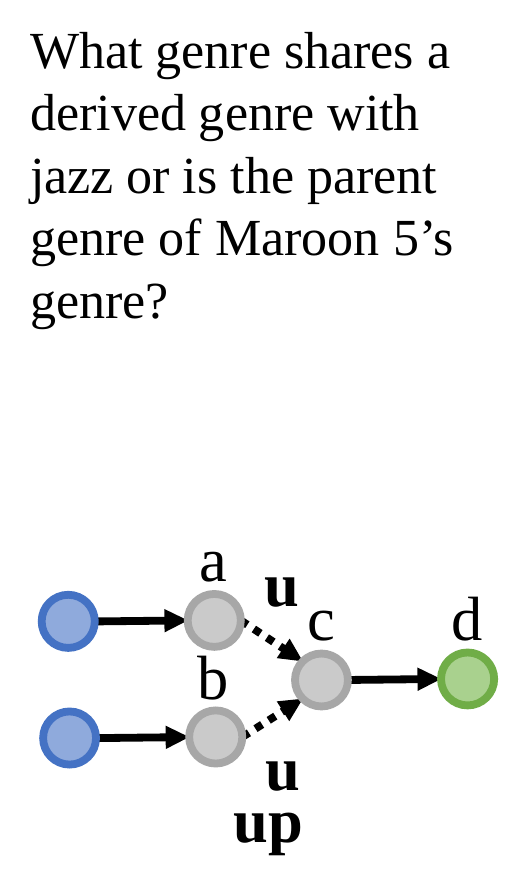}
    \end{minipage}
    \hspace{0.02\textwidth}
    \begin{minipage}{0.8\textwidth}
        \begin{adjustbox}{max width=\textwidth}
            \scriptsize
            \begin{tabular}{lccccc}
                \toprule
                \bf{Query} & \multicolumn{5}{l}{$q = ?d : \exists c: (\text{ParentGenre}(c, \text{jazz}) \lor \text{Artist}(c, \text{Maroon 5})) \land \text{ParentGenre}(c, d)$} \\
                \midrule
                \multirow{2}{*}{\bf{Variable}} & \multicolumn{3}{c}{\bf{Top Predictions ($\geq 0.1$)}} & \bf{Random} & \bf{Filtered} \\
                & \bf{Easy} & \multicolumn{2}{c}{\bf{Hard}} & \bf{Ground Truth} & \bf{Ranking} \\
                \midrule
                a & \stack{\easy{drum and bass} \\ \easy{post-rock} \\ \easy{technical death metal}} & \stack{\tp{Dixieland} \\ \tp{bossa nova} \\ \fp{sophisti-pop}} & \stack{\tp{soul jazz} \\ \fp{doo-wop} \\ \fp{progressive bluegrass}} & \stack{bluegrass music \\ (easy)} & 1 \\
                \midrule
                b & \stack{\easy{techno} \\ \easy{ska punk} \\ \easy{jazz fusion}} & \stack{\fp{power pop} \\ \fp{Britpop} \\ \fp{synth-pop}} & \stack{\fp{rhythm and blues} \\ \fp{post-punk} \\ \fp{electro}} & \stack{pop music \\ (easy)} & 1 \\
                \midrule
                c & \stack{\easy{jazz fusion} \\ \easy{jazz rap} \\ \easy{techno}} & \stack{\fp{jam band} \\ \fp{sophisti-pop} \\ \fp{doo-wop}} & \stack{\fp{power pop} \\ \fp{progressive metal} \\ \fp{electro}} & \stack{pop music \\ (easy)} & 1 \\
                \midrule
                d & \stack{\easy{gospel music} \\ \easy{folk music} \\ \easy{blues}} & \stack{\fp{folk rock} \\ \fp{blues rock} \\ \tp{hard rock}} & \stack{\tp{psychedelic rock} \\ \fp{pop rock} \\ \tp{glam rock}} & \stack{dance music \\ (hard)} & 96 \\
                \bottomrule
            \end{tabular}
        \end{adjustbox}
    \end{minipage}
    \vspace{-0.5em}
\end{table*}
\begin{table*}[!h]
    \centering
    \begin{minipage}{0.17\textwidth}
        \centering
        \includegraphics[width=\textwidth]{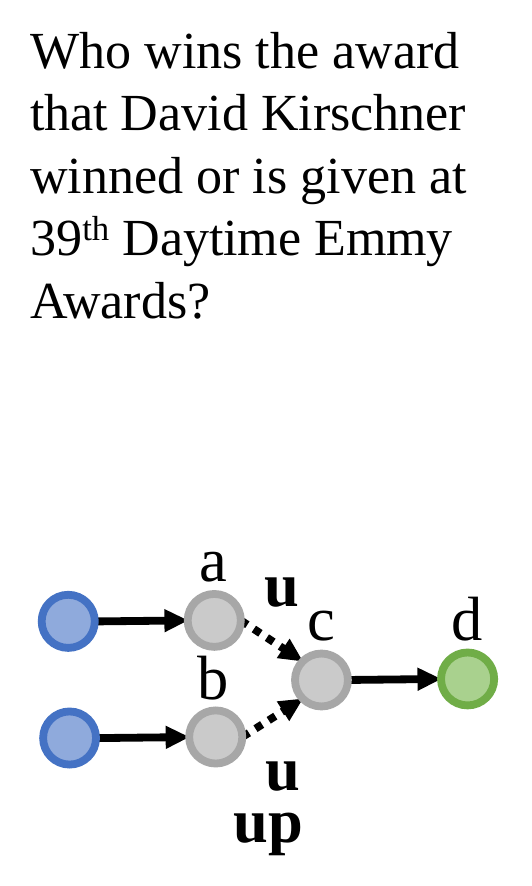}
    \end{minipage}
    \hspace{0.02\textwidth}
    \begin{minipage}{0.8\textwidth}
        \begin{adjustbox}{max width=\textwidth}
            \scriptsize
            \begin{tabular}{lccccc}
                \toprule
                \bf{Query} & \multicolumn{5}{l}{$q = ?d : \exists c: (\text{Award}(\text{David Kirschner}, c) \lor \text{Ceremony}(c, \text{39th Daytime Emmy Awards})) \land \text{Award}(d, c)$} \\
                \midrule
                \multirow{2}{*}{\bf{Variable}} & \multicolumn{3}{c}{\bf{Top Predictions ($\geq 0.1$)}} & \bf{Random} & \bf{Filtered} \\
                & \bf{Easy} & \multicolumn{2}{c}{\bf{Hard}} & \bf{Ground Truth} & \bf{Ranking} \\
                \midrule
                a & \stack{Daytime Emmy Award for \\ \easy{Outstanding Animated Program} \\ - \\ -} & \stack{Academy Award for \\ \fp{Best Animated Short Film} \\ Primetime Emmy Award for \\ \fp{Outstanding Comedy Series} \\ Academy Award for \\ \fp{Best Adapted Screenplay}} & \stack{Academy Award for \\ \fp{Best Story} \\ Academy Award for \\ \fp{Best Original Screenplay} \\ Golden Raspberry Award for \\ \fp{Worst Director}} & \stack{Daytime Emmy Award for \\ Outstanding Animated Program \\ (easy)} & 1 \\
                \midrule
                b & \stack{Daytime Emmy Award for \\ \easy{Outstanding Animated Program} \\ Daytime Emmy Award for \\ \easy{Outstanding Talk Show Host} \\ Daytime Emmy Award for \\ \easy{Outstanding Younger Actress in Drama}} & \stack{Daytime Emmy Award for \\ \tp{Outstanding Drama Series} \\ Writers Guild of America Award for \\ \fp{Best Adapted Screenplay} \\ Writers Guild of America Award for \\ \fp{Best Original Screenplay}} & \stack{Daytime Emmy Award for \\ \fp{Outstanding Talk Show} \\ \fp{soap opera} \\ \fp{game show}} & \stack{Daytime Emmy Award for \\ Outstanding Game Show \\ (easy)} & 1 \\
\bottomrule
            \end{tabular}
        \end{adjustbox}
    \end{minipage}
    \vspace{-0.5em}
\end{table*}
\begin{table*}[!h]
    \centering
    \begin{minipage}{0.17\textwidth}
        \centering
        \includegraphics[width=\textwidth]{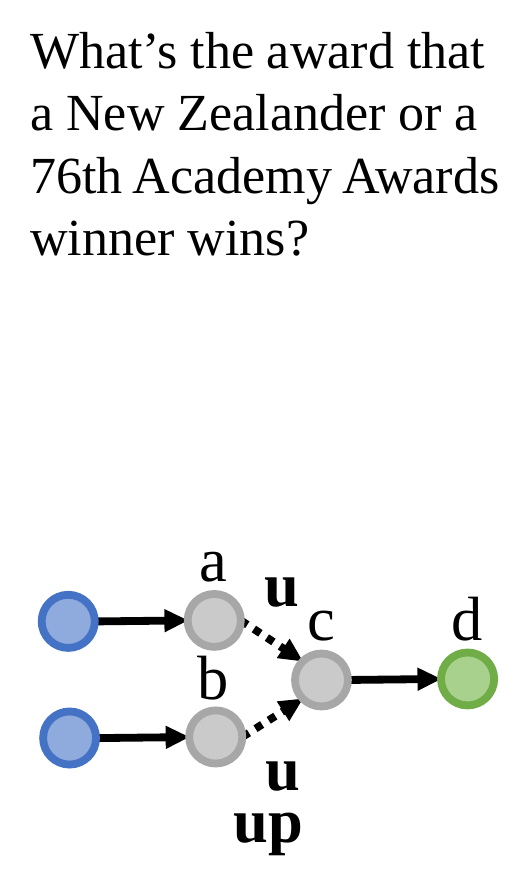}
    \end{minipage}
    \hspace{0.02\textwidth}
    \begin{minipage}{0.8\textwidth}
        \begin{adjustbox}{max width=\textwidth}
            \scriptsize
            \begin{tabular}{lccccc}
                \toprule
                \bf{Query} & \multicolumn{5}{l}{$q = ?d : \exists c: (\text{Nationality}(c, \text{New Zealand}) \lor \text{AwardWinner}(\text{76th Academy Awards}, c)) \land \text{Award}(c, d)$} \\
                \midrule
                \multirow{2}{*}{\bf{Variable}} & \multicolumn{3}{c}{\bf{Top Predictions ($\geq 0.1$)}} & \bf{Random} & \bf{Filtered} \\
                & \bf{Easy} & \multicolumn{2}{c}{\bf{Hard}} & \bf{Ground Truth} & \bf{Ranking} \\
                \midrule
                a & \stack{\easy{Keith Urban} \\ \easy{Chris Wood} \\ \easy{Jemaine Clement}} & \stack{\tp{Graeme Revell} \\ \tp{Fran Walsh} \\ \fp{Jack Thompson}} & \stack{\fp{Yoram Globus} \\ \fp{Christopher Columbus} \\ \fp{David Petraeus}} & \stack{Karl Urban \\ (easy)} & 1 \\
                \midrule
                b & \stack{\easy{Andrew Stanton} \\ \easy{Tim Robbins} \\ \easy{Annie Lennox}} & \stack{\fp{Focus Features} \\ \fp{Bill Murray} \\ \fp{Peter Weir}} & \stack{\fp{Brian Helgeland} \\ \fp{Scarlett Johansson} \\ \fp{Henry Bumstead}} & \stack{Howard Shore \\ (hard)} & 15 \\
                \midrule
                c & \stack{\easy{Peter Jackson} \\ \easy{Richard Taylor} \\ \easy{Keith Urban}} & \stack{\tp{Graeme Revell} \\ \fp{Jack Thompson} \\ \fp{Focus Features}} & \stack{\fp{Peter Weir} \\ \fp{Bill Murray} \\ \fp{Brian Helgeland}} & \stack{Howard Shore \\ (hard)} & 50 \\
                \midrule
                d & \stack{Saturn Award for \\ \easy{Best Costume} \\ BAFTA Award for \\ \easy{Best Special Visual Effects} \\ MTV Video Music Award for \\ \easy{Best Female Video}} & \stack{Academy Award for \\ \fp{ Best Supporting Actor} \\ BFCA Award for \\ \fp{Best Supporting Actor} \\ Independent Spirit Award for \\ \fp{Best Male Lead}} & \stack{BFCA Award for \\ \fp{Best Cast} \\ Tony Award for \\ \fp{Best Musical} \\ BFCA Award for \\ \fp{ Best Actress}} & \stack{Golden Globe Award for \\ Best Original Score \\ (hard)} & 26 \\
\bottomrule
            \end{tabular}
        \end{adjustbox}
    \end{minipage}
    \vspace{-0.5em}
\end{table*}
\begin{table*}[!h]
    \centering
    \begin{minipage}{0.17\textwidth}
        \centering
        \includegraphics[width=\textwidth]{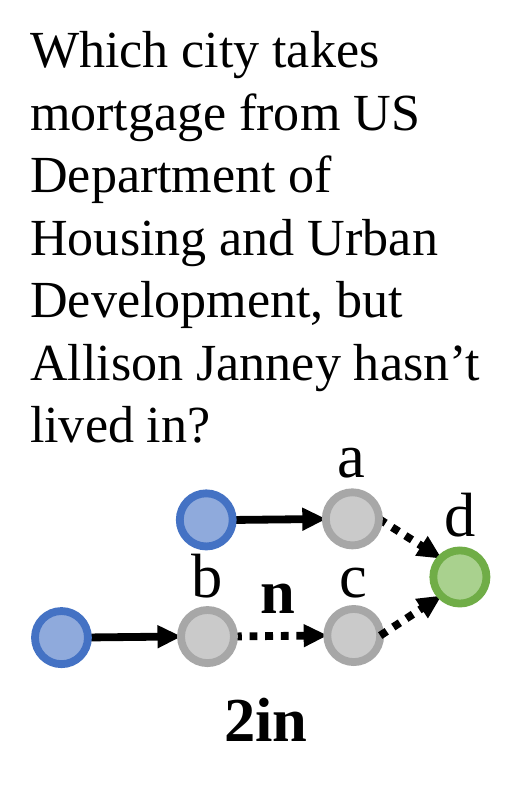}
    \end{minipage}
    \hspace{0.02\textwidth}
    \begin{minipage}{0.8\textwidth}
        \begin{adjustbox}{max width=\textwidth}
            \scriptsize
            \begin{tabular}{lccccc}
                \toprule
                \bf{Query} & \multicolumn{5}{l}{$q = ?d : \text{MortgageSource}(d, \text{US Department of HUD}) \land \neg\text{PlaceLive}(\text{Allison Janney}, d)$} \\
                \midrule
                \multirow{2}{*}{\bf{Variable}} & \multicolumn{3}{c}{\bf{Top Predictions ($\geq 0.1$)}} & \bf{Random} & \bf{Filtered} \\
                & \bf{Easy} & \multicolumn{2}{c}{\bf{Hard}} & \bf{Ground Truth} & \bf{Ranking} \\
                \midrule
                a & \stack{\easy{Tulsa} \\ \easy{Terre Haute} \\ \easy{Santa Monica}} & \stack{\tp{Meridian} \\ \tp{Paducah} \\ \tp{Natchez}} & \stack{\tp{Albany} \\ \tp{Biloxi} \\ \tp{Waukesha}} & \stack{Anderson \\ (hard)} & 1 \\
                \midrule
                b & \stack{- \\ - \\ -} & \stack{\fp{Boston} \\ \fp{Massachusetts} \\ \fp{South Carolina}} & \stack{\fp{New York City} \\ \fp{Ohio} \\ \fp{Plainfield}} & \stack{Dayton \\ (hard)} & 8 \\
                \midrule
                d & \stack{\easy{Lewis County} \\ \easy{Hidalgo County} \\ \easy{Delaware County}} & \stack{\tp{Meridian} \\ \tp{Spokane County} \\ \tp{Albany}} & \stack{\tp{Odessa} \\ \tp{High Point} \\ \tp{Paducah}} & \stack{Anderson \\ (hard)} & 8 \\
                \bottomrule
            \end{tabular}
        \end{adjustbox}
    \end{minipage}
    \vspace{-0.5em}
\end{table*}
\begin{table*}[!h]
    \centering
    \begin{minipage}{0.17\textwidth}
        \centering
        \includegraphics[width=\textwidth]{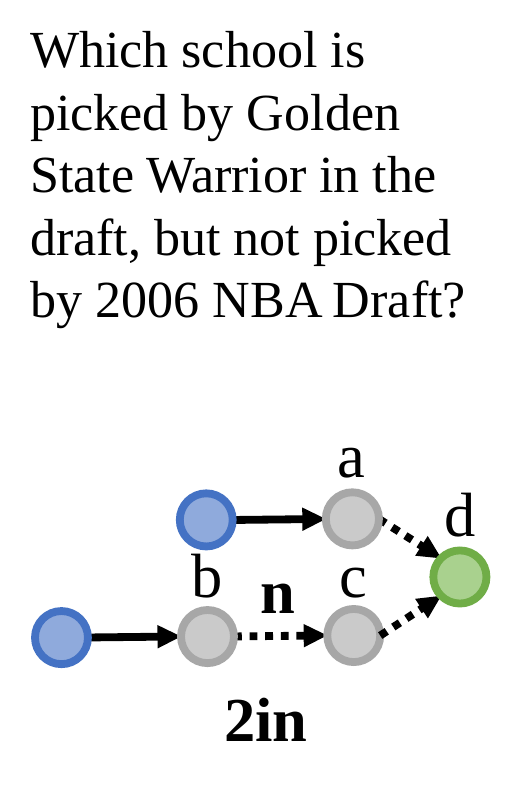}
    \end{minipage}
    \hspace{0.02\textwidth}
    \begin{minipage}{0.8\textwidth}
        \begin{adjustbox}{max width=\textwidth}
            \scriptsize
            \begin{tabular}{lccccc}
                \toprule
                \bf{Query} & \multicolumn{5}{l}{$q = ?d : \text{DraftPickSchool}(\text{Golden State Warriors}, d) \land \neg\text{DraftPickSchool}(\text{2006 NBA Draft}, d)$} \\
                \midrule
                \multirow{2}{*}{\bf{Variable}} & \multicolumn{3}{c}{\bf{Top Predictions ($\geq 0.1$)}} & \bf{Random} & \bf{Filtered} \\
                & \bf{Easy} & \multicolumn{2}{c}{\bf{Hard}} & \bf{Ground Truth} & \bf{Ranking} \\
                \midrule
                a & \stack{\easy{Temple University} \\ \easy{Stanford University} \\ \easy{George Washington University}} & \stack{\fp{Wake Forest University} \\ University of Texas \\ \fp{at Arlington} \\ \fp{Centre College}} & \stack{\fp{University of Vermont} \\ Walsh School \\ \fp{of Foreign Service} \\ \fp{The Wharton School}} & \stack{University of Florida \\ (hard)} & 7 \\
                \midrule
                b & \stack{\easy{Duke University} \\ \easy{University of Kentucky} \\ \easy{University of South Carolina}} & \stack{\fp{University of Kansas} \\ \fp{New Mexico State University} \\ \fp{University of Florida}} & \stack{University of North Carolina \\ \fp{at Chapel Hill} \\ \fp{Boston College} \\ \fp{University of Nevada, Reno}} & \stack{Louisiana State University \\ (easy)} & 1 \\
                \midrule
                d & \stack{\easy{University of San Diego} \\ \easy{College of the Holy Cross} \\ \easy{Southern Methodist University}} & \stack{\fp{Columbia Business School} \\ \fp{The Wharton School} \\ Walsh School \\ \fp{of Foreign Service}} & \stack{\fp{Harvard Business School} \\ Stanford Graduate School \\ \fp{of Business} \\ \fp{Stanford Law School}} & \stack{University of Florida \\ (hard)} & 401 \\
                \bottomrule
            \end{tabular}
        \end{adjustbox}
    \end{minipage}
    \vspace{-0.5em}
\end{table*}
\begin{table*}[!h]
    \centering
    \begin{minipage}{0.17\textwidth}
        \centering
        \includegraphics[width=\textwidth]{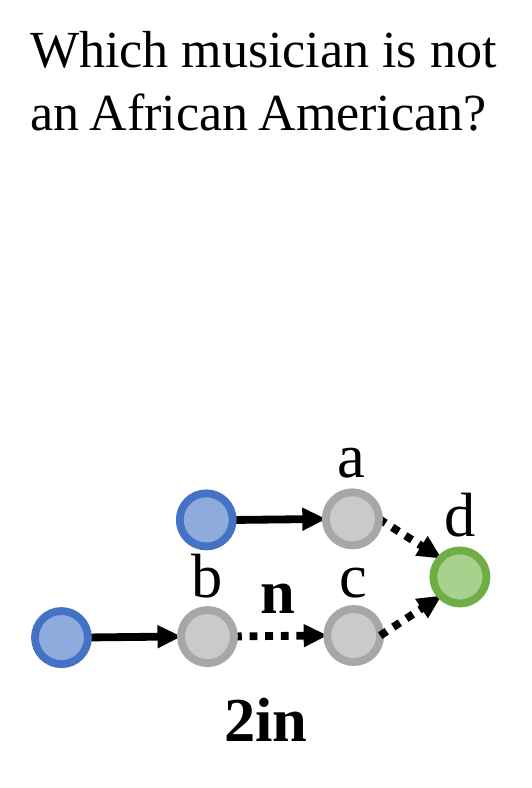}
    \end{minipage}
    \hspace{0.02\textwidth}
    \begin{minipage}{0.8\textwidth}
        \begin{adjustbox}{max width=\textwidth}
            \scriptsize
            \begin{tabular}{lccccc}
                \toprule
                \bf{Query} & \multicolumn{5}{l}{$q = ?d : \text{Profession}(d, \text{musician}) \land \neg\text{People}(\text{African American}, d)$} \\
                \midrule
                \multirow{2}{*}{\bf{Variable}} & \multicolumn{3}{c}{\bf{Top Predictions ($\geq 0.1$)}} & \bf{Random} & \bf{Filtered} \\
                & \bf{Easy} & \multicolumn{2}{c}{\bf{Hard}} & \bf{Ground Truth} & \bf{Ranking} \\
                \midrule
                a & \stack{\easy{Peter Hyams} \\ \easy{Adam Levine} \\ \easy{James Marsters}} & \stack{\fp{Al Jourgensen} \\ \fp{Walter Scharf} \\ \fp{Loudon Wainwright III}} & \stack{\tp{Jon Lord} \\ \fp{Aimee Mann} \\ \fp{Trey Anastasio}} & \stack{Kid Rock \\ (hard)} & 173 \\
                \midrule
                b & \stack{\easy{Fred Williamson} \\ \easy{Maya Rudolph} \\ \easy{Ben Harper}} & \stack{\fp{Yolanda Adams} \\ \fp{LL Cool J} \\ \fp{Flo Rida}} & \stack{\fp{Barry White} \\ \fp{Fat Joe} \\ \tp{Timbaland}} & \stack{Pharrell Williams \\ (easy)} & 1 \\
                \midrule
                d & \stack{\easy{Sting} \\ \easy{Guy Pearce} \\ \easy{Wendy Melvoin}} & \stack{\fp{Arjen Anthony Lucassen} \\ \fp{Georg Solti} \\ \fp{Dmitri Shostakovich}} & \stack{\fp{Jerry Duplessis} \\ \fp{Franz Schubert} \\ \tp{David Gilmour}} & \stack{Kid Rock \\ (hard)} & 278 \\
                \bottomrule
            \end{tabular}
        \end{adjustbox}
    \end{minipage}
    \vspace{-0.5em}
\end{table*}
\begin{table*}[!h]
    \centering
    \begin{minipage}{0.17\textwidth}
        \centering
        \includegraphics[width=\textwidth]{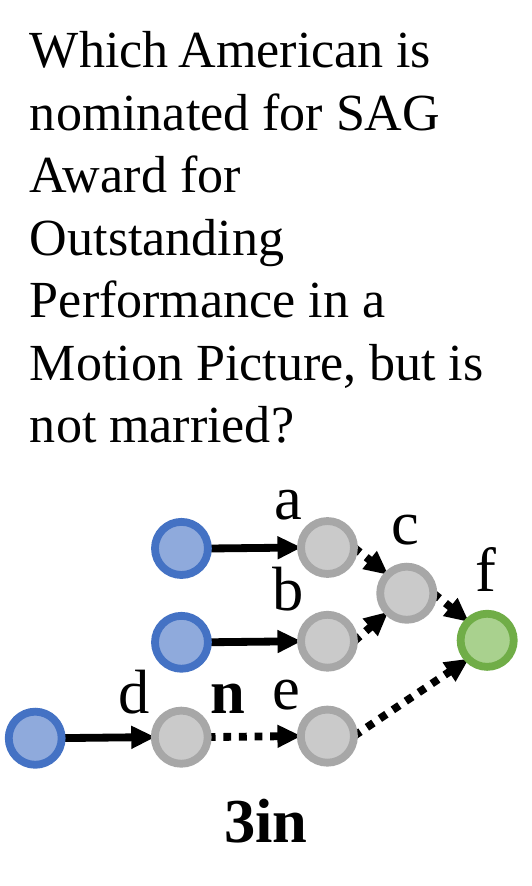}
    \end{minipage}
    \hspace{0.02\textwidth}
    \begin{minipage}{0.8\textwidth}
        \begin{adjustbox}{max width=\textwidth}
            \scriptsize
            \begin{tabular}{lccccc}
                \toprule
                \bf{Query} & \multicolumn{5}{l}{$q = ?f : \text{Nominated}(f, \text{SAG Award for Outstanding Performance in a Motion Picture}) \land \text{Nationality}(f, \text{USA}) \land \neg \text{MarriageType}(f, \text{marriage})$} \\
                \midrule
                \multirow{2}{*}{\bf{Variable}} & \multicolumn{3}{c}{\bf{Top Predictions ($\geq 0.1$)}} & \bf{Random} & \bf{Filtered} \\
                & \bf{Easy} & \multicolumn{2}{c}{\bf{Hard}} & \bf{Ground Truth} & \bf{Ranking} \\
                \midrule
                a & \stack{\easy{Joey Fatone} \\ \easy{Judy Greer} \\ \easy{Dan Hedaya}} & \stack{\tp{Alan Tudyk} \\ \tp{Peter Stormare} \\ \tp{Sam Rockwell}} & \stack{\tp{Jerry Stiller} \\ \tp{Edward Herrmann} \\ \tp{Larry Hagman}} & \stack{Charlize Theron \\ (hard)} & 1 \\
                \midrule
                b & \stack{\easy{Hannah Arendt} \\ \easy{Sylvia Plath} \\ \easy{Vladimir Nabokov}} & \stack{\tp{Ad-Rock} \\ \fp{George III of Great Britain} \\ \tp{Moe Howard}} & \stack{\tp{Tim Duncan} \\ \tp{Frank Herbert} \\ \tp{Lamar Odom}} & \stack{Charlize Theron \\ (easy)} & 1 \\
                \midrule
                c & \stack{\easy{Thomas Jane} \\ \easy{Billy Zane} \\ \easy{Kerry Washington}} & \stack{\tp{Paul Dooley} \\ \tp{Amy Adams} \\ \tp{Jake Gyllenhaal}} & \stack{\tp{Mo'Nique} \\ \tp{Alan Tudyk} \\ \tp{Peter Stormare}} & \stack{Charlize Theron \\ (hard)} & 1 \\
                \midrule
                d & \stack{\easy{Wong Kar-wai} \\ \easy{Zhang Yimou} \\ \easy{Roger Daltrey}} & \stack{\tp{Hugh Dancy} \\ \tp{Greta Scacchi} \\ \tp{Preston Sturges}} & \stack{\tp{Jenny McCarthy-Wahlberg} \\ \tp{Todd Field} \\ \tp{Nathaniel Hawthorne}} & \stack{Joel Silver \\ (easy)} & 1 \\
\bottomrule
            \end{tabular}
        \end{adjustbox}
    \end{minipage}
    \vspace{-0.5em}
\end{table*}
\begin{table*}[!h]
    \centering
    \begin{minipage}{0.17\textwidth}
        \centering
        \includegraphics[width=\textwidth]{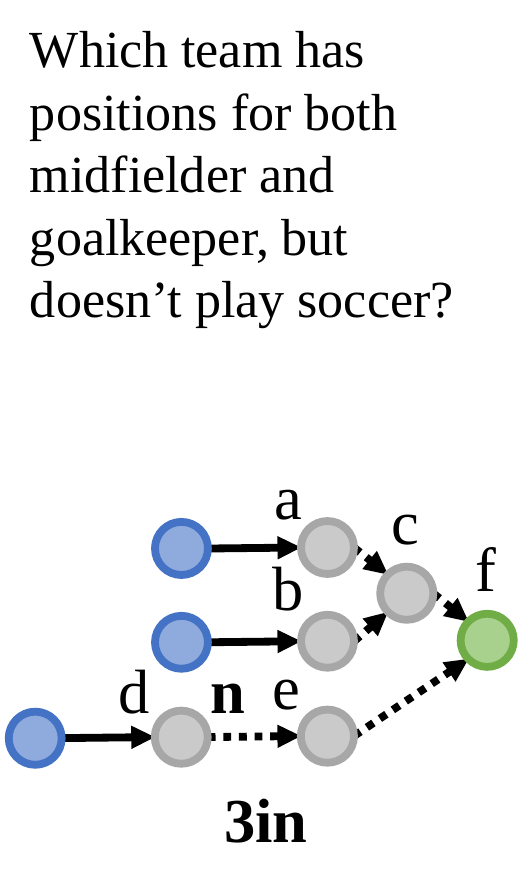}
    \end{minipage}
    \hspace{0.02\textwidth}
    \begin{minipage}{0.8\textwidth}
        \begin{adjustbox}{max width=\textwidth}
            \scriptsize
            \begin{tabular}{lccccc}
                \toprule
                \bf{Query} & \multicolumn{5}{l}{$q = ?f : \text{Position}(f, \text{midfielder (soccer)}) \land \text{Position}(f, \text{goalkeeper (soccer)}) \land \neg \text{Sport}(f, \text{soccer})$} \\
                \midrule
                \multirow{2}{*}{\bf{Variable}} & \multicolumn{3}{c}{\bf{Top Predictions ($\geq 0.1$)}} & \bf{Random} & \bf{Filtered} \\
                & \bf{Easy} & \multicolumn{2}{c}{\bf{Hard}} & \bf{Ground Truth} & \bf{Ranking} \\
                \midrule
                a & \stack{\easy{Hannover 96} \\ \easy{RB Leipzig} \\ \easy{FC Bayern Munich}} & \stack{\fp{Bulgaria soccer team} \\ \fp{Kazakhstan soccer team} \\ \fp{Croatia soccer team}} & \stack{\fp{Chile soccer team} \\ \fp{FC Torpedo Moscow} \\ \fp{FC Energie Cottbus}} & \stack{Wimbledon F.C. \\ (hard)} & 130 \\
                \midrule
                b & \stack{\easy{RB Leipzig} \\ \easy{Hannover 96} \\ \easy{FC Volga Nizhny Novgorod}} & \stack{\fp{D.C. United} \\ \fp{F.C. Lorient} \\ \fp{Coritiba F.C.}} & \stack{\fp{South Africa soccer team} \\ \fp{Croatia soccer team} \\ \fp{Suwon Samsung Bluewings FC}} & \stack{Wimbledon F.C. \\ (easy)} & 1 \\
                \midrule
                c & \stack{\easy{Hannover 96} \\ \easy{RB Leipzig} \\ \easy{FC Volga Nizhny Novgorod}} & \stack{\fp{D.C. United} \\ \fp{F.C. Lorient} \\ \fp{Bulgaria soccer team}} & \stack{\fp{Kazakhstan soccer team} \\ \fp{Coritiba F.C.} \\ \fp{FC Torpedo Moscow}} & \stack{Wimbledon F.C. \\ (hard)} & 202 \\
                \midrule
                d & \stack{\easy{India soccer team} \\ \easy{Levante UD} \\ \easy{Singapore soccer team}} & \stack{\tp{Croatia soccer team} \\ \tp{Tunisia soccer team} \\ \tp{Colombia soccer team}} & \stack{\tp{1. FSV Mainz 05} \\ \tp{Esporte Clube Bahia} \\ \tp{Legia Warsaw}} & \stack{PFC CSKA Moscow \\ (easy)} & 1 \\
\bottomrule
            \end{tabular}
        \end{adjustbox}
        \footnotetext{In reality, any team has a midfielder or a goalkeeper plays soccer, and this query has no answer. The failure of generating this query is due to the incompleteness of FB15k-237.}
    \end{minipage}
    \vspace{-0.5em}
\end{table*}
\begin{table*}[!h]
    \centering
    \begin{minipage}{0.17\textwidth}
        \centering
        \includegraphics[width=\textwidth]{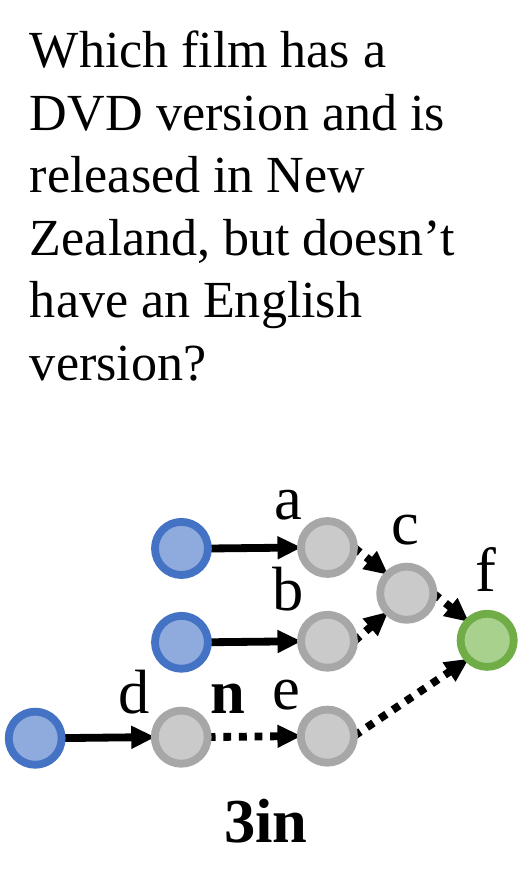}
    \end{minipage}
    \hspace{0.02\textwidth}
    \begin{minipage}{0.8\textwidth}
        \begin{adjustbox}{max width=\textwidth}
            \scriptsize
            \begin{tabular}{lccccc}
                \toprule
                \bf{Query} & \multicolumn{5}{l}{$q = ?f : \text{ReleaseMedium}(f, \text{DVD}) \land \text{ReleaseRegion}(f, \text{New Zealand}) \land \neg \text{Language}(f, \text{English})$} \\
                \midrule
                \multirow{2}{*}{\bf{Variable}} & \multicolumn{3}{c}{\bf{Top Predictions ($\geq 0.1$)}} & \bf{Random} & \bf{Filtered} \\
                & \bf{Easy} & \multicolumn{2}{c}{\bf{Hard}} & \bf{Ground Truth} & \bf{Ranking} \\
                \midrule
                a & \stack{\easy{The Verdict} \\ \easy{The Interpreter} \\ \easy{The Love Guru}} & \stack{\tp{The Matrix 3} \\ \tp{Get Him to the Greek} \\ \tp{Shattered Glass}} & \stack{\tp{A View to a Kill} \\ \tp{Airplane!} \\ Superman/Batman: \\ \fp{Public Enemies}} & \stack{Detective Dee and \\ the Mystery of the Phantom Flame \\ (hard)} & 219 \\
                \midrule
                b & \stack{\easy{The Tree of Life} \\ \easy{Total Recall} \\ \easy{Prometheus}} & \stack{Snow White and \\ \tp{the Huntsman} \\ \fp{Submarine} \\ \fp{Tyrannosaur}} & \stack{\fp{This Is England} \\ \fp{The Host} \\ \tp{Argo}} & \stack{Detective Dee and \\ the Mystery of the Phantom Flame \\ (easy)} & 1 \\
                \midrule
                c & \stack{\easy{The Expendables 2} \\ \easy{Soul Surfer} \\ \easy{National Security}} & \stack{\tp{The Butterfly Effect} \\ \fp{On the Road} \\ \fp{Monsters, Inc.}} & \stack{\tp{Pineapple Express} \\ Snow White and \\ \tp{the Huntsman} \\ \fp{Hyde Park on Hudson}} & \stack{Detective Dee and \\ the Mystery of the Phantom Flame \\ (hard)} & 98 \\
                \midrule
                d & \stack{\easy{Zero Dark Thirty} \\ \easy{Slaughterhouse-Five} \\ \easy{The Butterfly Effect}} & \stack{\tp{Going My Way} \\ \tp{She's Having a Baby} \\ \tp{Julius Caesar}} & \stack{\tp{The Wizard of Oz} \\ \tp{The Country Girl} \\ \tp{The Right Stuff}} & \stack{Dangerous Liaisons \\ (easy)} & 1 \\
\bottomrule
            \end{tabular}
        \end{adjustbox}
    \end{minipage}
    \vspace{-0.5em}
\end{table*}
\begin{table*}[!h]
    \centering
    \begin{minipage}{0.17\textwidth}
        \centering
        \includegraphics[width=\textwidth]{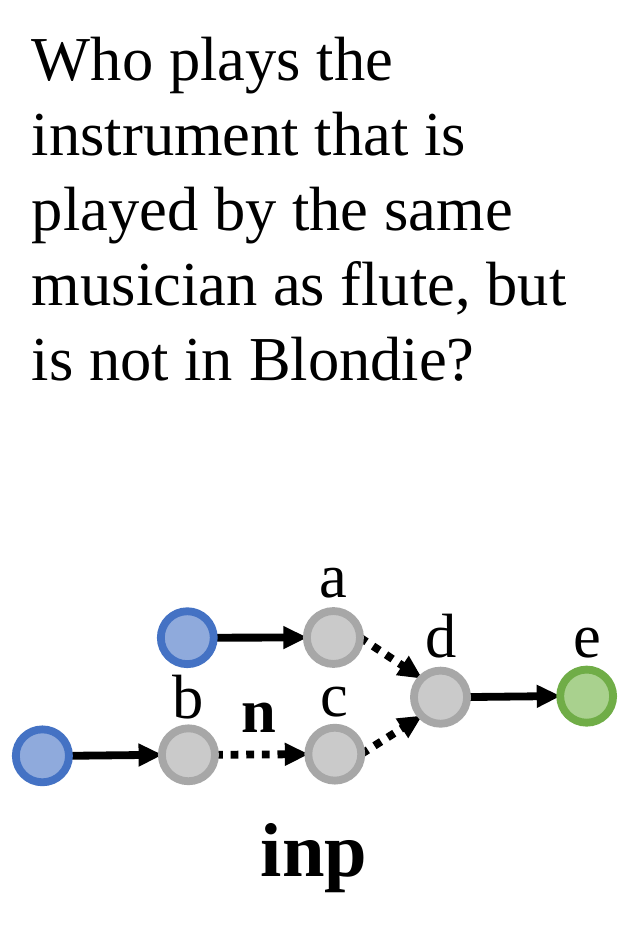}
    \end{minipage}
    \hspace{0.02\textwidth}
    \begin{minipage}{0.8\textwidth}
        \begin{adjustbox}{max width=\textwidth}
            \scriptsize
            %
        \end{adjustbox}
    \end{minipage}
    \vspace{-0.5em}
\end{table*}
\begin{table*}[!h]
    \centering
    \begin{minipage}{0.17\textwidth}
        \centering
        \includegraphics[width=\textwidth]{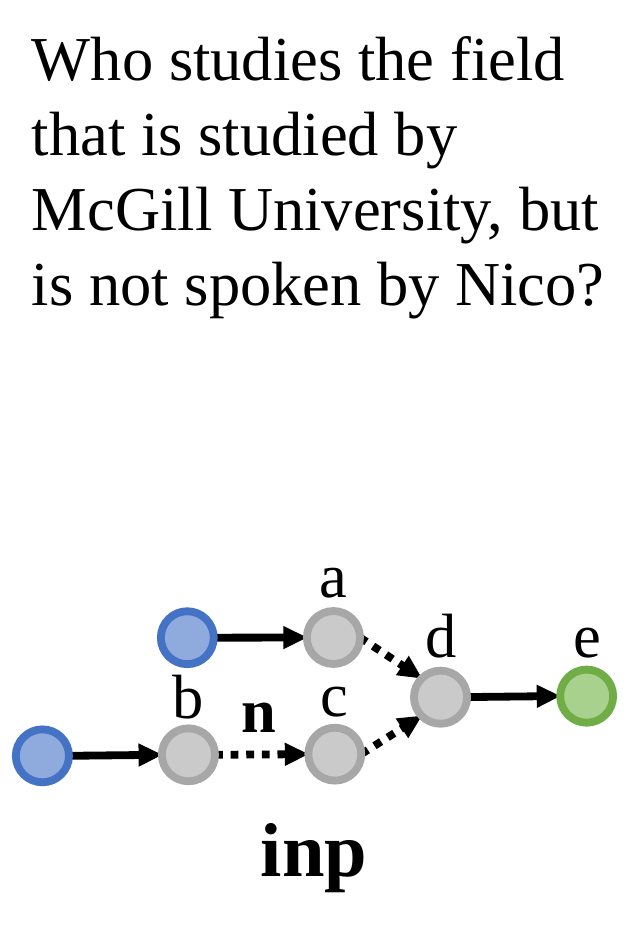}
    \end{minipage}
    \hspace{0.02\textwidth}
    \begin{minipage}{0.8\textwidth}
        \begin{adjustbox}{max width=\textwidth}
            \scriptsize
            %
        \end{adjustbox}
        \footnotetext{In reality, McGill University studies all the fields that our model predicts for variable $a$ and $d$. The predictions are considered wrong because the corresponding facts are missing in FB15k-237.} 
    \end{minipage}
    \vspace{-0.5em}
\end{table*}
\begin{table*}[!h]
    \centering
    \begin{minipage}{0.17\textwidth}
        \centering
        \includegraphics[width=\textwidth]{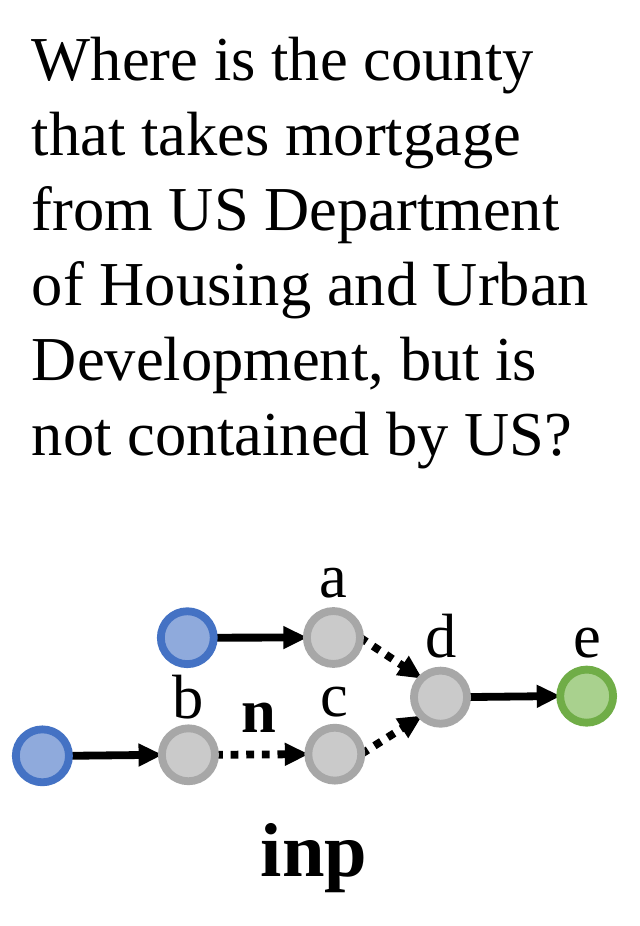}
    \end{minipage}
    \hspace{0.02\textwidth}
    \begin{minipage}{0.8\textwidth}
        \begin{adjustbox}{max width=\textwidth}
            \scriptsize
            %
        \end{adjustbox}
        \footnotetext{In reality, US Department of HUD only provides mortage to US cities, and this query has no answer. The failure of generating this query is due to the incompleteness of FB15k-237.}
    \end{minipage}
    \vspace{-0.5em}
\end{table*}
\begin{table*}[!h]
    \centering
    \begin{minipage}{0.17\textwidth}
        \centering
        \includegraphics[width=\textwidth]{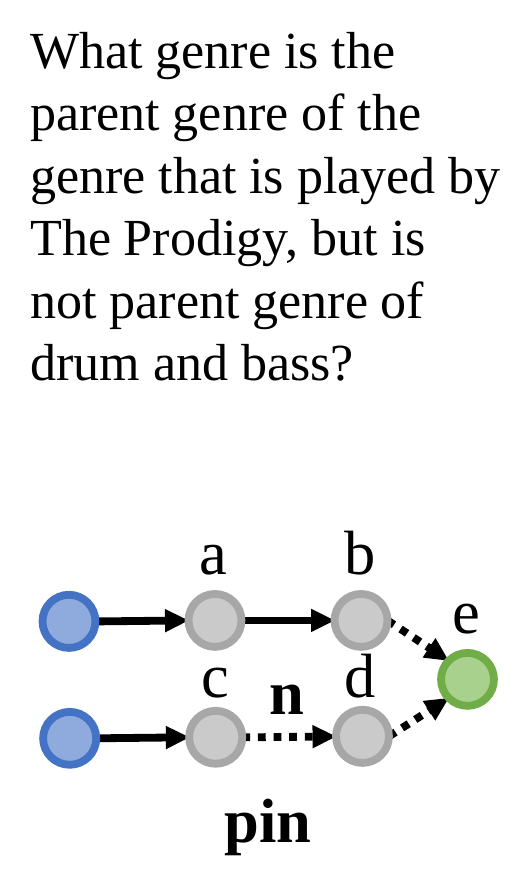}
    \end{minipage}
    \hspace{0.02\textwidth}
    \begin{minipage}{0.8\textwidth}
        \begin{adjustbox}{max width=\textwidth}
            \scriptsize
            %
        \end{adjustbox}
    \end{minipage}
    \vspace{-0.5em}
\end{table*}
\begin{table*}[!h]
    \centering
    \begin{minipage}{0.17\textwidth}
        \centering
        \includegraphics[width=\textwidth]{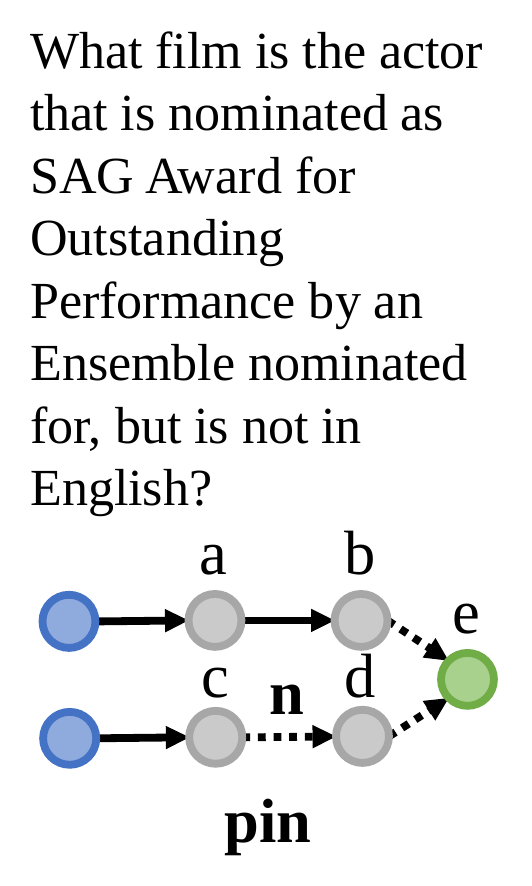}
    \end{minipage}
    \hspace{0.02\textwidth}
    \begin{minipage}{0.8\textwidth}
        \begin{adjustbox}{max width=\textwidth}
            \scriptsize
            %
        \end{adjustbox}
        \footnotetext{In reality, all films that win SAG Award have English versions. The failure of generating this query is due to the incompleteness of FB15k-237.}
    \end{minipage}
    \vspace{-0.5em}
\end{table*}
\begin{table*}[!h]
    \centering
    \begin{minipage}{0.17\textwidth}
        \centering
        \includegraphics[width=\textwidth]{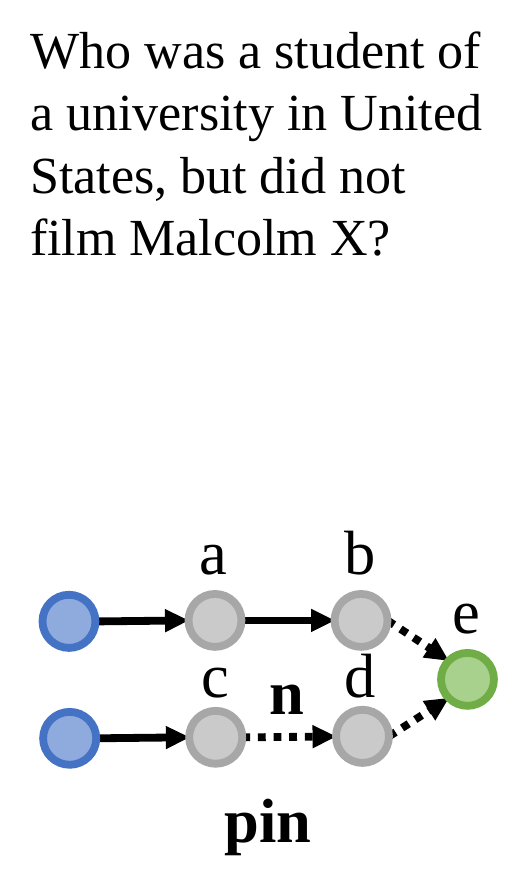}
    \end{minipage}
    \hspace{0.02\textwidth}
    \begin{minipage}{0.8\textwidth}
        \begin{adjustbox}{max width=\textwidth}
            \scriptsize
            %
        \end{adjustbox}
    \end{minipage}
    \vspace{-0.5em}
\end{table*}
\begin{table*}[!h]
    \centering
    \begin{minipage}{0.17\textwidth}
        \centering
        \includegraphics[width=\textwidth]{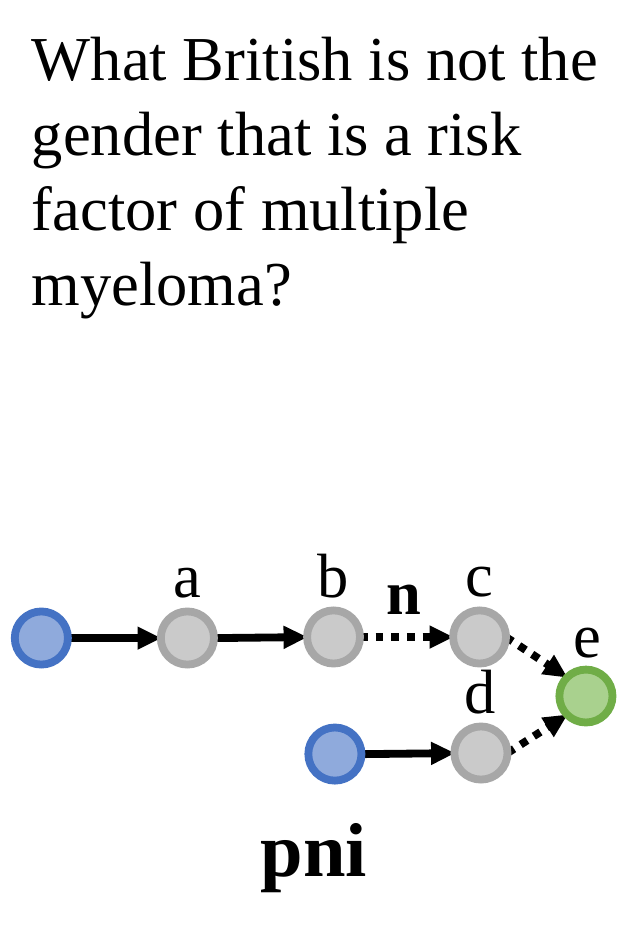}
    \end{minipage}
    \hspace{0.02\textwidth}
    \begin{minipage}{0.8\textwidth}
        \begin{adjustbox}{max width=\textwidth}
            \scriptsize
            %
        \end{adjustbox}
    \end{minipage}
    \vspace{-0.5em}
\end{table*}
\begin{table*}[!h]
    \centering
    \begin{minipage}{0.17\textwidth}
        \centering
        \includegraphics[width=\textwidth]{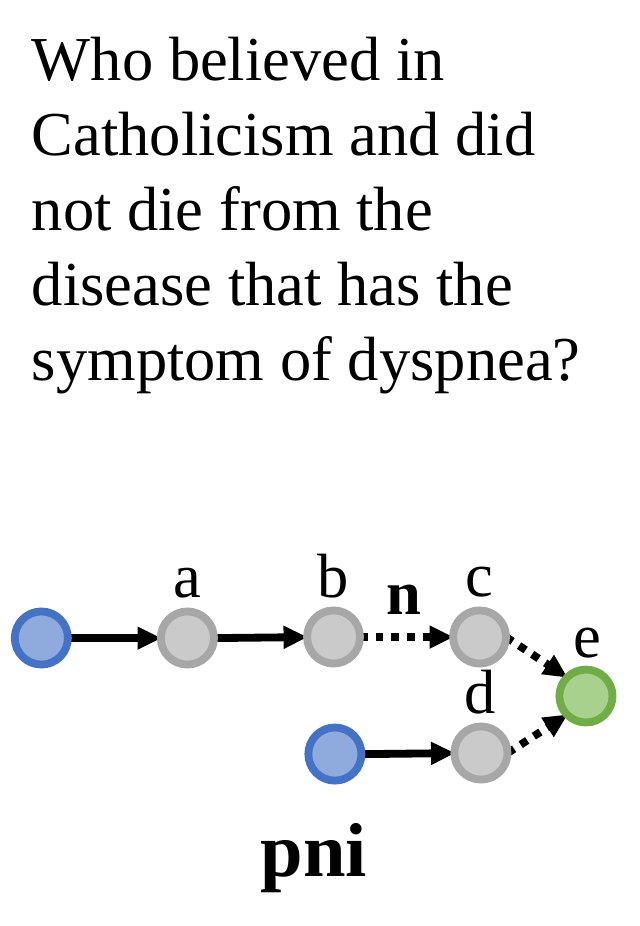}
    \end{minipage}
    \hspace{0.02\textwidth}
    \begin{minipage}{0.8\textwidth}
        \begin{adjustbox}{max width=\textwidth}
            \scriptsize
            %
        \end{adjustbox}
    \end{minipage}
    \vspace{-0.5em}
\end{table*}
\begin{table*}[!h]
    \centering
    \begin{minipage}{0.17\textwidth}
        \centering
        \includegraphics[width=\textwidth]{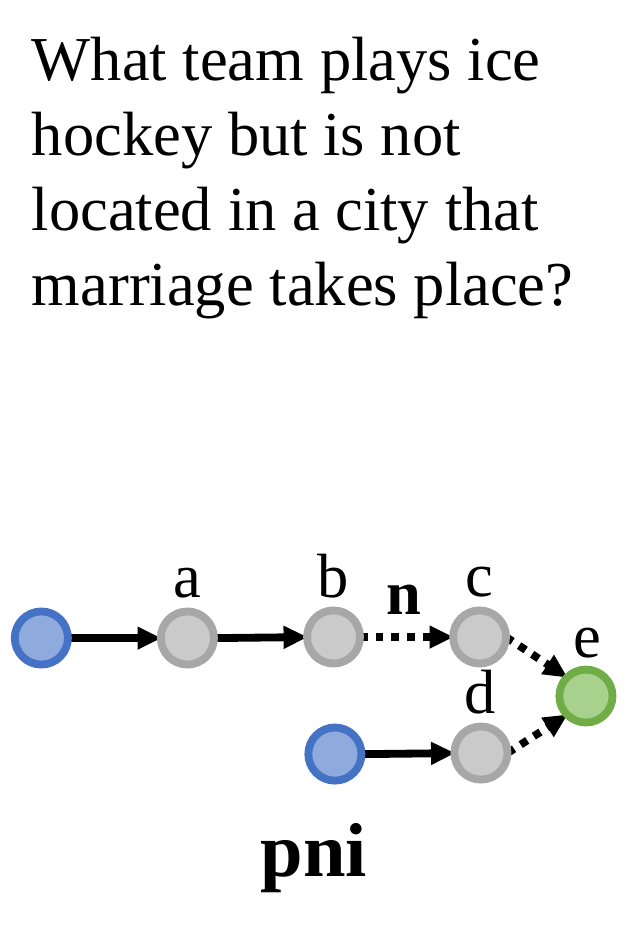}
    \end{minipage}
    \hspace{0.02\textwidth}
    \begin{minipage}{0.8\textwidth}
        \begin{adjustbox}{max width=\textwidth}
            \scriptsize
            %
        \end{adjustbox}
        \footnotetext{In reality, marriage can take place in any city, and this query has no answer. The failure of generating this query is due to the incompleteness of FB15k-237.}
    \end{minipage}
    \vspace{-0.5em}
\end{table*}

\end{document}